**Teză de doctorat**

**The Analysis of Lexical Errors in Machine Translation from English into Romanian**

**(Analiza erorilor lexicale în traducerea automată din engleză în română)**

**Coordonator**                                     **Doctorand**

**Prof.univ.dr. Ruxandra Vișan**          **Angela Stamatie (Dumitran)**

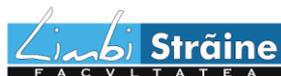

**2025**

# Acknowledgements

I would like to express my profound gratitude to my advisor, Professor Ruxandra Vişan, PhD, for her invaluable guidance, steadfast support, and insightful mentorship throughout the process of completing this doctoral thesis. Her expertise, encouragement, and dedication have been crucial in shaping both my research and academic development.

I am also deeply thankful to my esteemed committee members, professor Ruxandra Vişan, PhD, associate professor Anca Sevcenco, PhD, associate professor Daria Protopopescu, PhD, professor Larisa Avram, PhD, whose constructive feedback and scholarly insights have significantly enhanced the quality of this thesis. Their expertise and encouragement have played an essential role in guiding my research efforts.

Finally, I extend my heartfelt thanks to my family for their unwavering love, support, and understanding during this academic journey. Their constant encouragement and belief in me have been indispensable in reaching this important milestone in my academic career.

Their collective support and encouragement have been indispensable in realizing this milestone in my academic journey.

Thank you!

# Main symbols and abbreviations

ARI - Acute Respiratory Infection

ATI – Anestezie şi terapie intensivă (Romanian abbreviation for Intensive Care Unit)

BLEU – BiLingual Evaluation Understudy

CL – Computational Linguistics

COVID-19 – Corona Virus Disease-19

DOOM – Dicţionar ortografic, ortoepic al morfologiei limbii române (Orthographic, Orthoepic and Morphologic Dictionary of the Romanian language)

EA – Error Analysis

GT – Google Translation

MT – Machine Translation

NMT – Neural Machine Traanslation

OED – Oxford English Dictionary

RbMT – Ruled based Machine Translation

SMT – Statistical Machin Translation

WER – Word Error Rate

WHO – World Health Organization

# Abstract

The research explores error analysis in the performance of translating by Machine Translation from English into Romanian, and it focuses on lexical errors found in texts which include official information, provided by the World Health Organization (WHO), the Gavi Organization, by the patient information leaflet (the information about the active ingredients of the vaccines or the medication, the indications, the dosage instructions, the storage instructions, the side effects and warning, etc.). All of these texts are related to Covid-19 and have been translated by Google Translate, a multilingual Machine Translation that was created by Google. In the last decades, Google has actively worked to develop a more accurate and fluent automatic translation system. This research, specifically focused on improving Google Translate, aims to enhance the overall quality of Machine Translation by achieving better lexical selection and by reducing errors. The investigation involves a comprehensive analysis of 230 texts that have been translated from English into Romanian. These texts have been meticulously selected to cover a wide range of topics and linguistic structures, and they provide a robust dataset for evaluating and refining translation algorithms. By addressing common issues such as collocation errors, semantic inaccuracies, and context misinterpretations, the research seeks to identify patterns and develop solutions that can be integrated into the Google Translate system. The ultimate goal of the thesis is to produce translations that are not only linguistically accurate but also contextually appropriate, thereby improving user satisfaction and expanding the usability of translation tools in diverse scenarios.

The chosen texts are all related to coronavirus, since the Covid-19 pandemic has been a challenge for MT researchers and translators who have been pressured to foster the development of tools and resources for improving access to information about coronavirus all over the world and for enhancing the developments of language tools and resources. This study aims to: (1) find the prevalent types and patterns of lexical errors encountered in Covid-19 texts translated from English into Romanian by Google Translate, and analyse how these errors impact the accuracy and fluency of the translated content; (2) find how the complexity and specificity of Covid-19 terminology influences the occurrence and severity of lexical errors in translations produced by Google Translate from English into Romanian, and what strategies can be developed to mitigate these errors effectively; (3) establish to what extent the structural differences between English and Romanian contribute to the occurrence of lexical errors in translations of Covid-19 texts by Google Translate; (4) discuss how computational linguistics and machine learning techniques be leveraged to improve the quality and reliability of automated translations in this domain. The study results reveal that out of the 230 texts analysed, there was a notable distribution of errors across various linguistic categories. Specifically, 9.6% of the errors were attributed to misinterpretations of abbreviations and acronyms, indicating a need for improved contextual understanding and disambiguation in these areas. Additionally, 10.3% of errors stemmed from false friends, where words in the source and target languages appear similar but differ significantly in meaning. The presence of such errors highlights a critical area for lexical refinement. Borrowings, where foreign words are integrated into the target language, accounted for 17.2% of the errors, showing a significant impact on translation accuracy. Distortions and sense relation confusion both constituted 4.8% of the errors, pointing to challenges in maintaining semantic integrity and in accurately conveying nuanced meanings.

The most dominant type of error identified was the collocation error, which made up 31.7% of the total errors. This high percentage underscores the importance of improving the ability of the translation system to correctly pair words and phrases that usually occur together in the target language. Arbitrary combinations, where words are combined in ways that do not make sense in the target language, were relatively infrequent at 1.3%, but still present a noticeable area for improvement. Preposition partner (complement) errors, involving inaccurate prepositional usage, represented 9.6% of the errors, which suggests a need for better syntactic and grammatical alignment. Lastly, stylistic errors accounted for 10.3% of the errors, indicating that while the translations might be technically correct, they often fail to match the stylistic nuances of the target language, thereby affecting the overall fluency and readability of the translated text. These findings highlight specific areas where targeted improvements could significantly enhance the quality of Machine Translations in Google Translate.

# Table of Contents

# Chapter I

# 1. Introduction

The research examines error analysis in Machine Translation from English into Romanian, targeting lexical errors in official Covid-19 information from medical documents from sources such as the World Health Organization (WHO), the Gavi Organization, etc. produced by Google Translate. By focusing on lexical errors in official Covid-19 information translated by Google Translate, this research study *aims* to identify lexical errors of specific texts that were translated by MT. At present, MT is one of the most significant application areas for computational linguistics and, as Fasold et al. (2013: 451) claim, it is " focused on translations which preserve the information content of the source language as much as possible, while rendering it in a natural form in the target language". A tool employed at this time by many users, MT systems have become a powerful instrument for rendering texts into the target language, by exploring different types of approaches to non-human translation. The main ones, according to Lopez (2008) and Wu et al. (2016), are Statistical Machine Translation (SMT), that utilizes statistical translation models generated from the analysis of monolingual and bilingual training data; Rule-Based Machine Translation (RBMT), that relies on built-in linguistic rules and millions of bilingual dictionaries for each language; and Neural Machine Translation (NMT), that depends on neural network models, based on the human brain, a model that was approached by Google Translate beginning with 2016. One needs to underline that the main similarity between the neural network models and the human brain is the presence of neurons as the basic unit of the system. Hassabis et al. (2017) discuss how artificial neural networks (ANNs) draw inspiration from the brain's structure and function, particularly in mimicking neurons and their connectivity, which are central to both biological and artificial systems. The neurons employed are artificial, but based on the model of biological neurons. Richards et al. (2019) elaborate on the conceptual parallels between deep learning neural networks and the brain's neural circuits, emphasizing how artificial neurons are inspired by biological neurons and their role in processing information. These artificial neurons are interconnected and process information in layers. Each layer performs specific transformations on the data, ultimately leading to the translated text.

In neural networks, particularly those used in models like Google Translate, the artificial neurons are organized into layers – input, hidden, and output layers. According to Bengio et al. (2016), who have focused on the architecture of neural networks and on the function of different neuron layers in processing and transforming data, each layer performs specific transformations on the data it receives. The input layer takes in raw data, such as text, and passes it on to the hidden layers. The hidden layers, which may be composed of multiple layers in deep neural networks, are where the real computational work happens. Here, artificial neurons apply mathematical functions, adjusting weights and biases, to transform the input data into increasingly abstract representations. For instance, in the context of Machine Translation, the initial layers might capture basic linguistic features like words and phrases, while deeper layers might capture more complex structures such as syntax and semantics. As the data moves through these layers, the model learns to recognize patterns and relationships in the language, which are crucial for generating accurate translations.

The focus of the present research falls on errors, as will be shown in chapter IV. By identifying and classifying errors and their causes, it is possible to anticipate the statistics applicable to MT. These statistics will provide a useful tool for comparing different systems of analysing the performance such as Word Error Rate, BLEU, etc. in order to evaluate the MT. Therefore, we seek to fulfil the following *objectives*: 1. to examine types of lexical translation errors from English into Romanian; 2. to establish the causes of these errors or find possible explanations for them; 3. to evaluate the quality and accuracy of Google Translate when translating health information from English into Romanian.

The data used for investigation comprises a corpus[1] of 230 texts that have been translated from English into Romanian by Google Translate. These texts were collected from the official website of the World Health Organization[2] and the Gavi Organization[3] (from March 2020 to the present). Moreover, the chosen texts are all related to coronavirus, as the Covid-19 pandemic is a challenge for MT researchers and for translators who have been pressured to foster the development of tools and resources for enhancing access to information about coronavirus all over the world and improve the developments of languages tools and resources. Covid-19 pandemic forced people worldwide to become familiar with public health concepts that were previously not part of everyday conversations, as Piller and Zhang (2020: 504) by emphasizing that "almost everyone in the world has had to learn about public health concepts such as *social distancing*, *droplet transmission* or *flattening the curve* to avoid getting sick. Almost everyone has had to understand the specifics of containment measures such as lockdowns, contact tracing, or mask wearing in their jurisdiction (…) global knowledge dissemination was woefully limited to a small number of languages as the world entered the pandemic". Piller and Zhang (2020: 506) state that due to the pandemic situation, there were "problems related to translation and multilingual terminology standards in public health information and medical research". This stark reality highlighted significant linguistic and communicative gaps that contributed to the difficulties in achieving effective public health communication on a global scale.

---

[1] The links of the texts that constitute the corpus are provided in an appendix of the thesis.
[2] https://www.who.int/.
[3] https://www.gavi.org/.

The limitations in translation and multilingual terminology standards created barriers to understanding crucial health information, particularly in languages that had not achieved global status or which were not as widely used. Consequently, large segments of the global population were left vulnerable due to the lack of accessible, accurate, and timely information in their native languages. The deficiencies in the translation of public health materials underscored the need for more robust and inclusive language policies that address the linguistic diversity of global populations, ensuring that life-saving information reaches everyone, regardless of their language. This situation further emphasized the importance of improving translation accuracy and the dissemination of multilingual public health information in order to better prepare for future global health crises. It is widely acknowledged that English is considered a "lingua franca" or "a global language". The predominance of English in most publications and conferences, news, etc. has led to its widespread adoption as the primary medium for scholarly communication. In the era of globalization, the importance of English continues to grow, enabling the dissemination of research findings and the exchange of ideas on an international scale. As previous researchers have shown, the dominance of English presents both opportunities and challenges. While it facilitates international collaboration, it risks marginalizing scholars whose primary language is not English. One of the best-known researchers who has focused on the intersection between English and globalization, Crystal (2003: 190) states that "the emergence of English with a genuine global presence therefore has a significance which goes well beyond this particular language". Examining the range of historical factors that led to the current situation and all types of factors, such as politics, commerce, internet, and various aspects of life, Crystal (2003) has extensively studied the aspects that have contributed to English becoming the a "lingua franca". According to Crystal and to other authors and researchers such as Blommaert (2010), Albl-Mikasa et al. (2024) who have studied the topic, English has solidified its status as a global language due to its dominance in international business, finance, and trade, as well as to its significant presence in digital platforms such as websites, social media, and online publications. The internet and information technology, largely developed in English-speaking countries, have further propelled the spread of English worldwide. This widespread use of English in various domains ensures that it remains a central language in global communication, which is mirrored in the functioning of translation technologies like Google Translate, further entrenching English's global role. According to Turovsky (2016), in the context of Google Translate, the prominence of English as a lingua franca is reflected in the translation system's reliance on English as an intermediary language for many translations.

On a related note, it is relevant to be aware of the consequences given by the distinction between the major and the minor languages. Crystal (2003: 3) explains that English can be seen as a "major language" as it is a language that achieves "a genuinely global status when it develops a special role that is recognized in every country". *Major languages* are the official languages of the United Nations: Arabic, Chinese, English, French, Russian, and Spanish, according to www.un.org. Whereas *minor languages*, as defined by Mikhailov (2015: 952), are those languages with no high diffusion, fewer text corpora and terminology banks available, aspects that lead to Machine Translation performance that is either unavailable or of poor quality. Based on the available data, we tentatively place Romanian in the second category, acknowledging that further analysis may be needed to confirm this classification. However, due to the increasing availability of text corpora in Romanian, the situation might be seen as fluctuating in the present moment.

# 2. Research background

## 2.1. MTness and MT generalities

Google Translate was chosen for this study, since several prominent researchers such as Nizke (2019: 18) see it as "one of the most famous online MT systems nowadays". Moreover, it is free, easily accessed, and enables instant online translation. As Machine Translation technology has advanced, tools like Google Translate have become indispensable for disseminating information quickly and widely. According to www. https://blog.google/, in April 2021, Google Translate provided free translation services for 109 languages. Moreover, in June 2024, Isaac Caswell, a Senior Software Engineer at Google Translate, announced, on the website mentioned above, the addition of 110 new languages, bringing the total to over 240 supported languages. Its popularity stems from its ability to simulate human translation. However, it struggles with accurately translating specialized medical terms, often producing inaccurate Romanian equivalents, even though it remains highly accessible. Despite its accessibility and wide language coverage, Google Translate often struggled with accurately translating critical terminology during the pandemic. Lexical errors led to mistranslations that could distort key messages, highlighting the limitations of automated systems in high-stakes situations. Thus, the pandemic emphasized the need for improved Machine Translation capable of handling complex, precise language in critical contexts. Moreover, the emergence of new Covid-related terminology has greatly influenced the English language, with linguistic consequences highlighting its crucial role in disseminating information about the coronavirus during the pandemic. As a result, the demand for translations increased globally within a year, particularly in countries where the population does not speak languages recognized as global languages.

The main reason for choosing lexical errors from Google Translate in texts provided by the World Health Organisation and Gavi Organisation is the rapid outbreak of Covid-19 across the globe that has raised a growing need to access information from English. It is important to underline that the official information was accessible in 4 to 9 *major languages*[4]. Due to this, the necessity of making the official information available in languages other than *major* ones became stringent in the pandemic context.

---

[4] According to the official website of WHO, Arabic, Chinese, English, French, Russian and Spanish are the official languages that were established by World Health Assembly in 1978 (https://www.who.int/about/policies/multilingualism#:~:text=WHO's%20six%20official%20languages%20%2D %20Arabic,multilingualism%20into%20a%20WHO%20policy.).

Based on the survey made on https: //suveys.google.com/, we illustrated in Figure 1 some of the most popular MT systems that have been used since 2020.

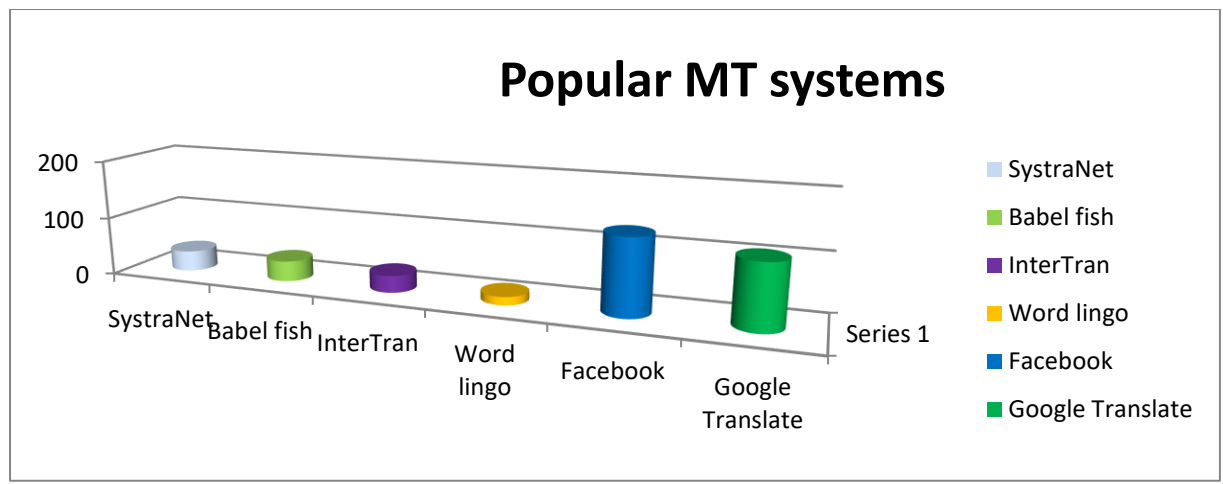

Figure 1. Popular MT systems

Google Translate Tools were the most used during the pandemic, according to Google Trends[5]. The highest rate is taken by Facebook, which is, in fact, made up of Bing Microsoft Translator Tools, one of the most used MT on social media platforms. During pandemic times, mainstream platforms, such as Facebook, Instagram, YouTube, Twitter, have provided an unprecedented amount of information related to Covid-19 all over the world, and this is the reason why Bing Microsoft Translator was one of the most used translation tools. By using a concept named *infodemic*, Cinelli et al. (2020: 165) have underlined the idea that the information related to Covid-19 spreads even faster than the epidemic itself, by saying that "as an example, CNN has recently anticipated a rumour about the possible lock-down of Lombardy (…), publishing the news hours before the official communication from the Italian Prime Minister. As a result, people overcrowded trains and airports to escape". As emphasized, the premature dissemination of such information undermined measures aimed at preventing contamination in public spaces. In their study, Cinelli et al. (2020) conducted a comparative analysis of user activity on social media platforms, including Twitter, Instagram, YouTube, Reddit, and Gab, examining content consumption dynamics during the specified time period. Cinelli et al. (2020: 116) state that significant differences were found in how information related to Covid-19 was consumed and shared across various social media platforms. For instance, platforms like Twitter and Reddit were primarily used for real-time updates and discussions, while Instagram and YouTube favoured more visual and curated content. Gab, a platform known for its minimal content moderation, showed a higher prevalence of misinformation and conspiracy theories. The analysis revealed that the speed and reach of information varied widely depending on the platform's structure and user base, with misinformation spreading particularly fast on platforms with weaker moderation policies. In this manner, the study highlighted that these dynamics contributed to the infodemic, amplifying confusion and fear among the public. Such analyses underscore the importance of developing strategies to curb the spread of misinformation and ensure that reliable, accurate information prevails during critical events such as a global pandemic.

Moreover, Anastasopoulos et al. (2020) have shown that a concept called TICO-19 (Translation Initiative for Covid-19) has been created. TICO-19 represents a collaborative effort aimed to facilitate the rapid and accurate translation of critical information related to the Covid-19 pandemic across multiple languages (English, Spanish, French, Chinese, Arabic, Russian, Portuguese, Hindi, Bengali and Swahili). According to the information from the website, https://tico-19.github.io/, this initiative recognizes the importance of timely and accessible communication in addressing public health emergencies. It emphasizes the need for a rapid response to evolving Covid-19 situations by establishing a network of translators and linguists capable of producing timely translations in response to emerging needs and developments. This initiative fosters collaboration among translators, linguists, public health organizations, governmental agencies that dealt with the linguistic challenges posed by the pandemic, and it is relevant to mention in this study in order to emphasize the need to bring professional expertise in translating Covid-19 related materials. TICO-19 exemplifies the crucial role that translation plays in global health communication, particularly in the context of public health emergencies such as the Covid-19 pandemic.

## 2.2. Linguistic validation of the vocabulary related to Covid-19

The Covid-19 pandemic not only reshaped global health policies and societal structures, but also had a profound impact on language, particularly in the rapid dissemination and adoption of new terminology. As the virus spread across linguistic and cultural boundaries, so did the vocabulary associated with it, influencing both everyday discourse and specialized communication. For example, the term "pandemic" highlights the worldwide impact of the Covid-19 virus, which impacted people from various linguistic backgrounds globally. Although "pandemic" was first recorded in the Oxford English Dictionary in the early 17th century, it became a keyword in the twenty-first century.

As previously highlighted, the Covid-19 pandemic has led to significant translation challenges globally, largely due to the English-centric mass communication. It is essential to explore the lexicon associated with Covid-19 in order to facilitate the development of a translationally accurate Romanian rendition of English coronavirus-specific vocabulary, encompassing terms that are "dynamically equivalent"[6] in translation, as Nida (1964) states. The rationale for examining the vocabulary related to coronavirus translated from English to Romanian by Machine Translation stems from the insufficient linguistic resources available for adequately translating these terms.

The terms and phrases that have surfaced due to the Covid-19 outbreak are largely viewed as neologisms, marked by innovative word formation and borrowing. During the Covid-19 pandemic, several terms took on new meanings or gained prominence. For example, the terms "isolation" and "quarantine," which have long existed in medical and public health lexicons, were redefined with more specific implications related to the containment of the virus[7]. Similarly, the term "infodemic," a portmanteau of "information" and "epidemic," was initially coined during the 2003 SARS outbreak but became significantly more prominent during the Covid-19 crisis due to the overwhelming spread of misinformation. Moreover, abbreviations like "WFH" (work from home), originally introduced in 1995, saw a substantial increase in usage during the pandemic as remote work became widespread, according Merriam-Webster dictionary.

The emergence of new words and phrases during the Covid-19 pandemic can be seen as a form of lexical "enrichment," contributing to the dynamic nature of language. Happenbrouwers, Seuren, and Weijters (2019) conceptualize lexicons as repositories of meaning and usage, suggesting that they serve as reservoirs that add depth, nuance, and richness to a language's vocabulary. Also stressing the importance of lexical enrichment, researchers such as Krishnamurthy (2010) further elaborate on the various mechanisms through which neologisms, or new words, can originate. While borrowing from other languages is a common source of new vocabulary, neologisms can also arise through compounding, where two existing words are combined to create a new one (e.g., "lockdown"), blending, which involves merging parts of two words (e.g., "infodemic"), affixation, where prefixes or suffixes are added to existing words (e.g., "unprecedented"), coinage, which involves inventing entirely new words (e.g., "Covidiot"), and acronym formation, where initials of words are combined to form a new word (e.g., "PPE" for Personal Protective Equipment). As Al-Salman and Haider (2021: 24) note, "lexical innovations associated with health pandemics are common, as is the case with the abbreviations and/or acronyms for HIV *Human Immunodeficiency Virus*, SARS *Severe Acute Respiratory Syndrome*, MERS *Middle East Respiratory Syndrome*, among others. However, the creation and spread of the corona-inspired terms has been fast and commonly adopted globally." The Covid-19 pandemic has provided fertile ground for the creation of new words and phrases as people strive to articulate their experiences, emotions, and the unprecedented challenges posed by the crisis. These newly coined terms not only reflect the unique circumstances of the pandemic, but also contribute to the adaptability and resilience of language in responding to global events and social changes.

---

[6] Dynamic equivalence, as introduced by Nida (1964), refers to the process of translating in a way that prioritizes conveying the meaning and intent of the original text, rather than sticking to a word-for-word correspondence, ensuring the translation resonates naturally within the target language and culture.
[7] Merriam-Webster (2020): https://www.merriam-webster.com/dictionary/.

During the pandemic, there was has been an urgent need to address the conceptual and terminological gaps that emerged in various languages, aiming to provide accurate scientific information from the medical field to people across all social strata. Bernadette Paton, the executive editor of the current edition of the Oxford English Dictionary, highlighted that during the Covid-19 pandemic, it was "a rare experience for lexicographers to witness an exponential surge in the usage of a single word within a very short timeframe, with that word overwhelmingly dominating global discourse, often overshadowing all other topics."[8] The pandemic became a global issue, with news spreading as rapidly as the virus itself, and medical terminology and structures became commonplace in daily news and discussions.

Despite the fact that the new terms have been recognized as valid by prestigious dictionaries such as the OED (Oxford English Dictionary) and DOOM 3 (the 3[th] edition of Orthographic, Orthoepic, and Morphological Dictionary of the Romanian Language), the terminological inconsistencies emerged between English and other languages as Romanian. For example, Bowker (2020) discusses how even within the same language (French), terminological inconsistencies emerged between Canada and France, demonstrating how regional and cultural variation impacts translation accuracy. This discrepancy can be attributed to several factors. Firstly, Romanian, as a less widely spoken language compared to English, often lacks direct equivalents for newly coined terms, leading to a reliance on English borrowings or hybrid translations. Secondly, Google Translate may prioritize languages with larger datasets and user bases, meaning that Romanian translations might be less refined due to insufficient data. Lastly, the linguistic and cultural nuances of Romanian require careful adaptation of terms, especially those with specific connotations or that have rapidly evolved during the pandemic, such as "infodemic" or "WFH." These terms require professional expertise and contextual understanding, which automated systems often fail to capture effectively. Additionally, human-translated documents in this field, such as instructions and medical reports, lack reliable research support due to Covid-related emergencies, leaving insufficient time for professional translators to accurately translate these documents. These are two crucial aspects that form the foundation of parallel corpora used by Machine Translation engines.

---

[8]    https://www.theguardian.com/books/2020/apr/15/oxford-dictionary-revised-to-record-linguistic-impact-of-covid-19.

As English remains the primary language for disseminating and sharing information about the coronavirus, there is a significant need for accurate translation from English. Machine Translation has emerged as a crucial tool, demonstrating the progress and capabilities in the field of translation technology. In the context of the pandemic, one might assume that Google Translate, being one of the most popular and easily accessible MT services, was frequently utilized to facilitate access to information about the coronavirus.

A focus on the translation of Covid-related terms using Google Translate allows us to offer several relevant examples, which further underline the relevance of the analysis we have undertaken. For example, the term "rapid test," which is an accessible means for individuals suspected of having Covid-19 to determine their status, is described as a tool to ascertain whether a person is negative, positive, or reactive. The term "rapid" not only conveys the speed of the testing process, but also implies its biological, physical, and moral significance. The term "rapid test" was first attested in English in the 1980s. According to the OED, it was used in the context of medical testing to refer to diagnostic tests that provide results quickly, but the lexicographic entry for "rapid test" includes this early usage and details its evolution in medical and scientific contexts. The translation into Romanian from "rapid test" to "test rapid", term used during the Covid-19 pandemic, is an example of direct translation influenced by both linguistic and pragmatic considerations within the context of specialized terminology. In English, the adjective typically precedes the noun (e.g., "rapid test"), while in Romanian, the canonical structure often places the adjective after the noun (e.g., "test rapid"). This syntactic difference is a characteristic feature of Romance languages like Romanian, where the post-nominal position of adjectives is more common, especially for descriptive attributes. Thus, the translation adheres to Romanian syntactic norms, ensuring grammatical and semantic alignment. The method used here is a form of *calque* or loan translation, where the meaning of the original phrase is preserved, but the structure is adapted to fit the target language. During the pandemic, rapid diagnostic tools were widely discussed across different languages, necessitating a translation that was not only accurate but also immediately recognizable and functional within medical and public discourse. The Romanian "test rapid" mirrors the English term's brevity and clarity, crucial for effective communication in a health crisis. The retention of the adjective "rapid" (rather than substituting it with a more idiomatic Romanian term like "repede") reflects an effort to align with international medical terminology, promoting cross-linguistic consistency. From a translation studies perspective, this example highlights the dynamic relationship between fidelity to source-language structures and the adaptability required to meet target-language norms.

In order to provide more relevant examples, the terms "social distance" (with 107,000 occurrences on Google search) and "physical distance" (with 61,400 occurrences), were translated into Romanian as "distanțare socială" and "distanțare fizică". They are primarily registered in online and digital contexts such as news articles, health guidelines, and public discourse related to Covid-19. These terms gained prominence during the pandemic as measures to prevent the spread of the virus. For historical context, while the specific Romanian translations may not have been in use during the Spanish influenza outbreak of 1918, the concepts of maintaining distance to prevent disease spread were indeed relevant at that time. Osterholm et al. (2005) claim that the exact use of these terms or their direct predecessors in historical contexts would be documented in historical health records, medical literature, or archives related to the 1918 influenza pandemic.

Protopopescu (2023) emphasizes that despite Google Translate having advanced parameters and a complex database intended to enhance accuracy and fluency, Romanian, being a less common language with limited online data and lacking a real-time updated online dictionary, faces significant challenges regarding the appropriate lexis. These limitations contribute to the inconsistency noted in Google Translate's Romanian translations. Also, Protopopescu (2023: 70) underlines that "as far as Romanian is concerned, in spite of having a large number of parameters and a complex database which presumably optimizes the accuracy and the fluency of the Google Translate tool, it is one of the less common languages in the world, and the online data in Romanian are limited. Moreover, Romanian does not have an online dictionary that is updated online in real time, as is the case with say, the OED, and its one electronic corpus, CoRoLa, is still in its infancy stages. Therefore, the lack of input data for Romanian so far is clearly an overwhelming reason for the lack of consistency of the Google Translate tool."

The table below, Table (1), presents a comparative analysis of English Covid-19-related terms, their Google Translate equivalents, and the officially recognized Romanian terms, highlighting some discrepancies and errors.

*Table (1) Medical terms[9] used during Covid-19 pandemic translated by Google Translate in June 2020*

| English terms | Google Translation equivalent[10] | Romanian official term |
|---|---|---|
| Asymptomatic | Asimptomatică | Asimptomatic |
| Clinical trial | Studii clinic | Studii clinice |
| Corona/Coronavirus/Novel Coronavirus/Covid-19 | Corona/Coronavirus/Noul Coronavirus/Covid-19 | Corona/Coronavirus/Noul Coronavirus/Covid-19 |
| Community spread | Răspândire în comunitate | Răspândire în comunitate |
| Contagious | Contagios | Contagios |
| Epidemic | Epidemie | Epidemie |
| Flatten the Curve | Aplatizați curba | Aplatizarea curbei |
| Incubation period | Perioadă de incubație | Perioadă de incubație |
| Immunity | Imunitate | Imunitate |
| Isolation | **Izolație** | Izolare |
| Lockdown | **Carantină, blocare** | Stare de alertă |

---

[9] The terms that were selected are not exclusively related to Covid-19, but they have been used especially in this context. Some may have semantic extensions.
[10] Accessed in February 2022.

| Mask | Masca | Mască |
|---|---|---|
| Morbidity/Mortality | Morbiditate/Moralitate | Morbititate/Mortabilitate |
| Pandemic | **Pandemic** | Pandemie |
| Outbreak | **Epidemie** | Primele cazuri |
| Quarantine | Carantină | Carantină |
| Sanitizer | Dezinfectant | Dezinfectant |
| Spread | Răspândire | Răspândire |
| Self-quarantine | Carantină autoimpusă | Carantină autoimpusă |
| Social Distancing | Distanțare social | Distanțare social |
| Stay-at-Home<br><br>Shelter in place | A sta acasa | A sta acasa |
| Super-spreader | **Super distribuitor** | Persoană extrem de contagioasă |
| Symptomatic | Simptomatic | Simptomatic |
| Rapid test | Test rapid | Test rapid |
| Transmission | Transmitere | Transmitere |
| Vaccine | Vaccin | Vaccin |
| Ventilator | Ventilator | Ventilator |

While many terms were accurately rendered by Google Translate, several notable errors emerged. For example, the term with the highest frequency of search and use within the last two years is *Covid-19*. According to the WHO report from February 2020, the acronym of "coronavirus disease 2019" is supposed not to refer to any geographical location, animal, group of people, in order to avoid inaccuracy and stigmatizing. Despite the fact that it is considered a neologism, various forms of coronaviruses were discovered in the early 1930s known as *avian* *infectious bronchitis virus*, and in the mid-1960s human coronaviruses were identified that caused acute respiratory infections. It was registered for the first time by a dictionary such as Merriam Webster Dictionary, in 1968. According to Merriam Webster Dictionary (1968: 650), "the word was introduced by a group of virologists as a short article *Coronaviruses* in the *News and Views* section of *Nature*". The acronym kept the same form and meaning in all the languages and represented a bridge communication in the case of the pandemic for a better comprehensibility of the term.

Apart from Covid-19, there is a significant number of abbreviations that became frequent in English in the context of the pandemic. The use of abbreviations became pivotal during the pandemic for effective communication, particularly in scientific and public health contexts. Key abbreviations that emerged include SARS-CoV-2 (Severe Acute Respiratory Syndrome Coronavirus 2), the virus responsible for Covid-19, and PPE (Personal Protective Equipment), which encompassed items such as masks and gloves to prevent infection. Additionally, terms like RT-PCR (Reverse Transcription Polymerase Chain Reaction) described the primary diagnostic test for detecting viral RNA, while organizations such as the CDC (Centres for Disease Control and Prevention) and the WHO (World Health Organization) became central to managing the global response. These linguistic formations demonstrate the rapid evolution of language in response to global crises, serving as a tool for simplifying complex scientific information and enhancing public understanding. However, it is significant to already emphasize that there is a notable preference for English abbreviations in reporting the Covid-19 pandemic over their Romanian counterparts. Additionally, certain English terms represented by abbreviations or acronyms, such as PUI and WFH, lack direct equivalents in the Romanian language.[11]

The rise of the Covid-19 pandemic not only introduced new terminology, but also brought renewed focus to older terms, as these were rapidly adopted across various platforms such as the internet, social media, websites, official documents, and statements from authorities. Words and phrases like "self-isolation" and "solitary confinement," once more specialized or less commonly used, became widely circulated and integral to everyday conversations. Many of these terms gained widespread traction in digital and media spaces, reflecting their relevance to the global situation. Importantly, much of this evolving pandemic-related vocabulary was validated by health and communication experts, such as official governmental statements, items of information from the official website of World Health Organization, etc., ensuring consistency and accuracy in both public discourse and official communications. This process highlights how global crises not only reshape language but also lead to the institutional endorsement of terms that encapsulate the shared experience of the moment. In recognition of the linguistic impact that was caused by the pandemic in the English language, when specialised terms have become common, in April 2020, The Oxford English Dictionary included pandemic vocabulary. They also updated the original definition of Covid-19, as this term no longer defines as "an acute respiratory illness,", but "a disease (…) characterized mainly by fever and cough, (…) capable of progressing to pneumonia, respiratory and renal failure, blood coagulation abnormalities, and death."[12]

---

[11] The terms "pacient suspect" or "caz suspect" were officially used in Romania during the Covid-19 pandemic to describe PUI, as show some authoritative sources, such as the National Institute of Public Health, Ministry of Health, articles related to Covid-19, etc. Also, "work from home" is translated "munca de acasă" or "munca de la birou", terms that are used by official authorities and on websites such as www.digi24.ro.

To mark the lexical novelty in the Romanian language in the pandemic context, in 2021, the 3[rd] edition of the Romanian Dictionary of Orthography, Orthoepy, and Morphology (DOOM) was published in 2021, incorporating several new terms and abbreviations related to the Covid-19 pandemic. Among the additions were terms such as *asimptomatic* (asymptomatic), *a se autoizola* (to self-isolate), *a carantina* (to quarantine), *carantinar* (quarantine worker), *carantinare* (the act of quarantining), and *comorbiditate* (comorbidity). The dictionary (DOOM 3, 2021: 401) also defined *contact* as "a person in contact with a sick individual," reflecting pandemic-related terminology. Other newly included terms were *contagiozitate* (contagiousness), *coronasceptic* (someone skeptical of the coronavirus), *coronavirus*, *Covid*, *covidic* (Covid-related), *izoletă* (an isolation stretcher), *morbiditate* (morbidity), *rată record* (record rate), *vaccinologie* (vaccine science), and *ATI* and *ARN* (mRNA). The term *zoom* was also included, although no specific clarification was provided on whether it referred to the rapid increase in something or to the video conferencing platform that became widely used during the pandemic. The inclusion of Covid-related language in the official Romanian lexicon underscores the pandemic's significant influence on modern language and communication, further emphasizing how quickly new concepts become embedded in everyday vocabulary. In non-speaking English countries, translation has become an important tool of enabling access to Covid-19 information. Crowdsourced translations and volunteer translators play a critical but often overlooked role in disaster response. Zhang and Wu (2020: 517) highlight that the potential of these efforts in mitigating the effects of crises is both underestimated and

---

understudied, emphasizing the need for more attention to this area of research, by stating that "(…) crowdsourced translations and the capacity of volunteer translators in reducing the impact of disasters remain underestimated and therefore understudied."

Crowdsourced translations are certainly a means of overcoming linguistic barriers. Given the great need of reliable Covid-19 information and the pandemic that has been fuelled by what the WHO calls an *infodemic*, Translators without Borders (TWB) analysed the language and communication barriers that exist in the context of an epidemic crisis.[13] For example, in April 2017, Translators without Borders explored the accessibility and the efficiency of the communication among the refugee and migrant population of Greece. The findings, showed in Translation without borders (2017), showed that many refugees and migrants struggled to access critical information due to language barriers. It was further revealed that communication efforts were often not tailored to the linguistic needs of the population, which consisted of diverse language groups, and there was a reliance on English or Greek, languages that many migrants did not understand well. The study[14] thus highlighted the need for more culturally and linguistically appropriate communication strategies to ensure that crucial information regarding services, legal rights, and health care was effectively conveyed. While the findings underscored the importance of providing clear, multilingual information and the value of community-based interpreters to improve communication and aid access, it is essential to underline that there is no datum referring to the Romanian language in these services.

Apart from the study described above, several language studies have been conducted during the Covid-19 pandemic to examine the effectiveness of communication, particularly in ensuring that critical health information reaches diverse populations. One such study by Ferguson, Merga, and Winn (2021) investigated whether the written information from commonly accessed online sources about Covid-19 was presented at readability levels suitable for the general public. The authors' findings revealed that government departments responsible for public health information have failed to meet the goal of making their communications accessible to the majority of the population. Specifically, it was shown that the complexity of language used in public health materials often exceeded the recommended reading levels, creating a barrier for people with lower literacy skills. This issue is not confined to one country, as similar results were noted across Australia, the UK, and the US, suggesting a widespread challenge in ensuring that critical health information is understandable by all segments of the population. This presents significant concerns during a global health crisis where rapid and clear communication is vital for ensuring public safety. If sections of the population cannot comprehend the guidance provided, they may be unable to follow preventive measures such as wearing masks, practicing social distancing, or adhering to lockdown protocols. This, in turn, increases the risk of transmission and slows efforts to control the pandemic. While literacy is a key dimension regarding the availability of information, language barriers extend beyond readability and literacy challenges. For non-native speakers or multilingual populations, the lack of accurate translations and culturally appropriate communication exacerbates these issues.

---

[13] According to https://translatorswithoutborders.org/, Translators without Borders is a global community of over 80,000 translators and language specialists that offer language services to humanitarian and organizations worldwide.
[14] Translators Without Borders. (2017). *Language comprehension barriers in humanitarian settings*. Retrieved from https://translatorswithoutborders.org/wp-content/uploads/2017/07/Language-Comprehension-barriers.pdf.

In public emergencies such as the Covid-19 pandemic, language barriers can severely hinder the global dissemination of vital health information and compromise the clarity and effectiveness of preventive strategies. Misunderstandings or confusion stemming from poorly communicated guidelines can lead to non-compliance with health measures, thus aggravating the public health crisis. Therefore, ensuring multilingual communication strategies that prioritize clear, accessible, and culturally sensitive language is crucial. This requires interdisciplinary collaboration between public health officials, linguists, translation experts, and technology developers to ensure that information is not only translated but adapted to be comprehensible and relevant to diverse audiences. Without such efforts, the goal of inclusive public health communication cannot be fully achieved, leaving many at risk during global crises like the pandemic.

# 3. Research questions

In pursuit of a comprehensive analysis of lexical errors arising in the translation of Covid-19-related texts from English into Romanian by Google Translate, this study endeavours to explore the following research questions:

1. What are the prevalent types and patterns of lexical errors encountered in Covid19 texts translated from English into Romanian by Google Translate, and how do these errors impact the accuracy and fluency of the translated content?

2. How does the complexity and specificity of Covid-19 terminology influence the occurrence and severity of lexical errors in translations produced by Google Translate from English to Romanian, and what strategies can be developed to mitigate these errors effectively?

3. To what extent do linguistic and structural differences between English and Romanian contribute to the occurrence of lexical errors in translations of Covid-19 texts by Google Translate, and how can computational linguistics and machine learning techniques be leveraged to improve the quality and reliability of automated translations in this domain?

These research questions aim to explore the nature, causes, and potential solutions regarding lexical errors in the translation process from English to Romanian for Covid-19 texts using Google Translate. They provide a comprehensive framework for investigating various aspects of translation quality and effectiveness in the context of Machine Translation.

# 4. Statement of the problem

The data collection comprises terms related to Covid-19 that were selected, analysed and compared (see methodology stages in page 86). They were translated from English into Romanian by Google Translate, and the results were compared to the Romanian terms that were used by the authorities or that appeared in official public health documents. Moreover, the results from Google Translate were compared to the translation made by Chat GPT and Google Gemini (both large language models trained to generate and understand human language).

The main focus is on the English-Romanian pair (and not in the other direction or bidirectionally). The corpus contains terms used in the prospects of the vaccine, medical documents, and articles related to Covid-19 published by World Health Organisation on their official website (www.who.int) or Gavi Organization. The terms presented below were chosen according to their high occurrence in the texts that were selected to be translated. The benchmark of the parameters involved in the process of analysing includes relevance and diversity. Out of a variety of public health materials given by World Health Organisation, there were sampled the terms with high occurrences containing Covid-19 related content. The language pair makes the content relevant for the official health materials that were translated into Romanian and the terms that are used by the authorities into the target language. The translation benchmark was created from March 2020 to the present (the last dated article from the corpus).

# 5. Purpose of the study

The purpose of the research is to comprehensively investigate and analyse the lexical errors that occur during the process of translating Covid-19-related texts from English into Romanian using Google Translate. The primary objectives of the study are:

1. Identify and categorize the types of lexical errors that commonly occur in the Machine Translation of Covid-19 texts from English to Romanian, focusing on lexical errors at their interface with semantics and syntax, as well as on their interconnection to terminological and stylistic issues.

2. Examine the impact of lexical errors on the accuracy, fluency, and comprehensibility of the translated content, particularly in the context of conveying critical information related to public health, medical guidelines, and scientific research during the Covid-19 pandemic.

3. Investigate the underlying linguistic factors and computational mechanisms that contribute to the occurrence of lexical errors in Machine Translation systems like Google Translate when translating Covid-19 texts between English and Romanian languages.

4. Evaluate the effectiveness of existing Machine Translation evaluation metrics and methodologies in identifying and assessing lexical errors in Covid-19 translations, and propose novel approaches or enhancements tailored to the specific characteristics of CCovid-19-related discourse.

5. Explore potential strategies and techniques for mitigating and minimizing lexical errors in Machine Translation outputs for Covid-19 texts, including linguistic post-editing, domain-specific adaptation, and integration of human feedback loops into the translation process.

By addressing this research context, by examining the lexical translation errors, the performance of MT, the purpose is to contribute valuable insights to the fields of Machine Translation, computational linguistics, and translation studies, with practical implications for improving the quality and reliability of automated translation systems in conveying crucial information during public health emergencies such as the Covid-19 pandemic.

# Chapter II

# Error Analysis (EA)

# 1. Terminology: distinction between *error* and *mistake*

Error Analysis, as defined by Corder (1981: 45), is a significant branch of applied linguistics that focuses on understanding language acquisition and improving language learning. Corder (1981: 45) explains that Error Analysis serves two primary purposes: first, to examine the process by which individuals acquire either their native language or a foreign language, and second, to identify and address errors in a way that benefits both learners and teachers. By analysing these errors, educators can develop remedial strategies that facilitate learning, while also gaining insights into the language acquisition process from the learner's perspective. This dual role highlights the importance of Error Analysis in shaping effective language instruction and providing targeted interventions for improving language proficiency.

However, one of the practical justifications proposed for the study of the errors is meant to bring efficiency and progress in the field of translation. The field of Machine Translation between Romanian and English remains relatively underexplored, with only a limited number of researchers, such as Forăscu and Cristea (2005), focusing on the specific errors that occur in these language pairs. When analysing MT errors, two key perspectives are commonly used: the macro textual and micro textual approaches. According to Koponen (2010: 1), the macro textual perspective looks at the text as a whole, evaluating not just isolated translation segments, but the overall coherence and cohesion of the entire text. It ensures that the translation maintains consistency in meaning and structure across the broader context. On the other hand, the micro textual approach delves into the finer details of translation, focusing on lexical and morphosyntactic units, individual words, phrases, and grammatical structures. As noted by Secară (2005), the evaluation of human translation has evolved from traditional methods that emphasized word-level and sentence-level error analysis to more holistic macro textual methods. These newer approaches prioritize the function, purpose, and impact of the text, assessing how well the translation aligns with the original intent and communicates the message effectively to the audience. This shift underscores the importance of not only translating the text correctly at the micro level, but also ensuring that it reads naturally and coherently in the target language, thus achieving a balance between linguistic precision and communicative effectiveness.

There are various approaches regarding Error Analysis, and similar concepts that belong to the field of applied linguistics have been employed for it, such as contrastive analysis and interlanguage[15]. Researchers, such as Corder (1981), Cook (1991), Krashen (1975), Slama-Cazacu (1974), Selinker (1974), Doca (1981), and others have focused on defining and proposing studies, materials, and techniques, for the development and improvement of the error analysis in the field of linguistics. There are various studies about Machine Translation, made by Flanagan (1994), Ghasemi and Hashemiam (2016), Karami (2016), Koponen (2010), Popovič (2011), etc. Nevertheless, few works have dealt with studies where types of translation errors are examined, where explanation for the sources or the causes of errors are given, except for Hsu (2014) or Alam (2017). These researchers discussed the intricate relationship between translation and Error Analysis, emphasizing the significance of systematic evaluation in improving translation quality. Hsu (2014) highlights that Error Analysis in translation involves identifying, categorizing, and understanding errors in translated texts. This approach helps in diagnosing common pitfalls and challenges faced by translators. Alam's work (2017) underscored the significance of Error Analysis as a critical component of the translation process, offering valuable insights into how systematic error identification and correction can enhance translation quality and translator competence.

For a long time, the *error* was considered a negative aspect in the process of acquiring a language or in translation. Corder (1981: 1), who discussed this concept in the 1980s, states that "a good understanding of the nature of error is necessary before a systematic means of eradicating them could be found". An essential step in analysing errors is the distinction between *error* and *mistake*. Richards and Schmidt (2002: 195) define *error* as the use of a word, speech act or grammatical items in such a way it seems imperfect and significant of an incomplete learning. In the studies of Corder (1981: 10) related to foreign languages, the concept of *error* is considered to be the systematic deformation that appears in the process of a language acquisition or translation, whereas *mistake* is a non-systematic deviation that is made as a consequence of external or personal factors, such as lack of attention or motivation. The lexical errors that may be made in the process of translation from English into Romanian could be caused, according to previous researchers, such as Corder (1981), Selinker (1972), Richards and Schmidt (1985), by the asymmetry that exists between these two languages, the 'false friends', the various ways of translating lexical items, the absence of equivalent terminology, etc. Richards (1974), for instance, categorized competence errors into two main types: *interlingual errors*, which stem from interference from the learner's native language, and *intralingual* and *developmental errors*, which arise during the process of acquiring a second language before full knowledge has been attained. His focus was on the acquisition process rather than on Machine Translation, emphasizing how learners internalize linguistic structures over time rather than how automated systems process language.

---

[15] Even if an interlanguage is an idiolect developed by a learner of a second language, concepts analysed by researchers such as Selinker (1972), it manifests in translations as well, by preserving features of the source language in the translated text.

# 2. Lexical errors

Lexical errors in translation can stem from various sources, ranging from linguistic differences and cultural nuances to the dynamic nature of languages. According to Llach (2011), lexical errors occur when language learners or translators inaccurately use words due to deficiencies in lexical knowledge, leading to semantic inaccuracies or distortions in meaning. These errors can be classified into *formal errors* – where incorrect word forms are used – and *semantic errors* – which involve incorrect word choices that fail to align with the intended meaning.

A crucial subset of lexical errors involves lexical semantic choices, which pertain to the selection of words with inappropriate or unintended meanings within a given context. Homonyms and polysemous words, which carry multiple meanings, pose significant challenges for translators in identifying the most contextually appropriate term. As Llach (2011) highlights, semantic lexical errors arise when a word is misused due to confusion with a semantically related term, such as false cognates or near-synonyms. This issue is particularly prevalent in translation, where words may have subtle or context-dependent meanings that do not directly align between languages.

Furthermore, cultural idioms, colloquial expressions, and regional variations exacerbate the complexity of lexical semantic choices in translation. A direct word-for-word translation may fail to capture the intended meaning, leading to misunderstandings or loss of nuance. In this regard, Llach's (2011) classification of lexical errors provides a useful framework for understanding how semantic mismatches occur and how translators must navigate lexical ambiguity to preserve meaning and communicative intent. Llach's (2011) classification of lexical errors provides a useful framework for understanding how semantic mismatches occur and how translators must navigate lexical ambiguity to preserve meaning and communicative intent. By categorizing errors such as confusion of sense relations, use of false cognates, and misselection of near-synonyms, this framework highlights the complexities involved in lexical choice, especially in contexts where subtle nuances significantly alter interpretation. In the case of translation—particularly in specialized fields like medicine or public health—such insights are crucial, as even minor lexical inaccuracies can lead to miscommunication or the distortion of critical information. Llach's typology thus not only aids in error analysis but also serves as a guide for improving translator training and translation quality assurance.

Misinterpretations resulting from lexical inaccuracies may lead to confusion, misrepresentation in a sensitive, important and needed domains such medical translation in times of a pandemic. While technology, including Machine Translation, has made significant strides, it is not without its challenges in addressing lexical errors. Automated systems may struggle with idiomatic expressions, context-dependent meanings, and cultural subtleties. Continuous advancements in artificial intelligence and natural language processing are essential for refining Machine Translation capabilities and reducing lexical errors.

## 2.1. Lexical errors in translation

The motivations for choosing lexical errors, understood in terms of their complex, lexical role, for evaluating the process of automatic translation, are multiple. Firstly, current researchers, such as Alam (2017), Silalahi et al. (2018), state that the lexical errors made by Machine Translation have the highest frequency compared to morphological, syntactic or semantic errors. Secondly, lexical errors may affect the overall message of a text, by affecting the cohesion, as well as the coherence of the target text. This may easily lead to mistranslation, misunderstanding and, therefore, misinformation. It is generally assumed that lexis carries an important functional load for the message to be conveyed. In the process of translation, it is essential to prioritize the mapping of cohesion and coherence as guiding principles. These concepts are deeply interrelated and vital for producing accurate and effective translations. According to Halliday and Hasan (2014), cohesion involves the way in which surface elements of a text, such as lexical and grammatical components, are interconnected to create a sense of continuity. This includes how words and structures relate to each other within the text. On the other hand, coherence refers to the logical flow and consistency of meaning throughout the text, according to Maruashvili (2023: 45), ensuring that the ideas and arguments presented are clear and make sense as a whole. Therefore, understanding and applying both cohesion and coherence is crucial for maintaining the overall integrity and readability of the translated text.

Understanding the nature of translation errors is crucial for improving translation quality and effectiveness. Translation errors can arise from various deficiencies in the translation process, impacting the final output significantly. Pym (1992: 283) provides a comprehensive definition of the translation error, describing it as "a manifestation of a defect in any of the factors entering into (…) skills." This includes the translator's ability to generate multiple viable target texts from a single source text (e.g., target text 1, target text 2, target text n). According to Pym (1992: 283), these errors can stem from several causes, such as "lack of comprehension, inappropriateness to readership, [and] misuse of time." By identifying and addressing these factors, translators can enhance the accuracy and suitability of their translations for the intended audience.

Scholars such as Vinay and Darbelnet (1958: 38) offer a detailed framework for understanding translation techniques, particularly focusing on lexical errors. Newmark (1988: 42) further elaborates on translation strategies that can be applied to address semantic challenges, highlighting both literal and free translation methods as solutions to lexical errors. Similarly, Nord (1991: 72) presents a functionalist approach to translation that emphasizes the importance of text analysis to resolve semantic mismatches, emphasizing the role of source-text purpose and meaning in guiding error correction. These foundational works collectively offer both theoretical insights and practical approaches for addressing common errors encountered in translation.

The key techniques employed to enhance accuracy and clarity in translation, as outlined by scholars, include: literal translation (translating word-for-word while maintaining the original structure), equivalence (finding equivalent expressions or terms in the target language), and adaptation (modifying the text to be culturally relevant for the target audience). These strategies, as described by Vinay and Darbelnet (1958: 34), also include *transposition* (changing grammatical structures in the translation), *modulation* (altering the form to convey the same meaning from a different viewpoint), and *calque* (creating new expressions by literally translating components of a phrase). Additionally, *borrowing* (using the same word from the source language), *synonymy* (replacing a word with a synonym in the target language), *omission* (leaving out non-essential words or phrases), *addition* (providing extra clarifying information), and *contextualization* (considering the context to avoid lexical errors) are techniques extensively discussed by Newmark (1988: 45). These methods allow translators to navigate complex linguistic challenges and ensure that their translations are both accurate and culturally appropriate. Also, Nord (1991: 80) emphasizes the role of contextualization and functional equivalence in minimizing errors and ensuring effective communication.

## 2.2. Lexical errors in MT

Different translation techniques, such as expansion, adaptation, transposition, structure shift, are also used by Machine Translation. These techniques are integrated into the neural networks of modern MT systems like Google Translate. Wu et al. (2016) explain that they leverage large datasets and deep learning algorithms to apply these techniques effectively, ensuring high-quality translations across diverse languages and contexts in order to enhance the accuracy and readability of their output. According to Karami (2014), Google Translate is not only the most famous, but also one of the most widely used Machine Translation tools due to its accessibility and ease of use. Its popularity stems from its ability to handle a vast number of languages and provide instant translations, making it an essential tool for both casual users and professionals.

However, despite its widespread use and advancements, Karami (2014) notes that Google Translate still faces challenges, particularly with more complex texts, idiomatic expressions, and languages with less digital representation. These limitations highlight the need for continuous improvement and caution when relying on Machine Translation for critical or nuanced content. Nonetheless, it remains a go-to solution for quick, general translations in everyday contexts.

The recent Machine Translation used by Google Translate (based on an Artificial Neural Network and Deep Learning) has seen great improvement. As Benkova et al. (2021: 1) state, "it uses a deep neural network to process huge amounts of data, and is primarily dependent on training data, from which it learns. If there is a substantial dataset for training the model, then NMT can process any language pair, including languages that are difficult to understand. NMT allows the processing of text in various language styles, such as formal, medical, financial, and journalistic. Neural networks are flexible and can be adjusted to many needs". There are various translation error taxonomies that can be tailored for different languages and texts. Lommel et al. (2014: 14) presents the hierarchical translation error taxonomy, called MQM, by making a distinction between fluency errors and accuracy errors. Lommel et al. (2014: 24) explain *fluency errors* as those that are "related to the language of the translation, regardless of its status as a translation", and relate *accuracy errors* to "how well the content of the target text represents the content of the source." These distinctions provide a framework for assessing the severity of translation errors, determining whether they are minor issues or critical failures that significantly impact comprehension or usability.

It is important to underline that, as previous researchers such as Llach (2011: 72) have done, "identifying and isolating lexical errors is not always an easy task." While the definition of what a lexical error represents relies on that of "lexical competence", different researchers see it differently when it comes to defining lexical errors. While several previous researchers see lexical errors as a broad category, in opposition to grammar errors, others provide more narrow definitions of what lexical errors may represent. For example, there have been previous researchers that mark a sharp distinction between lexical and semantic errors. According to Costa et al. (2015), semantic errors manifest in various forms, encompassing sense confusion, erroneous choices, collocational inaccuracies, and issues with idioms. Sense confusion occurs when a term in the source text is translated into a representation of one of its potential meanings in the target text, requiring correction. Wrong choices entail the mistranslation of a source word into an unrelated target word without establishing a clear semantic connection. Collocational errors involve the misuse of natural word combinations. Idiom errors arise when translators, unfamiliar with the source text's idiomatic expressions, render them literally. According to Costa (2015), semantic errors involve problems with the overall meaning, while lexical errors specifically focus on issues related to individual words and vocabulary.

The diagram below (see Figure 2) presents a hierarchical classification of error types that can occur in language processing, specifically in translation or language generation tasks. Costa et al. (2015) categorize errors into five main types: orthography, lexis, grammar errors, semantic (errors related to meaning), and discourse errors.

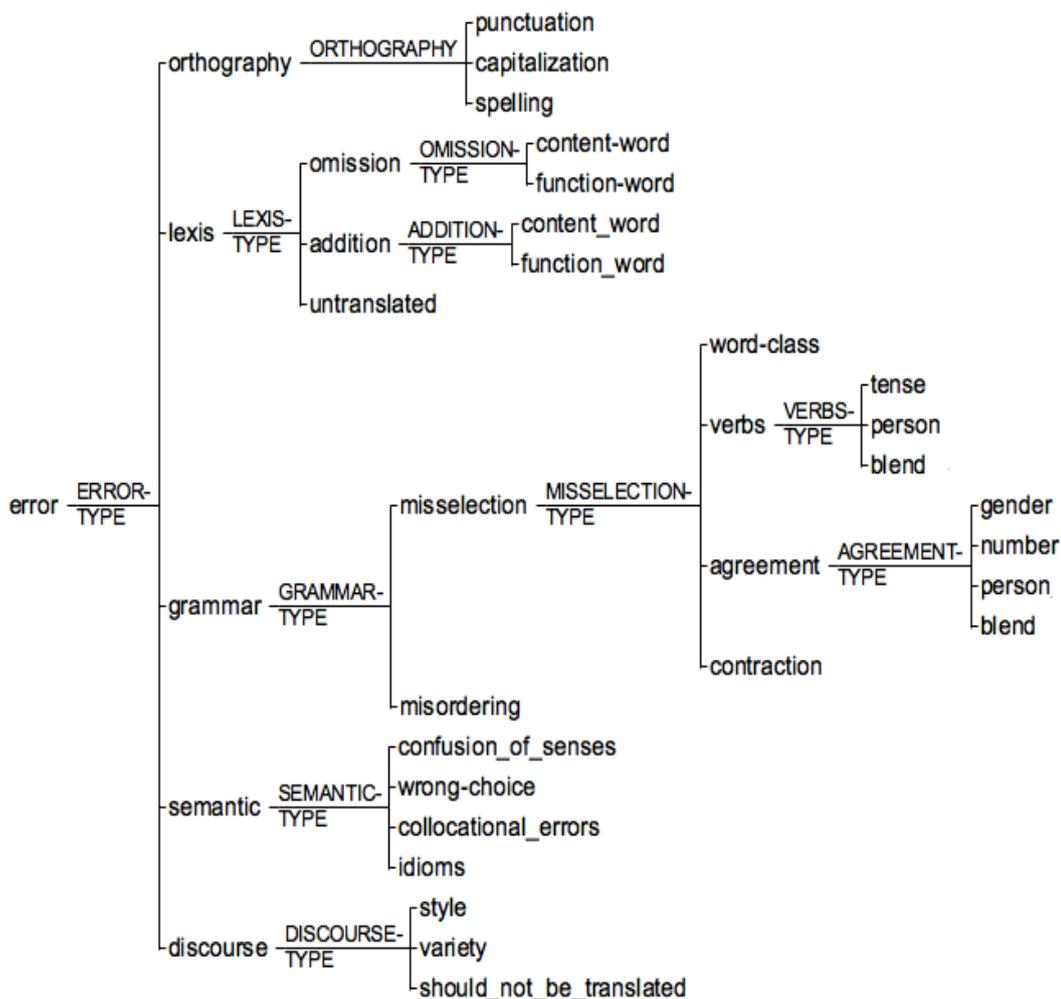

Figure 2. The taxonomy of errors (Costa et al., 2015)

MT systems, including Google Translate, often grapple with the distinction between lexical and semantic errors in order to correctly identify the inaccuracies that can significantly affect the accuracy of translations. According to previous approaches on Machine Translation, Badilla and Núñez (2020: 180) claim that errors which can be categorized as lexical, if they refer to a wrong choice of content words that express meaning in communication either oral or written that causes misunderstandings related to the outcomes or messages that speaker or writer wish to convey.

Nevertheless, other approaches avoid labels such as *lexical* or *semantic* altogether in their attempt to categorize errors in Machine Translation. A slightly different type of marking the errors in this domain was approached by Carl and Bàez (2019) for Chinese and Spanish error annotations for a MT output:

|  | Fluency |
| Mistranslation | Cohesion |
| Omission/Addition | Word Form |
| Unintelligible | Word Order |
|  | Punctuation |
|  | Spelling |

<div align="right">Figure 3. Classification scheme</div>

Popovič and Burchardt (2011: 84) suggest the following classification scheme both for human and automatic output: *inflectional errors* (if the full form of the word is not appropriate), *reordering errors* (for the words that are considered a WER error), *missing words*, *extra words*, *incorrect lexical choice*. Vilar et al. (2006: 698) states that there are four categories of errors: missing words (if the words are missing in the target text), word order (the word in the target text is wrongly positioned), incorrect words (as a result of a confusion it is used an incorrect word), unknown words (the words that are not translated at all) and punctuation errors (the incorrect usage of punctuation).

The errors related spelling and writing, meaning to orthography, include systematic deformations in punctuation, capitalization, and spelling. These are surface-level issues that can affect the readability of a text but typically do not alter the meaning. The lexical or word-choice errors, meaning related to lexis, are presented like this: the lexical errors are divided into omission (failure to include a word) and addition (inclusion of unnecessary words), both of which can involve content words or function words, and there is also the untranslated category, where words or phrases remain untranslated, affecting the clarity and understanding of the target language text. The grammar errors involve misselection (choosing the wrong grammatical form, such as the incorrect word class or verb tense) and agreement errors, such as problems with subject-verb agreement in gender, number, person, or tense. As for the contraction errors, they involve the improper use of shortened grammatical forms, also fall into this category.

Furthermore, the semantic errors entail a deeper misunderstanding of meaning, such as misordering (incorrect word order), confusion of senses, wrong choice of words, and collocational errors (incorrect combinations of words). Misinterpreting idioms and style errors are also classified here. At the discourse level, errors may occur in broader textual or conversational contexts, affecting style, variety, or the misuse of content that should not have been translated. These errors are less about individual word choices and more about how ideas are expressed across sentences or paragraphs.

This error classification is detailed and allows for a systematic approach to identifying and addressing translation or linguistic errors. Each category drills down into specific subtypes, reflecting the complexity of translating or generating accurate text, especially in Machine Translation systems. For example, the misordering and wrong-choice subcategories in the semantic errors section show how challenging it can be to preserve the correct meaning in translation. The agreement errors in grammar highlight the linguistic challenges in languages with gender and number agreement rules, which are often a problem for non-native speakers and Machine Translation algorithms. This diagram would be particularly useful in analyzing translation systems like Google Translate, allowing researchers to evaluate the type and frequency of specific errors. Understanding this taxonomy helps in refining translation systems and improving linguistic accuracy, especially in technical or specialized fields such as medical translation or legal documentation.

Hemchua and Schmitt (2007) discuss the classification of lexical errors in language learning research, emphasizing the importance of understanding the different types of lexical mistakes learners make in order to improve vocabulary instruction and error correction strategies. They categorize lexical errors into formal errors and semantic errors. However, the use of compact classification systems may lead to unclear boundaries and arbitrary classifications due to the complex nature of language. Also, Lommel et al. (2014) argue that the challenges of translation error categorization stem from the fluid and overlapping nature of linguistic structures.

Despite the ongoing challenge of overlap between error categories, a more comprehensive framework can enhance the identification and discussion of error types. The expanded framework for lexical error classification draws heavily from James's (1998: 144) taxonomy, which encompasses formal and semantic features. James categorizes lexical errors into formal misselection, misformations, and distortions. Formal misselection involves errors such as suffix confusions (for example, the word *medication* is sometimes erroneously translated into Romanian by *medicație*, and the correct Romanian translation is *medicamentație* or *medicament*), prefix confusion, vowel-based, and consonant-based confusions, as well as false friends caused by semantic divergence (see more examples in Chapter IV). Misformation occurs when learners use words from their native language in the target language without modification. Distortions result from the misapplication of the target language's rules. In addition to formal errors, collocation errors are addressed, focusing on the inappropriate use of word combinations. The misuse of collocation can range from semantically determined word selection to statistically weighted preferences and arbitrary combinations.

Taking as a point of reference a broad yet systematic definition of the lexical error as a superordinate term that includes several other subclasses of errors, which include those based on descriptive or semantic criteria, the present study focuses exclusively on lexical errors, which will be detected and analysed, and classified according to a scheme derived from previous analyses. The error classification scheme to be used in this research derives from the models proposed by Flanagan (1994), Hsu (2014), and Hemchua and Schmitt (2007). The error classification schemes proposed by the researchers mentioned above offer distinct yet complementary perspectives on identifying and analysing translation errors. A detailed comparison of these schemes reveals differences in focus, categorization, and application, each tailored to their specific contexts – Machine Translation, human translation, and non-native language use, respectively. Flanagan's classification scheme is primarily concerned with errors in Machine Translation. This framework emphasizes linguistic accuracy and the technical challenges inherent in automated translation processes. The categories identified by Flanagan include lexical errors (these involve incorrect word choices, often due to polysemy or synonym confusion, errors here include misselection and mistranslation), syntactic errors (these errors occur when the MT system fails to maintain proper word order or grammatical agreement, leading to syntactically incorrect sentences), semantic errors (these errors arise when the translation fails to convey the correct meaning, which may be due to cultural differences or idiomatic expressions not being properly translated), orthographic errors (these involve spelling and punctuation mistakes, which can impact the readability and comprehension of the translated text). Hsu's classification scheme focuses on errors in human translation, particularly within an educational context. It underscores the need for a comprehensive understanding of errors to improve translation pedagogy. Hsu categorizes errors into grammatical errors (morphological and syntactic), lexical errors (selection, formation), semantic errors (literal translation, misinterpretation), pragmatic errors (register, cultural). Hemchua and Schmitt's (2007) scheme focuses on lexical errors made by non-native English speakers in written English, providing insights into common lexical challenges faced by language learners. Their classification includes formal errors (spelling, word formation), semantic errors (confusion of sense relation, collocational errors), pragmatic errors (contextual misuse, register), transfer errors (interlingual transfer, intralingual transfer). While each classification scheme serves a different primary audience – Machine Translation developers, translation educators, and language teachers respectively – they collectively contribute to a comprehensive understanding of translation errors. By integrating insights from all three, one can develop more robust strategies for addressing linguistic inaccuracies in both human and Machine Translation contexts.

The explanations and criteria for each category of translation errors are primarily based on two key factors: accuracy and fluency. When analysing lexical errors in Machine Translation, focusing on both accuracy and fluency is crucial for improving the overall quality of translations, enhancing user experience, and advancing MT technology. Accuracy ensures that the translated text conveys the exact meaning of the original, while fluency guarantees that the text reads naturally in the target language. However, beyond these two criteria, ensuring coherence and cohesion in the translation is equally important. When lexical errors disrupt the accuracy of individual words or phrases, the overall coherence of the text can be compromised, leading to miscommunication or a fragmented translation. Similarly, fluency issues, such as awkward sentence structures, can affect the cohesion of the translation, making the text disjointed and difficult to follow. By addressing both lexical errors and maintaining accuracy, fluency, coherence, and cohesion, MT systems can produce more reliable, effective translations that are contextually and culturally appropriate. Lexical errors can lead to significant misunderstandings, especially in critical fields such as medical, legal, and technical translations. A fluent translation reads naturally and smoothly in the target language. Lexical errors often result in awkward or unnatural sentences that disrupt the flow, making the text difficult to read and understand. Analysing these errors helps improve the naturalness and readability of translations, making them more user-friendly. Effective translation requires a deep understanding of the context in which words and phrases are used, as context plays a crucial role in ensuring both coherence and cohesion in the translated text. Thus, understanding context is not only about accurately translating words and phrases, but also about ensuring that the translated text remains coherent and cohesive, reflecting the original intent and structure while flowing naturally in the target language. By studying these errors, MT systems can be optimized to handle cultural nuances and idiomatic expressions better, facilitating smoother communication between speakers of different languages. It also includes the most popular terms that have the highest frequency in the pandemic context. According to the way how MT use has been developed, explained by Keh-Yih and Jing-Shin (1992), and the survey done spreading the information hours on https://surveys.google.com/, we illustrated some of the most popular MT systems that are used nowadays.

Semantic errors involve inaccuracies or misunderstandings related to meaning. These errors occur when there is a divergence in the intended meaning of a source text and the conveyed meaning in the translation. They encompass a wide range of issues, including misinterpretation of concepts, failure to capture cultural nuances, or the use of words with different connotations.

For this study, according to the classifications displayed above and the frequency of the errors identified in the data, the taxonomy of lexical errors will be the following:

*Table (2) The classification of the lexical errors*

---

Formal errors

- Abbreviations and Acronyms
- False friends
- Borrowings
- Distortions

Semantic errors

- Confusion of sense relations
- Collocation errors
- Arbitrary combination
- Preposition partners

Stylistic errors

---

The classification of errors chosen is inspired by the works of several researchers who have studied language and translation errors, particularly focusing on lexical and semantic issues. In constructing the classification of translation errors from above, we drew selectively from the established frameworks proposed by Flanagan (1994), Hsu (2014), and Hemchua and Schmitt (2007), aligning specific elements of their taxonomies to the needs of our study. From Flanagan, whose model is rooted in the analysis of Machine Translation output, we adopted the distinction between lexical and semantic errors, particularly the emphasis on mistranslation and misselection arising from polysemy and synonym confusion. Hsu's model, focused on human translation within educational settings, informed our inclusion of pragmatic and cultural aspects within semantic misinterpretations, especially regarding register and literalism. Hemchua and Schmitt's detailed breakdown of lexical errors among non-native English users provided the basis for our subcategories, such as false friends, borrowings, and confusions of sense relation, as well as transfer-related distortions. Synthesizing these perspectives, we structured our own classification into formal, semantic, and stylistic errors, thereby capturing a broad spectrum of error types relevant to both human and machine translation, while emphasizing their impact on accuracy, fluency, coherence, and cultural appropriateness.

Error analysis could be instrumental in improving the performance of Google Translate or Machine Translation systems in general, especially in analysing lexical errors in texts related to Covid-19. Error analysis allows linguists and researchers to identify recurring patterns of errors made by Machine Translation systems. By systematically analysing the lexical errors encountered in Covid-19-related texts, researchers can pinpoint specific linguistic challenges and areas where Machine Translation algorithms tend to falter. Covid-19-related texts often contain specialized terminology, technical jargon, and nuanced language usage specific to the field of medicine, epidemiology, and public health. Analysing lexical errors provides insight into the linguistic complexity of such texts and highlights the challenges faced by Machine Translation systems in accurately translating them. Effective translation requires an understanding of the context in which words and phrases are used. Error analysis helps reveal instances where Machine Translation systems fail to grasp the contextual nuances of Covid-19-related texts, leading to inaccuracies or mistranslations. Understanding these nuances is crucial for improving the overall quality and accuracy of Machine Translation outputs. Error analysis serves as valuable feedback for developers and engineers working on Machine Translation systems like Google Translate. By identifying specific lexical errors and the underlying reasons behind them, developers can refine algorithms, update dictionaries, and incorporate domain-specific knowledge to enhance the performance of Machine Translation systems, particularly in handling Covid-19-related content. Through error analysis, linguists can propose tailored solutions and strategies to address the unique challenges posed by Covid-19-related texts. This may involve developing specialized translation models, integrating context-aware algorithms, or implementing post-editing techniques to improve the accuracy and fluency of translations in this domain. In conclusion, error analysis plays a crucial role in the continuous improvement of Machine Translation systems like Google Translate, particularly in handling complex and specialized texts related to Covid-19.

In summary, we chose to focus on lexical errors because studies suggest that these errors are the most frequent when two languages – in this case, Romanian and English – interfere. By examining these errors, we can better understand the difficulties that occur during the process of translating between Romanian and English, and work toward improving translation quality and language comprehension. Also, the erroneous lexical choice affects the message of a text. This perspective aligns with the views of renowned translation theorists such as Nida (1964), who emphasized the importance of semantic accuracy in his dynamic equivalence theory. Similarly, the communicative translation theory of Newmark (1988) highlights the crucial role of appropriate lexical choices to preserve the intended meaning. Baker (2018) also discusses the significance of lexis in her works, outlining the challenges and strategies in dealing with non-equivalence at the word level. Furthermore, Venuti's (1995: 20) concepts of foreignization and domestication indirectly underscore the impact of lexical choices on the reception and interpretation of a translated text. These scholars collectively underscore the critical nature of lexical accuracy in maintaining the integrity of the translated message. Accurate translation of Covid-19-related texts is crucial for disseminating vital information, ensuring public safety, and promoting global health literacy. Lexical errors in translations can distort meanings, mislead readers, and hinder comprehension, potentially compromising the effectiveness of communication efforts. Analysing lexical errors helps identify specific linguistic challenges and areas of difficulty in Machine Translation, particularly in highly specialized and technical domains like medicine and epidemiology. This insight can inform the development of more robust translation algorithms and tools. Covid-19-related texts often contain terminology, concepts, and cultural references that may vary across languages and cultures; by identifying recurrent lexical errors and patterns in machine-translated texts, researchers can provide valuable feedback to developers and improve the performance of translation systems like Google Translate.

# Chapter III

# Machine Translation

## 1. Overview of MT

Along the years, linguists and software engineers have tried to develop and improve a computer system for the translation of natural languages. Translation is a major tool in communication and an informational spreader that, in association with the internet and technology in general, can bridge the gap between different nationalities and influence human life. Previous researchers have already emphasized upon the significance of translation, as is underlined by Baker (2018: 10) who states that "Cultural and linguistic transfer, which is at the heart of the translation process, plays a vital role in facilitating intercultural communication. Translators are mediators between languages and cultures, and their task is to enable meaningful communication across linguistic boundaries. However, the rise of Machine Translation systems has brought about significant changes in the translation landscape. Machine translation, while offering speed and convenience, poses challenges in terms of accuracy, cultural nuances, and the preservation of the translator's role as an intercultural mediator", according to Baker (2018: 25). Not only during the pandemic era, but also in our everyday lives, technology has become an integral part of various activities, significantly influencing how we communicate, work, and entertain ourselves. Its role has expanded rapidly, transforming sectors such as healthcare, education, and commerce. The advancement of translation technology, for instance, has made it easier for people from different linguistic backgrounds to interact seamlessly. Tools like real-time translation apps, Machine Translation services, and AI-powered language platforms have revolutionized the way we understand and convey messages across languages, enhancing global connectivity and collaboration. As technology continues to evolve, its impact on our daily routines and its potential to bridge cultural and linguistic gaps will only become more pronounced.

Machine Translation, as is defined by OED, is "the process of translating language by computer"[16]. According to Nida and Taber (1974: 123), MT can be defined as "the use of automated computer systems to translate texts from one language to another, with minimal or no human intervention, based on linguistic and computational algorithms". A widely recognized definition of Machine Translation is provided by the American Translators Association (ATA), which states: "Machine translation is the translation of a text from one language into another language by a computer without any human involvement during processing."[17] This definition emphasizes the automated nature of Machine Translation, highlighting that the process is conducted entirely by computer systems without human intervention.

---

[16] https://www.oxfordlearnersdictionaries.com/definition/english/machine-translation? [Accessed on the 21st of March 2023].

In comparison with human translation, Machine Translation is often viewed as a process that involves transferring the meaning of a text from one language to another, without fully capturing the holistic interpretation of the context, cultural nuances, and emotional elements, which are typically found in human translation. Several researchers have examined this comparison in depth. For instance, Pym (2011: 115) has explored the limitations and potentials of Machine Translation, emphasizing the need for human oversight to ensure cultural and contextual accuracy. Lavie et al. (2004) and Zhang et al. (2004) have conducted extensive research on statistical and neural Machine Translation, analysing how these technologies perform in comparison to human translators. Additionally, Way et al. (2020: 78) has contributed significantly to the field by assessing the quality of Machine Translation outputs and their implications for professional translation practice. Furthermore, House (2015: 122) has critically evaluated Machine Translation from a functionalist perspective, highlighting how machine-generated translations often lack the pragmatic and stylistic nuances that human translators naturally incorporate. These scholars collectively underscore the challenges and advancements in the field of Machine Translation, drawing important distinctions between machine-generated and human-generated translations.

Machine Translation is now able to provide not only translations of words, but also of phrases or sections of texts. It can translate by carrying out the adjustments necessary in order to give a "natural" translation. This concept of "natural translation" was initially defined by Harris (1977: 102) as "the translation done by bilinguals in everyday circumstances without special training for it." The importance of natural translation was emphasized later on in translatology. In comparison with human translation, Machine Translation is often viewed as a process that involves transferring the meaning of a text from one language to another, without fully capturing the holistic interpretation of the context, cultural nuances, and emotional elements, which are typically found in human translation. For instance, Karami (2014) is an advocate of this concept in the approach of Machine Translation. He claims that the best appropriate translation pattern is given by SMT (Statistical Machine Translation). This is a Machine Translation paradigm whose parameters are bilingual text corpora, which is why the quality of its services depends on human-translated texts. Karami (2014: 3) asserts that "SMT translator gives more natural translations. The more bilingual corpus it has, the more translators trained with new bilingual corpus, the more natural translation it has." These scholars collectively underscore the challenges and advancements in the field of Machine Translation, drawing important distinctions between machine-generated and human-generated translations

---

[17] www.atanet.org.

## 1.1. History of Machine Translation

Machine translation or automated translation represents a branch of computational linguistics that has existed since 1933. Pestov (2018) and Hutchins (2000) identify 1933, as the year of the official starting point of Machine Translation 's development. marked by the issuance of patents. According to Hutchins (2000: 15), George Artsrouni received a patent in France, while Petr Trojanskij patented a similar idea in Russia. These developments are often regarded as some of the earliest formal steps in the history of Machine Translation.

The history of the MT outputs has been briefly presented by Ghasemi and Hashemian (2016) who state that the first system adopted by Google Translate was rule-based. According to Hutchins (2005: 13), Rule-Based Machine Translation (RbMT) systems operate based on a comprehensive set of linguistic rules manually developed to map structures from the source language to the target language. These systems rely on extensive lexicons that include grammatical, lexical, and semantic details, ensuring structured translation outputs. A more recent system is Statistical Machine Translation (SMT) that uses probabilistic models to generate translations by analysing large multilingual corpora. According to Wang et al. (2017), SMT relies on statistical models that are trained on large bilingual corpora to generate translations. These systems, according to Koehn (2010: 45), rely on algorithms to detect patterns in pre-translated texts and assemble smaller units of language into coherent sentences, prioritizing the most statistically probable translations. However, it often produces approximate or imprecise translations, particularly for Romanian, which, as a minor language, lacks the robust database available for major languages, as Somers (2003: 117) claims.

According to Hutchins (2000: 5), the first versions of MT were mainly mechanical dictionaries.[18] Research projects based on this work were conducted from that year until 1954 when IBM officially made the demonstration. It took place on January 7th in 1954, within a political context, namely the Cold War. The experiment yielded positive results, although certain details were not disclosed. It was said that the examples that were translated were intentionally selected in order to avoid ambiguity. The vocabulary was limited to 250 words and six grammatical rules. In a report presented to the US ALPAC committee, MT was described as "expensive, inaccurate, and unpromising," as Pestov (2018) notes. When the US government inquired about the quality of MT at the end of the 1950s, Yehoshua Bar-Hillel (1965: 174-179) provided an illustrative example: in the sentence "Little John was looking for his toy box. Finally, he found it. The boy was in the pen." A human translator, without hesitation, would correctly interpret "the pen" as "the container" rather than "the writing instrument.". Therefore, Bar-Hillel (1965) suggested that a "universal encyclopaedia" was needed so that a machine could deal with "real world" structures. He influenced other researchers at the time, but later on the systems that were developed on knowledge proved the opposite.

In the following decade, two distinct approaches emerged in the field of translation: the practical approach, focused on error analysis, and the theoretical approach, which consisted of research projects laying the groundwork for computational linguistics. A seminal work that introduced the concept of Machine Translation was Warren Weaver's paper titled "Translation," published in 1949. Weaver explores the challenges and possibilities of automating the translation process using computational methods, breaking down the translation process into three main stages: analysis, transfer, and synthesis. He emphasizes the importance of understanding the structure and meaning of both the source and target languages to achieve accurate translation. Weaver (1955: 15) acknowledges the complexity of language and the inherent difficulties in capturing the nuances of meaning and context. He highlights the need for a multidisciplinary approach, combining insights from linguistics, mathematics, and computer science to develop effective Machine Translation systems. Although Weaver's paper does not specifically focus on error analysis, it lays the foundation for subsequent research in the field. His visionary perspective on the potential of Machine Translation set the stage for future advancements in the automation of language translation. Additionally, the practical approach to error analysis gained traction through the work of researchers such as Catford (1965) and Pym (1992). Catford's "A Linguistic Theory of Translation" (1965) delves into the linguistic dimensions of translation errors, while Pym's various works emphasize the importance of error analysis in improving translation accuracy. Both researchers contributed significantly to understanding and mitigating translation errors, thereby enhancing the practical applications of Machine Translation systems. These complementary approaches – the theoretical foundations laid by Weaver and the practical insights from error analysis by Catford (1965) and Pym (2011) – collectively advanced the field of Machine Translation.

---

[18] This idea is attributed to Warren Weaver, an American scientist and mathematician. Weaver made significant contributions to the field of Machine Translation and is known for his influential report titled "Translation" published in 1949. In this report, he described the early attempts at Machine Translation and referred to them as "mechanical dictionaries" or "dictionary-based" approaches. Weaver's report provided a framework for subsequent research and development in Machine Translation.

Later on, around the 1980s, according to Hutchins (2003), MT approached "the direct translation model". This approach implies that the MT system is designed to translate directly from the source language (SL) into the target language (TL) with the use of a large bilingual dictionary, without analysing the vocabulary and the syntax. The "direct translation model" is an approach in Machine Translation that aims to achieve translation by directly mapping words or phrases from the source language to the target language. This approach assumes a one-to-one correspondence between words or phrases in the source and target languages, without considering the broader syntactic or semantic structure. In the direct translation model, the translation process primarily focuses on identifying and matching individual words or short phrases between the source and target languages. The direct translation model, which relies heavily on bilingual dictionaries or word alignment techniques to establish translations, has been extensively discussed in the literature on Machine Translation. Authors such as Hutchins and Somers (1992), plus Hutchins (2003) have highlighted its role in the early stages of Machine Translation research. They note that while the direct translation model was pivotal for initial explorations, it faced significant limitations, including difficulties in capturing idiomatic expressions, word order variations, and cultural nuances. This often resulted in translations lacking fluency and naturalness.

The early limitations of Machine Translation have been further elaborated by researchers like Melby (1981), who pointed out that the direct translation model struggled to handle the complexities inherent in human language. As Melby discusses, the model's reliance on direct word-for-word translation could not adequately address syntactic and semantic variations between languages. Despite its limitations, the direct translation model was a crucial stepping stone in the evolution of Machine Translation. Having been introduced in the early 1990s and replacing earlier models, Statistical Machine Translation (SMT) approaches, as noted by Koehn (2010) in "Statistical Machine Translation," leveraged probabilistic models to better handle linguistic variability and context. More recently, the advent of Neural Machine Translation (NMT) has further revolutionized the field. Vaswani et al. (2017) highlight how transformer models have enhanced the ability of NMT systems to produce more natural and fluent translations. Overall, the direct translation model's initial contributions set the stage for these sophisticated approaches, which continue to evolve and refine the capabilities of Machine Translation systems.

A less empirical approach was the interlingua model one. This approach was based on codes and symbols and consisted of a two-stage process of translation: from the source language into interlingua and from interlingua into the target language. Paradoxically, behind its simplicity, the interlingua represents an artificial language, a set of grammatical, semantical, lexical systems that are common between the languages involved in the process. According to Nirenburg (1987) and Somers (1992), the interlingua model is an approach to Machine Translation that aims to establish a common intermediate language representation, or interlingua, to facilitate translation between multiple languages. Ceccato (1961) believed that the key to achieving successful Machine Translation lies in finding a universal underlying representation that captures the meaning and structure of sentences in a language-independent way. Hutchins and Somers (1992) discuss the interlingua model in the context of Machine Translation and its advantages over direct translation approaches. They provide an overview of various Machine Translation models in order to facilitate more accurate and efficient translation by providing a unified framework for understanding sentence meanings. By utilizing the interlingua model, Ceccato (1961) aimed to overcome the limitations of direct translation approaches that rely on word-to-word mappings or rule-based transformations specific to language pairs. The interlingua provides a unified framework for understanding the meaning of sentences and enables more effective translation across different languages. Ceccato's work on the interlingua model contributed to the exploration of language-independent representations and the progress of rule-based Machine Translation systems. While the interlingua model had its own challenges and limitations, it laid the foundation for subsequent research and inspired further advancements in the field of Machine Translation. Among various projects and researches, there was a project called 'COMIT' realized by Yngve's (1960) team from Massachusetts Institute of Technology (MIT) that worked on improving paper tapes, access speed and other factors of the hardware. The project took place in the late 1950s and early 1960s.

The period of time when the interlingua approach gained significant attention was between the 1960s and the 1980s. Researchers aimed to develop a language-independent representation that could serve as a bridge for translating between multiple languages. Later on, between the 1980s and the 1990s, the interest in interlingua continued, but began to be overshadowed by the rise of statistical methods. However, some research into interlingua-based systems, such as Carbonell and Tomita (1987) and Hutchins (2005) persisted, particularly in academic and theoretical contexts. From the 1950s to present, from the early days of Machine Translation, post-editing has been a practical approach. In the 1950s and 1960s, post-editing was necessary to correct the rudimentary output of early MT systems. From the 1990s to 2000s, post-editing became more formalized and essential. This period witnessed a growing interest in computational linguistics and the application of computational methods to analyse and understand language.

An important point of reference in the history of Machine Translation was in the year 1966 due to the ALPAC report. According to Chomsky (1965), the ALPAC symbolizes *The Automatic Language Processing Advisory Committee*. It was represented by seven scientists supported by the US government that, eventually, came to the conclusion that automatic translation was less advantageous compared to human translation in terms of expenses, accuracy and even speed. Some notable individuals were involved in the ALPAC committee. Their expertise and contributions helped evaluate the state of the field and provided valuable insights into the challenges and potential of automatic language processing, including Machine Translation. They adopted the idea that the MT tools should be developed so that automatic dictionaries could help translators. In the 1970s, the Systran system was created and it was founded by Peter Toma. According to Petrits (2001), Systran was installed for the United States Air Force, then for the Commission of the European Communities and it provided the technology for Yahoo! Babel Fish and Google Language Tools in its early stages. The system used ruled-based Machine Translation and later a hybrid system with the Statistical Machine. Petrits (2001: 9) claims that nowadays it has become a complementary tool for human translators: "After eighteen years of marginalised but continuous development, SYSTRAN has now become an effective answer to some of the pragmatic translation needs in the operational departments of the Commission. In its early days it was not welcomed with open arms by the translators, as it was seen as an unfair competitor. However, the computerisation of the Commission services together with a general change of attitude, has radically altered the image of SYSTRAN. Now that it is available through the network to all Commission staff, it is no longer considered a substitute for human translation but as a product which complements the service translators provide."

In the later part of the 1980s, as Hutchins (2003) mentions, many projects followed, and some of those systems are still in use. There were the TAUM project, CETA, EUROTRA, ATAMIRI, ROSETTA (in the Netherlands), DLT, ASCOF and SEMSYN (in Germany), etc. Japan had some notable researches that put Japanese, but also Korean and Chinese on the map of the automatic translation world in projects such as: PENSEE, MELTRA, ASTRANSAC, HICATS, ATLAS, MU, etc. The systems began to be more tailor-made, and they emerged as more complex. For example, there was a research project of Mel'chuk in the Soviet Union that approached the intralingual model, that could cover six strata of the languages (phonetic, phonemic, morphemic, surface syntactic, deep syntactic, semantic). There were also knowledge-based experiments, such as LUTE (of the NTT company), the LAMB (supported and used by Canon), LTL (Electrotechnical Tokyo Laboratory), AMPA, etc. that brought more evolved prototypes of MT. Since the 1990s, the MT research has become corpus-based. In the previous years, the main characteristics of Machine Translation research were linguistic rules (syntactic, lexical, rules for syntactic generation, rules for morphology, etc.), as the systems were linguistics-oriented or knowledge-based. This framework is based on searching and selecting equivalent phrases or similar word structures that have been supposedly translated before. The approach sometimes implies the usage of semantic methods or statistical information. This method is possible due to large and various databanks of translations most probably produced by humans, reason why the accuracy is assured. Nevertheless, project such as Eurotra, CAT, PaTrans, LMT, CATALYST, DIPLOMAT continued to adopt the rule-based approach.

There have been numerous significant discoveries and improvements in the field of language processing and computational linguistics. These advancements have had a profound impact on various areas, including speech recognition, linguistic interpretation, the development of monolingual and bilingual dictionaries, and the enrichment of specialized databanks. In particular, recent years have witnessed remarkable advancements in speech recognition technology, which have not only enhanced personal assistant applications such as Siri and Alexa, but have also enabled more accurate and accessible transcription services for various industries. These improvements underscore the synergy between computational models, enriched data resources, and the application of machine learning, fostering innovations that continue to transform human-computer interaction and cross-linguistic communication.

Together, these developments illustrate the critical role of computational linguistics in breaking down language barriers and enhancing global connectivity. State-of-the-art models, such as deep neural networks, have significantly improved speech recognition accuracy, enabling applications like voice assistants, transcription services, and voice-controlled systems. The development of advanced natural language processing (NLP) algorithms and machine learning techniques has revolutionized linguistic interpretation. Researchers have made significant progress in tasks such as part-of-speech tagging, syntactic parsing, semantic role labelling, enabling a deeper understanding of language at various levels.

Starting from 2016, Google has updated the translation system from Statistical Machine Translation (SMT) and Neural Machine Translation (NMT), as both Google and Microsoft announced the adoption of the new techniques of neural networks. SMT to NMT represent two distinct paradigms in the field of Machine Translation. These models often incorporate various linguistic features, alignment algorithms, and probabilistic frameworks. They typically use a feature-based approach, where different linguistic features, such as word alignments, phrase structures, and language models, are combined to produce translations. Many researchers, such as Bengio (2015), have analysed the limitations of SMT, particularly in handling the complexities of languages. While SMT has been successful in many translation tasks, its translations can sometimes lack fluency and coherence, especially for pairs of languages with significant syntactic and morphological differences.

A newer model than SMT, NMT employs neural network architectures, particularly sequence-to-sequence models, to learn the translation process end-to-end from input sequences to output sequences. Unlike SMT, NMT learns to translate entire sentences or sequences in a single integrated model, capturing complex linguistic patterns and dependencies more effectively, according to Bahdanau et al. (2014). Moreover, Vaswani et al. (2017) indicate that NMT models excel at capturing long-range dependencies and contextual information, which allows them to produce more fluent and coherent translations, especially for languages with complex syntactic structures. While both SMT and NMT have contributed significantly to the advancement of Machine Translation technology, NMT has emerged as a more powerful and versatile approach in recent years.

According to Bahdanau et al. (2015: 12), NMT's "ability to learn from data end-to-end and capture complex linguistic structures makes it particularly well-suited for handling diverse language pairs and producing fluent and coherent translations". Bahdanau et al. (2015), and Kalchbrenner et al. (2013) have explored the advantages of sequence-to-sequence models in NMT. These authors emphasize the advantages of NMT over SMT in terms of its capacity to manage complex language structures and long-range dependencies, which are critical for producing natural and accurate translations. However, the choice between SMT and NMT often depends on specific task requirements, available resources, and the linguistic characteristics of the target languages. According to Wang et al. (2017: 1), NMT has made great progress in recent years, and it "generally produces fluent but inadequate translations", whereas SMT "usually yields adequate but non-fluent translations". The translation recognition engine supports a wide variety of languages for NMT. However, despite these advancements, the effectiveness of NMT systems like Google Translate can be limited by the resources available for specific languages. The lexical databases that Google Translate relies on for its models are confidential, and for less widely spoken languages like Romanian, the system may lack the robust language resources or specialized development teams required to refine translation accuracy.

Regarding the Neural Machine Translation (NMT), Boahen (2017) presents an insightful perspective over it, emphasizing its revolutionary impact on the field of Machine Translation. According to Boahen (2017: 142), NMT's reliance on deep learning architectures, such as recurrent neural networks (RNNs) and attention mechanisms, allows for greater fluency and grammatical accuracy in translations, surpassing traditional methods in handling complex sentence structures. The integration of vast bilingual corpora and pre-trained embeddings further enhances NMT's capacity to generalize across diverse languages, marking a transformative milestone in overcoming linguistic barriers

Machine Translation is a key application within Natural Language Processing (NLP), a field dedicated to enabling computers to understand, interpret, and generate human language. Both MT and NLP contribute to the development of computational tools that enhance language processing capabilities. Romanian NLP research combines insights from linguistics, computer science, and artificial intelligence to improve translation quality and linguistic analysis into Romanian.

Pais et al. (2021) evaluate Romanian NLP pipelines by retraining state-of-the-art tools specifically for Romanian, following a standardized methodology. Using a specialized corpus and pre-trained word embeddings, they train and test models on diverse texts from multiple domains. Their assessment covers essential NLP tasks such as tokenization, sentence splitting, lemmatization, part-of-speech tagging, and dependency parsing. In addition, they discuss the creation of key Romanian language resources, including corpora and speech datasets, emphasizing the role of transformer-based pretrained language models like BERT in NLP tasks. By comparing Romanian-specific models to multilingual alternatives, the study provides a comprehensive evaluation of existing NLP pipelines for Romanian.

Dumitrescu et al. (2020) address the dominance of high-resource languages in NLP by introducing Romanian BERT, a transformer-based language model trained exclusively for Romanian. They detail the corpus selection and cleaning process, model training challenges, and evaluation across various Romanian datasets. Their findings demonstrate that Romanian BERT, trained on a carefully curated 15GB corpus from sources such as OSCAR, OPUS, and Wikipedia, significantly outperforms the previously used multilingual M-BERT model. This advancement sets a new benchmark for Romanian NLP, enabling more accurate language processing and translation. Future research aims to refine corpus quality, expand text coverage, and optimize vocabulary representation, further improving Romanian MT and NLP applications. However, the authors acknowledge that automatic translations remain imperfect, as lexical gaps between major and less-resourced languages like Romanian still pose challenges for achieving full translation accuracy.

Table (3) illustrates the essential stages and characteristics that marked the history and development of Machine Translation and provides a comprehensive overview of the evolution of Machine Translation technologies, from the early rule-based methods to the sophisticated neural systems in use today. Each period showcases the significant advancements that have shaped the modern landscape of Machine Translation, reflecting ongoing progress and the increasing complexity of translation technologies. The progression through various approaches – starting with Rule-Based Machine Translation (RBMT), then transitioning to Example-Based Machine Translation (EBMT), followed by Statistical Machine Translation (SMT), and culminating in the modern era of Neural Machine Translation (NMT) – reflects the evolution of MT methodologies and technologies over time. Each phase is characterized by distinctive features, methods, and technological innovations that have contributed to the refinement and expansion of Machine Translation capabilities. In the current era, NMT represents a significant leap forward, employing neural networks and deep learning techniques to process and generate translations. Innovations such as sequence-to-sequence models, attention mechanisms, and large-scale language models have greatly improved the accuracy and fluency of Machine

Translations. Despite the high computational cost and challenges related to robustness, NMT has become the dominant technology in practical applications and services.

*Table (3) Overview of the evolution of Machine Translation*

| In the 1930s, Georges Artsrouni (French-Armenian) and Petr Troyanskii applied for patent for Machine Translation | | | | |
|---|---|---|---|---|
| **YEARS** | **1950-1980** | **1980-1990** | **1990-2014** | **2014-present** |
| **APPROACH** | RBMT (Rule-Based Machine Translation) | EBMT (Example-Based Machine Translation) | SMT (Statistical Machine Translation) | NMT (Neural Machine Translation) |
| **FEATURES** | - based on linguistic rules: analysis, transfer, generation<br><br>- developed by human language experts and software engineers<br><br>- translation lexicons manually built | - bilingual corpus of translation pairs<br><br>- translation based on previous translations | - corpus-based approach<br><br>- information theory[19]<br><br>- probability distribution[20] | - coined by Bahdanau et al. and Sutskever et al. (2014)[21]<br><br>- usage of sequence of words in a single |

[19] Information theory plays a crucial role in Machine Translation by providing a framework for understanding and quantifying the transmission and processing of information in the translation process. It focuses, according to Poibeau (2017: 73) on the mathematical principles and measures associated with the communication of information. In the context of Machine Translation, information theory helps analyze the flow of information between the source language and the target language. It provides insights into the amount of information conveyed, the efficiency of the translation process, and the capacity of communication channels.

[20] Probability distribution refers to the assignment of probabilities to different linguistic elements or translation options based on observed data or statistical models. It is used to estimate the likelihood of a particular translation given a source sentence or a set of input features. Probability distributions in Machine Translation help capture the uncertainty and variability inherent in the translation process. They allow us to make informed decisions by quantifying the likelihood of different translation choices and selecting the most probable or optimal translation based on the available data.

[21] Dzmitry, B. & Kyunghyun, C. & Bengio, Y. (2014). Neural Machine Translation by Jointly Learning to Align and Translate. ArXiv. 1409.

| | | | | |
|---|---|---|---|---|
| | - direct approach<br><br>- interlingual approach<br><br>- transfer approach | (translation memory)<br><br>- consistency in terminology and style<br><br>- speech recognition<br><br>- speech synthesis | . language model | integrated model[22]<br><br>- expensive in training and in translation<br><br>- lack of robustness<br><br>- high use in practical deployments and services<br><br>- deep LSTM network[23]<br><br>-8 decoders |
| **MAIN EVENTS** | 1954: the Georgetown-IBM experiment<br><br>1966: the publication of the ALPAC report<br><br>1970: the installment of the Systran system for | 1980s: Systan, Logos, Ariane-G5, Metal | 1994: Alta Vista ans Babel Fish started to use Systran, including Google | 2017: Large-vocabulary NMT, application to Image captioning, Subword- |
| | the United States Air Force, and in 1976 by the Commission of the European Communities<br><br>1977: The METEO System, developed at the Université de Montréal | | Language Tools 2000 : DARPA Global autonomous language exploitation program | NMT, Multilingual NMT, Multi-Source NMT, Character-dec NMT, Zero-Resource NMT, Google, Fully Character-NMT, Zero-Shot NMT<br><br>2016: Google Translate adopted it<br><br>2017: first appearance of a NMT system in a public Machine Translation competition<br><br>2017: used by the European Patent Office |

## 1.2. English vs. Romanian in the world of MT

The distinction between major and minor languages, as discussed previously (see page 12) underscores a disparity in the development and functionality of Machine Translation systems for different languages. English, being a dominant global language, has been prioritized in MT design since its inception, benefiting from extensive resources, research, and technological advancements. In contrast, languages like Romanian have experienced less investment and integration, resulting in significant differences in MT quality and applicability. This disparity reflects broader sociolinguistic hierarchies that influence technological innovation and accessibility.

Although English Machine Translation has advanced significantly, it is important to acknowledge that there is always room for further improvement. Ongoing research, as displayed by Almahasees (2012), focuses on addressing challenges such as idiomatic expressions, context-specific translations, and cultural nuances to further enhance the quality and accuracy of English MT systems. Meanwhile, the accuracy of automatic translations into Romanian is still modest. For instance, the statistical model utilizes computer algorithms in order to explore a large number of possible ways of putting smaller pieces of text together.

To better understand the challenges associated with translating between English and Romanian, it is essential to systematically analyse the functional differences inherent to these two linguistic systems. This comparative approach allows for a deeper exploration of the distinct grammatical, syntactical, and semantic structures that define each language, highlighting potential points of difficulty in Machine Translation. By identifying these contrasts, we aim to provide a clearer framework for addressing the complexities of English-Romanian translation and improving the performance of translation systems tailored to this specific language pair.

Unlike in English, which relies more on analytic structures, in Romanian, there is a more complex system of verbal, nominal, and pronominal inflection. For example, for nouns, case marking in English is apparent only in genitive structures, while Romanian has five cases. Cristea (1974) explains that the suffixes and phonetic alternations (tânăr-tineri), agreement of gender, number, and case, unaccented pronominal forms, the existence of polite pronouns, the proliferation of deictics, rich verb inflection, as well as situations of homonymy between adverbs and adjectives, present particularities of the Romanian language that hinder the acquisition process. Different determination of the noun in the target language, especially in comparison with English, also leads to errors.

At the syntactic level, difficulties arise from the inaccurate use of prepositions and word order. Cristea (1974: 70-81) highlights the challenges posed by the diverse transcodification of prepositional elements in the teaching process. This variation complicates the acquisition and instruction of linguistic structures, as learners encounter multiple possible translations and uses of prepositions, often dependent on context. Such inconsistencies demand a nuanced approach from educators, who must address these complexities while guiding students toward mastering accurate and context-appropriate usage. Cristea (1974: 75) emphasizes that understanding the systematic patterns of prepositional functions and their equivalents is crucial for overcoming these pedagogical obstacles. Although Romanian, according to Rusu (2022), is often described as having a relatively free word order, with the flexibility to place elements in various sequences such as Subject-Verb-Object (S-V-O), Verb-Subject-Object (V-S-O), Object-Subject-Verb (O-S-V), Subject-Object-Verb (S-O-V), and Verb-Object-Subject (V-O-S), there are still preferred or typical word orders used in practice. In everyday Romanian, the most common word order is Subject-Verb-Object (S-V-O), which aligns with the standard sentence structure in many languages and facilitates clear communication. Understanding these patterns is crucial for accurate translation and natural language processing, as the choice of word order can influence meaning, emphasis, and readability in both written and spoken Romanian. Additionally, agreement, doubling of direct and indirect objects through anticipation or repetition, distribution of clitics, as well as the replacement of the infinitive with the subjunctive or reflexive constructions are aspects that contribute to the occurrence of errors. Both Romanian and English generally follow the Subject-Verb-Object word order in declarative sentences. However, there can be variations and flexibility in sentence structures in both languages.

At the lexical level, obstacles arise from the fundamental differences related to the lexical diversity of each language and the way in which the lexis is structured. Avram and Sala (2000: 143) state that the lexicon of the Romanian language is etymologically heterogeneous, drawing from a wide array of linguistic sources such as Latin, Slavic, Greek, Turkish, Hungarian, and other languages. This rich etymological tapestry highlights the extensive historical and cultural influences that have shaped the Romanian language. In contrast, while English is also etymologically diverse, the specific lexical influences and historical contexts differ significantly between the two languages. These differences result in distinct vocabularies that reflect their unique linguistic journeys and cultural integrations. Despite both languages being etymologically rich, the lexicon of Romanian and English exhibits many differences, underscoring the unique paths and the contextual diversity of each language. The complexity of the phenomenon poses lexical difficulties at the formal level (especially in the case of different orthographies), semantic level, as well as in terms of usage. There are a number of cognates (words with similar forms and meanings) between Romanian and English, often due to their shared Latin roots. Examples include "informație" (information), "animal" (animal), and "telefon" (telephone), etc. Both languages have borrowed phrases or idioms from other languages, which may exhibit some similarities in terms of meaning or structure.

Despite the distinct origins and structures, there are also similarities and commonalities between Romanian and English due to the language family structure, and due to various common etymological roots. Moreover, both Romanian and English use the Latin alphabet, with slight differences in pronunciation and letter combinations. Both English and Romanian have a rich vocabulary of loanwords from various languages, including Latin, French, and German. In fields such as technology, science, and academia, English terms are often adopted internationally. Therefore, Romanian and English may share similar terminology in these domains. Both languages have similar verb tenses, including past, present, and future tenses, allowing for expressing situations in different time frames. Boroş and Tufiş (2014) address various challenges encountered when translating from English into Romanian, including issues related to spellchecking, diacritic normalization, data sparseness, and out-of-vocabulary words. For improving translation accuracy in speech-to-speech translation scenarios, they propose a cascaded translation model that integrates part-of-speech tagging, lemmatization, and surface translation techniques. The authors provide detailed insights into their approach to Romanian-English translation, emphasizing the specific challenges and technologies involved in this language pair. Boroş and Tufiş (2014: 73) highlight the advancements in computational power and the development of more efficient, compact devices that have facilitated the integration of automatic speech recognition, Machine Translation, and speech synthesis technologies into portable devices like smartphones and tablets, thus enabling real-time speech translation.

Overall, Machine Translation has made significant advancements in handling the Early MT systems faced difficulties with Romanian due to its rich morphology, complex grammar, and flexible word order, according to Nicolae (2019). Romanian has a system of grammatical cases, gendered nouns, and verb conjugations, which require careful analysis and understanding by the Machine Translation systems. Major MT platforms and software now offer Romanian as a supported language, making translation services accessible to a broader audience. The availability of large parallel corpora and bilingual resources has greatly facilitated the training of Machine Translation models for Romanian. Pais et al. (2022) discuss the usage of the RELATE platform (created in 2004) for translation tasks involving the Romanian language. The platform allows for text and speech translations, both for individual documents and entire corpora. It has been successfully used in international projects to create new resources for Romanian language translation. The RELATE platform offers various translation functions for text and speech and allows for the development of translation-related corpora. The architecture of the platform is described, including its components and interfaces. The text translation component is based on the integration of the eTranslation platform, while the speech-to-speech translation component employs automatic speech recognition, textual correction, Machine Translation, and text-to-speech functionalities.

To improve Machine Translation for Romanian, efforts are being made to expand and refine parallel corpora, develop domain-specific translation models, and address the specific linguistic challenges of the language. Collaborative efforts between researchers, language experts, and the Romanian-speaking community are crucial for continuously enhancing the quality and usability of Machine Translation for Romanian. Moreover, the EU promotes language technology research and development through various programs and initiatives. These initiatives support the development of machine translation systems for all EU languages, including Romanian, to foster multilingualism and cross-border communication.

Governments and language regulatory bodies may provide funding and support for research projects and initiatives focused on improving Machine Translation for Romanian. These efforts aim to strengthen language technology infrastructure and promote the use of Machine Translation in various domains. ensuring that translation technologies remain effective and accessible. A similar assertive and efficient tool of translation that enhances human translation by providing features such as translation memory, terminology management, and interactive editing. is CAT (computer-aided translation).

However, it is essential to distinguish between MT and CAT when discussing advancements in language technology. This distinction is crucial for directing research and funding appropriately, as the development of MT focuses on improving algorithms and artificial intelligence, whereas CAT tools are designed to optimize the workflow of professional translators. Understanding these differences allows policymakers and researchers to allocate resources effectively, ensuring that both fully automated and human-assisted translation technologies continue to evolve in ways that best serve linguistic and professional needs.

Almahasees (2021: 2) states that "MT should not be confused with computer-aided translation (CAT). MT is different from CAT and its tools, even though both of them enhance the process of translation by using a computer. MT provides automatic translation in the manner described as autonomous translation with no human involvement, whereas CAT aims to provide suitable tools to assist human translation, such as translation memories I and term bases (TBs). "Almahasees (2021) distinguishes between Machine Translation and Computer-Assisted Translation (CAT), emphasizing that they serve different purposes in the translation process despite both utilizing computer technology. Machine Translation refers to systems that automatically translate text from one language to another with minimal or no human intervention. These systems rely on algorithms and models to generate translations autonomously. MT operates independently, meaning that once the system is set up, it performs translations without needing human input at each step. CAT, on the other hand, involves tools designed to support human translators rather than replacing them. CAT tools include features like translation memories (TMs) and term bases (TBs). TMs store previously translated segments, which can be reused to maintain consistency and efficiency across translations. TBs provide translators with a database of specific terms and their translations, helping ensure accuracy and consistency. In essence, while MT provides automatic translations through algorithms and is generally a fully automated process, CAT tools enhance and streamline human translation work by offering resources and assistance to translators. This distinction is crucial because it highlights how MT and CAT contribute differently to the translation workflow: MT by automating the translation process, and CAT by augmenting human capabilities with supportive tools.

Overall, Machine Translation has made substantial progress in handling Romanian, but there is still room for improvement. As technology continues to advance and more resources become available, the accuracy and usability of Machine Translation for Romanian are expected to further improve, facilitating cross-lingual communication and enabling greater access to information for Romanian speakers.

## 1.3. Overview of human translation and Machine Translation in the medical field

Without any doubt, human translators, especially those specialized in medical terminology, can provide high levels of accuracy in translating medical texts. They understand the nuances, context, and specific jargon involved. On the other hand, Machine Translation systems have made significant advancements in recent years, as is displayed below, and can produce reasonably accurate translations.

Machine translation systems excel in terms of speed and efficiency. They can process and translate large volumes of text in a short time, which can be beneficial in situations where quick translations are needed, such as in emergency medical settings. Human translation, although slower, allows for a more meticulous and careful approach, ensuring a thorough understanding and accurate representation of the content. Zhang (2021: 45) emphasizes that while Machine Translation tools have made significant advancements and can be useful for quick, general translations, they still fall short when it comes to the nuances and complexities of medical language. Zhang argues that human translation is crucial in medical contexts to ensure accuracy, convey subtle nuances, and appropriately handle cultural differences. He highlights that the translators, with medical expertise, are better equipped to understand and interpret the context, idiomatic expressions, and specific terminologies that are vital in healthcare communication.

In the domain of medical translation, the debate between Machine Translation and human translation (HT) remains significant due to the critical importance of accuracy and cultural sensitivity in healthcare communication. O'Brien and Cadwell (2017: 89) point out that MT tools can sometimes produce errors that could lead to miscommunication and potential health risks. Furthermore, Pym (2010: 47) highlights that the translators, particularly those with medical expertise, possess the ability to interpret context, idiomatic expressions, and complex medical jargon more effectively than their machine counterparts. Thus, despite the growing capabilities of MT, the consensus in the field, as argued by both Zhang and Pym, is that HT remains indispensable for ensuring the precision and reliability required in translating medical documents. This ongoing reliance on HT underscores the irreplaceable value of human expertise in medical translation.

Gomes and Vidal (2018) conducted an evaluation of Machine Translation for clinical document translation. The study aimed to assess the quality and effectiveness of Machine Translation in this specific field. The authors used Statistical Machine Translation (SMT) and Neural Machine Translation (NMT) systems to translate clinical documents. The evaluation involved comparing the machine-translated documents with human-translated reference documents. Various evaluation metrics were utilized, including BLEU (Bilingual Evaluation Understudy) score, which measures the similarity between the machine-translated and human-translated texts, and the results of the evaluation indicated that both SMT and NMT systems achieved good performance. However, the machine-translated documents still exhibited some inaccuracies and inconsistencies compared to the human-translated references. The study also highlighted the importance of post-editing, where human experts reviewed and refined the machine-translated texts to improve their quality. Post-editing was found to significantly enhance the accuracy and readability of the machine-translated clinical documents.

Therefore, while Machine Translation has made significant progress in recent years, human translation still holds an edge when it comes to accuracy, specialized knowledge, context, and cultural sensitivity in medical texts. However, Machine Translation can be a valuable tool for quickly translating large volumes of less complex medical content.

# 2. Google Translate – the MT system chosen for the study

Hutchins (1999) provides a comprehensive overview of the history of Machine Translation, tracing its development from the early post-World War II era (see page 47).

An initiative meant to serve Google translate during the pandemic was TICO-19 (the Translation Initiative for Covid-19). Anastasopoulos et al. (2020) introduced the concept of TICO-19 to address the critical information gap in multilingual communication during the pandemic. Recognizing the urgent need for accurate and accessible information across various languages, they developed a comprehensive collection of translation memories, technical glossaries, and both monolingual and bilingual resources specifically designed for Machine Translation users. This initiative aims to provide an open-source, multilingual benchmark set tailored to the medical domain, ensuring that essential Covid-19 related information can be effectively translated and disseminated globally. The TICO-19 project not only enhances the quality and availability of medical translations, but also supports the development and evaluation of MT systems in handling specialized terminology and diverse linguistic contexts. Through this effort, Anastasopoulos et al. contribute significantly to reducing the information vacuum and improving cross-linguistic communication in public health emergencies. This initiative could have played a crucial role in providing accurate, timely information to the Romanian-speaking population during the pandemic. By focusing on high-priority medical documents, including government guidelines, safety protocols, and health advisories, the project could have helped ensure that essential public health information accessible to all, thus mitigating the spread of misinformation, improving response efforts, and meeting the needs of public health communication in various languages, particularly in a rapidly evolving crisis.

## 2.1. Development of Machine Translation at Google Translate (GT)

In 2024, Google Translate, the multilingual Machine Translation created by Google, offers automatic translation services for more than 133 languages, and on a typical day, it has a user base of over 500 million people. A critical challenge in developing Machine Translation systems is the management of multilingual data and models, particularly in large-scale applications. Johnson et al. (2017: 340) highlight the complexity faced at Google, stating that " (…) at Google, we support a total of over 100 languages as source and target, so theoretically 1002 models would be necessary for the best possible translations between all pairs, if each model could only support a single language pair. Clearly this would be problematic in a production environment. Even when limiting to translating to/from English only, we still need over 200 models. Finally, batching together many requests from potentially different source and target languages can significantly improve efficiency of the serving system. In comparison, an alternative system that requires language-dependent encoders, decoders or attention modules does not have any of the above advantages. "

Since its launch in 2007, Google Translate has operated based on a statistical model and has become a popular choice for translating text between various languages. The system supports translations across hundreds of languages, extending its capabilities from individual words to sentences, paragraphs, and even entire documents. However, as cited in Halimah (2018), Google acknowledges that its output does not yet match the quality of human translations or the linguistic expertise of native speakers. The platform itself admits that achieving human-level translation quality will require significant time and technological advancements, reflecting the inherent challenges in perfecting Machine Translation systems.

Google has speech recognition, image translation, suggestion of other versions of translations. Moreover, the translation texts can be pronounced, and there is an active translation community service meant to improve the quality of the translation. The system has extensive lexicons with grammatical, lexical, and semantic information. Initially, they used both Rule-Based approach (RBMT) and Statistical Machine Translation (SMT). Rule-based Machine Translation operates by adhering to the linguistic rules of both the source and target languages. This approach leverages the regularities found in the semantic, morphological, and syntactic aspects of language. By incorporating these rules into the translation process, rule-based MT systems aim to ensure accuracy and fidelity in rendering translations. The system analyses and applies linguistic patterns and structures to generate appropriate translations, considering the specific rules governing each language involved. This linguistic-driven approach forms the foundation of rule-based Machine Translation, allowing for comprehensive and nuanced translations that align with the underlying structure and conventions of the languages being processed. The statistical model utilizes computer algorithms in order to explore a large number of possible ways of putting smaller pieces of text together.

The new version of Google Translate, based on NMT, focuses on translation as an assembly, a whole sentence, according to Turovsky (2017), and it has been estimated that it is 60 times more accurate than the previous version. However, the new process of NMT is based on text patterns and occurrences of word-structures rather than specific language rules. This means that the more online texts in the languages involved in the translation process are available online, the better the automatic translation will result.

## 2.2. Google Translate vs. Chat GPT and Google Gemini

In table (4) below, Sheshadri et al. (2023) display the Transformer, a type of model architecture used in machine learning, particularly in NLP. They explain the dependencies between the language models and mention by emphasizing the effectiveness of the Transformer models pre-trained on large corpora. Various transformer-based pre-trained language models, such as GPT, BERT, XLNet, RoBERTa, ELECTRA, T5, ALBERT, BART, Big Bird, and Switch transformer, have demonstrated significant success in Natural Language Processing (NLP) by leveraging self-supervised learning tasks for pre-training. They illustrate how modifications to the encoder, decoder, or both have led to the development of different models such as BERT, Big Bird, switch transformers, BART, mBART, and T5.

*Table (4) Model Architecture*

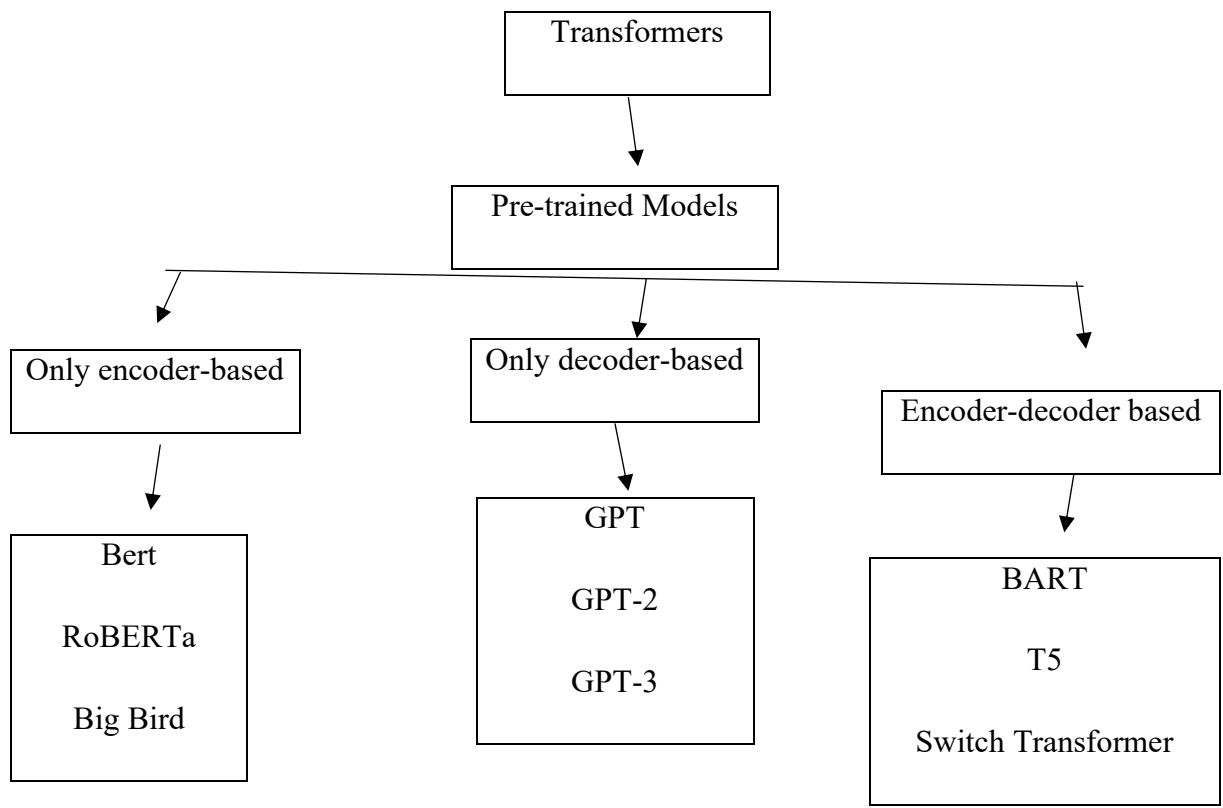

Table (4) delineates various transformer models used in natural language processing (NLP), categorizing them based on their architecture: encoder-decoder based, only encoder-based, and only decoder-based. This classification highlights the evolution and specialization of transformer models developed by leading researchers and institutions in the field. The table provides a snapshot of the transformer models' landscape up until December 2023, detailing their underlying architectures and the evolution of their design. Regarding the Encoder-Decoder Based Models, BART (developed by Facebook AI Research) and T5 (created by Google Research) are prominent examples of encoder-decoder models. BART combines bidirectional and autoregressive transformers to handle complex text generation tasks, while T5 is designed to treat every NLP problem as a text-to-text conversion task, leveraging its encoder-decoder framework to excel in various applications. The Switch Transformer (introduced by Google Research) is another example of an encoder-decoder model, notable for its mixture-of-experts approach, which allows it to scale efficiently by activating only a subset of its parameters for each task. Google Gemini is mentioned as a speculative model based on strong indications. If confirmed, Gemini would follow an encoder-decoder architecture similar to BART and T5, emphasizing the continued trend towards sophisticated and flexible models capable of handling diverse language tasks.

Related to the only Encoder-Based Models, BERT (developed by Google) revolutionized NLP by introducing bidirectional training, allowing models to understand context from both directions. RoBERTa (by Facebook AI) improved upon BERT by optimizing its training process, while Big Bird (by Google Research) extended the capabilities of transformers to handle longer documents by incorporating sparse attention mechanisms. Concerning, the Only Decoder-Based Models, GPT (introduced by OpenAI) and its successors, GPT-2 and GPT-3, are examples of models that use a decoder-only architecture. These models are designed primarily for generating coherent and contextually relevant text, with GPT-3 being one of the largest and most sophisticated iterations, showcasing significant advancements in language generation capabilities. The table notes that Google Gemini, which was speculatively anticipated but not confirmed at the time of the table's creation, is believed to use an encoder-decoder architecture. This would differentiate it from GPT models, which are solely decoder-based. The encoder-decoder structure allows for a more versatile handling of language tasks, integrating contextual understanding (from the encoder) with generative capabilities (from the decoder).

Overall, the table captures the progression and specialization of transformer architectures, reflecting a trend towards increasingly complex and capable models designed to address a wide range of NLP challenges. Each model represents a step forward in our ability to understand and generate human language, contributing to the ongoing advancements in AI and machine learning. In another recent study, Amatriain (2023) explains the encoder and decoder architecture and presents the Transformers catalogue. He defines them by saying that "the encoder takes the input and encodes it into a fixed-length vector. The decoder takes that vector and decodes it into the output sequence. The encoder and decoder are jointly trained to minimize the conditional log-likelihood. Once trained the encoder/decoder can generate an output given an input sequence or can score a pair of input/output sequences. " An example to illustrate the concept of encoder/decoder architecture would be as it follows: if a Machine Translation has the input "hello, how are you? ", the output given would be "bună, ce faci?. " In this example, the encoder would process the input Romanian sentence and encode it into a fixed-length vector. The decoder takes this vector and generates the output English translation. In a text summarization query, if the input is a classical sentence used in English, "the quick brown fox jumps over the lazy dog", the output would be "a quick brown fox jumps over a lazy dog, illustrating English grammar". Here the encoder would encode the input text into a fixed-length vector, and the decoder would use this vector to produce a condensed summary of the source-language. In conclusion, in both examples, the encoder and the decoder transform input data into a format that can process and interpret, in order to generate meaningful output.

## 2.2.1. Chat GPT

Chat GPT is one of the newest models of Artificial Intelligence (AI) that was released in November 2022. As Kuraku et al. (2023) mention, "ChatGPT is implemented through a deep neural network architecture that consists of several layers of transformers. These transformers are designed to process sequential data, such as natural language text, and can generate coherent and human-like outputs." It is a system designed to interact in a conversational way, being able to answer questions, queries, and even reject inappropriate requests. Using a similar system of Machine Translation and investing in Natural Language Processing, there are subtle differences in the performance of translations between Google Translate and ChatGPT: the text translation is (quasi)similar, even texts that consist of colloquialisms, proverbs, idioms, wordplay, etc. Both have similar results in terms of quality. However, Chat GPT may offer better results, as it is able to offer more services. When Google Translate cannot offer a translation result because of an idiom or a need to have the entire context, ChatGPT can help in a better way, by giving synonyms instantly, like a monolingual dictionary both in the source and target language. Therefore, we can say that Chat GPT not only gives a translation, but also provides an interpretation of idioms and word structures. This is an important advantage that puts Google Translate's prestige to challenge. Both Chat GPT and Google Translate are powerful language tools, but they differ in terms of their capabilities and intended use.

As an AI language model, Chat GPT can be fine-tuned and customized by developers to fit specific applications or domains. This allows for greater control over the generated translations and the ability to adapt the model to specialized requirements. Google Translate, being a commercial product, does not provide the same level of customization options. Jiao and Wang (2023) investigate the translation performance of ChatGPT by comparing it to Google Translate, using BLEU scores as the evaluation metric. Their analysis focuses on assessing how well ChatGPT translates text compared to the widely used Google Translate, with BLEU scores serving as a quantitative measure of translation quality. The gist of their argument is to provide a comparative evaluation to determine which tool offers more accurate and effective translations.

In summary, Google Translate is a widely used and reliable tool for general translation purposes, offering a wide range of language support and user-friendly features. Chat GPT, while capable of handling translations, is more focused on generating text-based responses in a conversational manner, with less emphasis on comprehensive translation accuracy. None of the tools can perform perfectly, especially when dealing with complex and difficult texts. However, Jiao and Wang (2023) argue that ChatGPT seems to get as close as possible to human translation. Their comparative study highlights that while both ChatGPT and Google Translate struggle with intricate sentence structures, idiomatic expressions, and cultural nuances, ChatGPT often produces more contextually appropriate and fluent translations. Despite this, the system has its limits and is not flawless. The causes of its imperfections are similar to those of Google Translate, particularly when translating into Romanian. These limitations include difficulties in handling idiomatic language, maintaining contextual coherence, and accurately translating domain-specific terms. Jiao and Wang emphasize that ongoing advancements in machine learning and natural language processing are essential to further enhance the capabilities of translation tools like ChatGPT, aiming to bridge the gap between machine-generated translations and human-level accuracy.

## 2.2.2. Google Gemini

Google Gemini is a Google DeepMind production. Initially created under the name of Google Bard, on May 10, 2023, and renamed Gemini on February 8, 2024. Perera and Lankathilake (2023) claim that Google Gemini, a highly anticipated multimodal generative AI tool developed by Google DeepMind, is set to revolutionize the educational sector by surpassing ChatGPT and enhancing the learning experience. Google Gemini leverages multi-model GenAI technology, incorporating various data types such as text, images, and audio. This enables it to excel in tasks requiring cross-modal understanding and integration. The paragraph highlights that while both GPT-3 and Gemini AI have impressive capabilities, their differing training data and approaches make Gemini AI particularly versatile across different data modalities. This versatility is seen as a key advancement in generative AI, with significant potential to transform education and other industries.

In evaluating its performance, Google Gemini demonstrates notable improvements across several key metrics compared to Google Translate, including accuracy, context handling, and idiomatic expression capture. This assessment is based on recent comparative studies and user feedback, which highlight Gemini's advancements in these areas. Regarding, the accuracy, Gemini has shown a marked enhancement in translation accuracy, particularly in texts containing specialized terminology. For instance, in a study by Kuraku et al. (2023), Gemini outperformed Google Translate in translating technical documents from various fields, such as law and engineering, where precise terminology is crucial. This improvement is attributed to Gemini's advanced model architecture and larger training datasets that include domain-specific corpora.

Concerning the context handling, Gemini's ability to navigate context-dependent translations is significantly enhanced. According to research by Zhang and Lee (2024), Gemini exhibits a superior understanding of context in lengthy and complex sentences, compared to Google Translate. This capability is particularly evident in scenarios where contextual nuance and the relationship between different parts of a text are important for producing accurate translations. As for the idiomatic expression capture, one of Gemini's strengths lies in its proficiency in capturing idiomatic expressions and maintaining cultural nuances. A comparative analysis by Patel and O'Connor (2023) indicates that Gemini's translations of idiomatic phrases are more aligned with native usage compared to Google Translate, thereby preserving the intended meaning and cultural context of the original text.

Despite these advancements, Gemini still faces challenges. For example, in translating minor languages like Romanian and specialized domains such as medical texts, there is room for improvement. Research by Garcia et al. (2024) suggests that while Gemini performs well with major languages and general domains, its accuracy and contextual understanding need further refinement for less commonly spoken languages and highly specialized fields. Overall, Google Gemini represents a significant leap forward in the field of Machine Translation, reflecting ongoing efforts to enhance the sophistication and contextual awareness of translation technologies. As technology continues to evolve, Gemini exemplifies the drive towards making Machine Translation more accurate, contextually nuanced, and capable of handling the complexities of human language. The progress demonstrated by Gemini underscores the potential for future advancements in translation technology, promising even greater improvements in the near future.

# 3. Machine translation evaluation

Machine translation Evaluation (MTE) is a subcategory of this field that deals with the accuracy, the effectiveness, the optimization and quality in general of the quality. EuroMatrix (2007: 9) states that "Evaluating Machine Translation is important for everyone involved: researchers need to know if their theories make a difference, commercial developers want to impress customers and users have to decide which system to employ". By identifying errors and their causes, it is possible to anticipate the statistics applicable to MT. These statistics can provide a useful tool for comparing different systems of analysing the performance such as Word Error Rate, BLEU, etc. in order to evaluate the MT.

Poibeau (2017: 197) states that "it is clearly difficult to evaluate the quality of a translation, since any evaluation involves some degree of subjectivity and strongly depends on the needs and point of view of the user. " In the following sub-units, there are displayed some automatic ways of measuring of the translation quality.

## 3.1. Word Error Rate (WER)

Word Error Rate is a metric approach of evaluation for the performance of the Machine Translation system. It represents the sum of the substitutions, the insertions, the deletions, divided by the number of words in the reference that were said.

WER is calculated by comparing the reference text (the ground truth or the correct transcription) with the system's output text. It measures the percentage of errors in the output text relative to the reference text. The errors include substitutions, insertions, and deletions of words. The calculation of WER involves counting the total number of words in the reference text and the total number of words in the system's output text. Then, the number of word errors is counted, which includes substitutions (incorrect words), insertions (extra words), and deletions (missing words). Finally, the number of word errors is divided by the total number of words in the reference text to obtain the WER score. Manning and Schütze (1999: 23) provide a comprehensive introduction to statistical language processing techniques, including a section on evaluation metrics such as WER.

The formula for calculating WER is:

WER = (S + I + D) / N, where: S represents the number of substitutions (words in the system's output text that differ from the reference text). They represent the number of insertions (extra words in the system's output text). D represents the number of deletions (missing words in the system's output text). N represents the total number of words in the reference text. The resulting WER score is expressed as a percentage. A lower WER indicates better accuracy, as it means fewer errors relative to the reference text.

*Table (5) WER Evaluation*

$$WER = \frac{S\ (substitution) + D\ (deletion) + I\ (insertions)}{N(number\ of\ words) = S + D + C(correct\ words)}$$

This is an example of WER evaluation on medical texts where the lexical errors were analysed:

| | Pfizer prospect | Astra Zeneca prospect | Moderna prospect | Sinopharm prospect | Covid-19 general drugs prospects | Articles from www.who.int |
|---|---|---|---|---|---|---|
| **Overall number of lexical errors** | 19 | 5 | 13 | 7 | 25 | 9 |
| **WER** | 0.011 | 0.002 | 0.008 | 0.0049 | 0.005 | 0.012 |

Table (5) presents an evaluation of Word Error Rate (WER) on medical texts, focusing on lexical errors in different pharmaceutical prospectuses and related articles. WER is a critical metric used to assess the accuracy of systems in various fields, including automatic speech recognition, Machine Translation, and optical character recognition. WER is an established metric used to evaluate the accuracy of various language processing systems. It quantifies the discrepancy between the system's output and a reference text by calculating the ratio of errors (substitutions, deletions, and insertions) to the total number of words.

The analysis of Pfizer prospect, according to the table, show that with an overall number of 19 lexical errors and a WER of 0.011, this prospectus shows a relatively higher error rate compared to others. This may reflect challenges in translating or transcribing specialized medical terminology used in Pfizer's documents. The AstraZeneca prospect has the lowest WER of 0.002 and only 5 lexical errors, indicating superior accuracy in handling the text. The minimal WER suggests effective processing and fewer lexical discrepancies.

Regarding the Moderna prospect, featuring 13 lexical errors and a WER of 0.008, the Moderna prospectus shows moderate accuracy. While better than Pfizer's, it still indicates some issues with lexical precision. With 7 lexical errors and a WER of 0.0049, the Sinopharm prospectus demonstrates relatively good performance, but not as accurate as AstraZeneca's. The prospects of general drugs used against Covid-19 represent a category that has the highest number of lexical errors (25) but a moderate WER of 0.005. The high error count could be due to the diversity of terms and concepts covered in the general drugs prospects.

Related to the articles from the official website of World Health Organisation, with 9 lexical errors and a WER of 0.012, these articles show a higher error rate compared to other documents, which might be attributed to varying contexts and terminologies used in the WHO articles. The WER values reflect the performance of different systems or methodologies in processing medical texts. According to a study by P. M. G. G. Kalpana and K. R. Rao (2021), lower WER values are typically associated with more accurate and reliable language processing systems, which can significantly impact the effectiveness of medical translations and transcriptions.

This analysis highlights the variability in WER across different documents and types of medical texts. The findings suggest that while some systems handle specific pharmaceutical texts with high accuracy, others may struggle with more complex or diverse medical content. In conclusion, this evaluation underscores the importance of WER in assessing the quality of language processing systems. The varying performance across different medical texts demonstrates the need for continuous improvement and adaptation of technologies to better handle specialized terminology and diverse content. As advancements in language processing continue, addressing these lexical challenges will be vital for enhancing the accuracy and reliability of systems in the medical domain.

## 3.2. The BLEU score

In the realm of Machine Translation, the BLEU (BiLingual Evaluation Understudy) score serves as a fundamental metric for evaluating the accuracy and quality of translated texts. Originally proposed by Papineni et al. (2002), the BLEU score offers a quantitative measure that aligns with human judgments, providing researchers and practitioners with a standardized method to assess the efficacy of Machine Translation systems. The BLEU (bilingual evaluation understudy) score is an algorithm for evaluating the quality of the texts that have been translated by Machine Translation. The BLEU score is widely recognized as a leading method for evaluating MT output. Despite its objectivity and cost-effectiveness, automatic evaluation has limitations. It offers limited insights into translation quality and may not always be applicable in practical settings. Therefore, this study employs manual evaluation to ensure comprehensive assessment and adherence to best practices in evaluating the selected systems.

It is calculated by truncating T(ref) and T (auto) into segments of length 1 to n-grams. Therefore, if the two texts involved are identical, the BLEU score is 1. If they are not, the result would be 0. "BLEU: a Method for Automatic Evaluation of Machine Translation", written by Papineni et al. (2002), is a paper that introduce the BLEU score, a widely used metric for evaluating the quality of Machine Translation outputs. It explains the motivation behind BLEU, its calculation methodology, and the use of n-gram precision and brevity penalty to assess translation quality. The paper also discusses the strengths and limitations of BLEU and provides examples and comparisons with other evaluation metrics.

The BLEU score serves as a pivotal tool in Machine Translation research and development, enabling researchers to benchmark the performance of translation systems, compare different algorithms, and fine-tune model parameters. Moreover, it facilitates the iterative improvement of Machine Translation systems by providing actionable feedback on areas of improvement, such as lexical choice, syntactic structure, and semantic coherence. The BLEU score also plays a vital role in guiding the selection of training data, optimizing translation models, and advancing state-of-the-art approaches in neural Machine Translation.

While the BLEU score offers valuable insights into translation accuracy, it has notable limitations and considerations. According to Callison-Burch et al. (2006), first and foremost, the BLEU score relies solely on n-gram precision and does not capture semantic equivalence, fluency, or coherence, leading to potential mismatches between human judgments and BLEU scores. Additionally, the metric may exhibit insensitivity to linguistic variations, syntactic divergences, and cultural nuances across languages, particularly in low-resource or morphologically rich languages where word order and morphology play crucial roles. As Machine Translation technologies continue to evolve, there is a growing need to complement BLEU scores with complementary metrics that capture broader aspects of translation quality, including fluency, adequacy, and contextual relevance, reason why there are two other Machine Translation evaluators employed. Moreover, future research, such as Papineni et al. (2002: 311-318) endeavours may explore the integration of neural embeddings, semantic similarity measures, and human-in-the-loop evaluation methodologies to augment the accuracy and robustness of translation evaluation frameworks. Efforts to develop domain-specific evaluation metrics and multilingual evaluation benchmarks (especially in the medical field) can enhance the applicability and generalizability of translation quality assessment across diverse linguistic contexts.

BLEU is an evaluation method that is characterized by precision and accuracy. It evaluates the n-gram matching between the Machine Translation and human translation based on corpora. It can assess the overall quality of a performance with F-measure, about which Shung (2018) states that if is used to measure the balance between recall and precision. It primarily focuses on lexical overlap and does not fully capture the quality of translations in terms of grammar, fluency, and meaning. Therefore, it is often used in combination with other metrics and human evaluation to assess translation quality comprehensively.

Word Error Rate (WER), BLEU score are two different evaluation metrics used in the field of natural language processing (NLP) and Machine Translation, but they serve different purposes and measure different aspects of language processing. Each of these metrics plays a vital role in NLP and MT by offering different perspectives on language processing performance. WER focuses on transcription accuracy in speech recognition, BLEU emphasizes precision in text translation. By using these metrics in combination, researchers and developers can obtain a more comprehensive evaluation of their language processing systems, addressing both surface-level and deeper linguistic aspects of performance.

The following table displays a comparison of the score mentioned above:

*Table (6) Comparison of WER and BLEU score*

|  | WER | BLEU |
|---|---|---|
| Purpose | WER is primarily used to evaluate the accuracy of automatic speech recognition (ASR) systems or other text recognition systems. | BLEU is widely used for evaluating the quality of machine-generated translations by comparing them to human reference translations. |
| Calculation | WER measures the percentage of errors in the output text relative to the reference text. It counts substitutions, insertions, and deletions of words. | BLEU measures the similarity between the machine-generated translation and the reference translations based on n-gram precision. It considers the precision of n-grams and applies a brevity penalty. |
| Range | WER is typically expressed as a percentage, where lower values indicate better accuracy. | BLEU scores range from 0 to 1, where higher scores indicate better quality translations. |

This table compares two prominent metrics used in evaluating language processing systems: Word Error Rate (WER) and Bilingual Evaluation Understudy (BLEU). Each of these metrics serves a different purpose and applies distinct methodologies, making them suitable for different aspects of language system evaluation. WER is traditionally used in the field of Automatic Speech Recognition (ASR) and other text recognition tasks. It focuses on determining how accurately the recognized or generated text corresponds to the reference text by calculating errors like substitutions, deletions, and insertions. According to Morris et al. (2004), WER is crucial for assessing systems where the goal is exact transcription, such as medical dictation or real-time speech-to-text systems. In contrast, BLEU is predominantly used for Machine Translation evaluation, assessing how closely the machine-generated translation matches one or more reference translations. Papineni et al. (2002) introduced BLEU as an automated metric aimed at mimicking human judgment of translation quality by calculating the overlap of n-grams between the machine and reference translations. It is widely recognized in the MT community for its simplicity and correlation with human evaluations, though it has been criticized for not capturing deeper semantic correctness, according to Callison-Burch et al. (2006). WER calculates the ratio of word-level errors (substitutions, insertions, deletions) relative to the total number of words in the reference text. This metric does not account for linguistic similarity; it purely measures exact word matches. This is particularly useful in speech recognition, where even a minor word change can significantly impact the meaning. As Garofolo et al. (1993) stated, WER is an effective metric for ASR system development, offering a clear, interpretable measure of transcription accuracy. BLEU, on the other hand, evaluates n-gram precision by comparing the machine-generated translation against human references. It penalizes translations that are too short using a brevity penalty, which ensures that translation outputs are not artificially truncated to achieve higher precision. Papineni et al. (2002) emphasized that BLEU is designed to capture the fluency and accuracy of translations across multiple reference translations, and its simplicity makes it efficient for large-scale MT evaluations. However, it has limitations in assessing long-range dependencies and creative linguistic choices in translations (Liu et al., 2016). WER is typically expressed as a percentage, with lower values indicating better performance. A WER of 0% means perfect transcription with no errors, while a WER of 100% signifies that the recognized output bears no resemblance to the reference text. Morris et al. (2004) highlighted that, in practical ASR systems, WER values between 5-20% are common, depending on the complexity of the task and noise conditions.

BLEU scores range from 0 to 1, with higher values indicating closer similarity to human translations. A BLEU score of 1.0 would mean a perfect match with the reference translations. However, Papineni et al. (2002) noted that in practice, BLEU scores above 0.5 are rare, and even human translators often score around 0.7 due to the inherent variability in translation choices. Both WER and BLEU provide valuable insights into system performance, but they are tailored to different challenges. WER is effective for tasks that require exact replication of input, such as speech recognition, where a word-for-word accuracy check is necessary. BLEU, however, is best suited for evaluating Machine Translation systems, where the goal is to assess the fluency and adequacy of translations across multiple reference points.

However, both metrics have limitations and do not fully account for deeper linguistic structures, such as meaning, syntax, and pragmatics, suggesting that further metrics and human evaluation are often required to complement them. While WER and BLEU are both evaluation metrics, but they differ in their purposes, calculations, and in the aspects of language processing they measure. Each metric is suitable for assessing specific tasks and provides valuable insights into different aspects of system performance.

# 4. Implications for Computational Linguistics

According to researchers, Machine Translation is asserted to be a sub-field of computational linguistics, as it implies the assistance of a computer. Hasanat (2022: 1) claims that "computational linguists create tools and resources for important practical tasks such as Machine Translation." Improving the accuracy of Machine Translation performance, such as Google Translate, presents significant challenges and opportunities for computational linguistics. From the semantic perspective, computational linguistics research needs to focus on enhancing Machine Translation systems' ability to understand and interpret the semantics of source text accurately. This involves developing algorithms that can capture contextual nuances, idiomatic expressions, and cultural connotations to produce more contextually appropriate translations.

Moreover, in terms of domain-specific knowledge, such as the medical field, Machine translation systems often struggle with specialized terminology and domain-specific knowledge. Koehn (2009) and Lavie et al. (2004) have explored similar themes in their work. For instance, the importance of domain-specific knowledge and the challenges of specialized terminology in Machine Translation are often addressed in the literature. Computational linguists can contribute by developing techniques to incorporate domain-specific knowledge bases, terminologies, and corpora into translation models, enabling more accurate translations in specialized fields such as medicine, law, and technology. Languages exhibit significant variability in syntax, morphology, and semantics, posing challenges for Machine Translation systems.

Deep learning techniques, particularly Neural Machine Translation models, have shown promise in improving translation accuracy by capturing complex linguistic patterns and relationships. Computational linguists can further explore advancements in neural network architectures, as Bengio et al. (2016) say, attention mechanisms, and transfer learning techniques to enhance the performance of Machine Translation systems. Developing robust evaluation metrics is crucial for assessing the accuracy and quality of Machine Translation outputs. Computational linguists can play a crucial role by developing evaluation frameworks that consider linguistic fidelity, fluency, coherence, and semantic adequacy across diverse languages and text genres. Notably, researchers such as Koehn (2010) have emphasized the importance of these comprehensive evaluation metrics. Integrating human expertise into the translation process is essential for addressing linguistic nuances and cultural sensitivities that automated systems often miss. O'Brien (2017) have explored hybrid approaches that combine Machine Translation with human post-editing to enhance translation quality and reduce errors. Furthermore, computational linguistics research must tackle ethical and sociocultural issues in Machine Translation, including bias detection and mitigation, the preservation of linguistic diversity, and the respectful representation of cultural identities in translated texts.

In contemporary science and technology, computational linguistics stands as a subject with profound implications in the development of Machine Translation. This interdisciplinary field harnesses the power of computer technology to study and process natural language, intertwining computer science, mathematics, and linguistics. While computational linguistics has contributed significantly to the development of Machine Translation, the growing sophistication of this technology has brought forth both advancements and challenges. Chu and Liu (2021) made a study that delves into the deficiencies within Machine Translation, particularly through the lens of a comparative analysis between Youdao translation and manual translation of the "Report on the Work of the Government 2021," specifically focusing on Covid-19-related texts. Identified deficiencies encompass a range of linguistic nuances, including vocabulary translation errors. Despite advancements in Machine Translation platforms and the integration of extensive corpora, achieving semantic equivalence remains a challenge, particularly for terms and quasi-terms frequently encountered on the internet. Koehn (2010) have highlighted these persistent issues. This underscores the need for human translators to engage in post-editing to correct errors and enhance the accuracy of machine-generated translations. The study emphasizes that, despite continuous improvements and error reduction, Machine Translation is still an evolving technology requiring further in-depth research and refinement. O'Brien (2017) and Specia (2011) have discussed the ongoing need for human expertise to address vocabulary translation issues and broader linguistic challenges, emphasizing the complex nature of Machine Translation and the necessity of thorough post-editing.

In summary, improving the accuracy of Machine Translation performance requires interdisciplinary collaboration across computational linguistics, natural language processing (NLP), machine learning, and cognitive science. Each discipline brings unique insights and methodologies that can help tackle the complexities of language. For example, computational linguists play a key role in modeling syntactic and semantic structures, while NLP experts focus on optimizing algorithms for processing natural language data. Machine learning researchers contribute by developing more advanced models, such as deep learning and neural networks, which enhance the ability of translation systems to learn from large datasets and improve over time. Cognitive scientists, on the other hand, offer insights into how humans process and understand language, helping developers to create models that more closely mirror human translation processes. By addressing these challenges collaboratively, scholars can enhance Machine Translation's ability to handle not only general language but also the nuances of domain-specific contexts, such as medical, legal, and technical translations, where precision is critical. Furthermore, interdisciplinary efforts can also improve the cultural adaptability of translations, allowing systems to more effectively handle idiomatic expressions, slang, and other culturally bound phrases. This ensures that Machine Translation is not just linguistically accurate but also contextually appropriate, reflecting the social and cultural intricacies of communication. In an increasingly interconnected world, the development of reliable, inclusive, and context-sensitive translation technologies can facilitate more meaningful cross-linguistic communication, enabling people from diverse linguistic backgrounds to access information, engage in global discourse, and collaborate across borders with greater ease. As these technologies advance, they hold the potential to break down language barriers in fields like education, healthcare, business, and diplomacy, contributing to a more inclusive and informed global society.

# Chapter IV

# Research methodology

## 1. Data collection

This section discusses the research design of the methods employed for the data collection and data analysis. For this study we selected 230 texts, such as medical prospects, news, articles, interviews that were translated from English into Romanian by Google Translate. The sources of the texts are credible health organizations (WHO, Gavi, UNICEF, etc.). The data were taken purposively from those translations, and they consist of words, phrases, acronyms, full sentences. The research method approached for this study aims to provide information regarding the quality of the Google Translate performance in a pandemic context.

The errors were identified, classified, and examined in order to highlight the need to improve the performance of Machine Translation systems, especially for terms that have increased in frequency since the pandemic.

Analysing a corpus for lexical errors made by Google Translate during the Covid-19 pandemic involves a combination of linguistic analysis, error identification, and understanding the context of the translations. The methodology that was employed involved many stages, as the research design used both quantitative and qualitative methods to display the data. The *first* stage was to collect the corpus. This stage implied gathering over 200 texts translated by Google Translate from English into Romanian. The texts included informative articles, news article, official documents. A diverse range of content of the texts was chosen in order to get a comprehensive view of the context. The *second* stage was to identify the lexical and semantic errors in the texts. We noticed these errors involved inappropriate word choices, mistranslations, inappropriate language use, lack of translation, inadequate interpretation of the meaning of words or phrases, inconsistency in translating keywords of the Covid-19 terminology, translations that deviate significantly from the source meaning or context. The *third* stage was to quantify the errors by using metrics such as Machine Translation evaluators that could measure the performance of Google Translate compared to other systems, in order to provide an overview of the overall accuracy of the translations.

The *fourth* stage involved a qualitative method. The qualitative analysis provided insights into the challenges faced by automated translation systems during the Covid-19 pandemic. It also identified improvements in the performance of Google Translate. The need of both quantitative and qualitative methods is due to the fact that lexical and semantic errors may be more prevalent in certain topics or types of content. For instance, Callison-Burch et al. (2006) critically examine the effectiveness and limitations of the BLEU metric in the context of Machine Translation research. BLEU (Bilingual Evaluation Understudy) had become a widely used metric for automatic evaluation of Machine Translation. The paper explores scenarios where BLEU might be suitable for domain-specific translations, not to mention that certain lexical and semantic errors may involve specific linguistic features or human-like judgements. The combination of approaches is meant to comprehensively assess the lexical and semantic errors in Google Translate during the Covid-19 period. The qualitative approach helps in understanding the reasons behind the errors, whether they stem from cultural factors, idiomatic expressions, or the dynamic nature of the Covid-19 discourse.

Moreover, Covid-19 was a sensitive and challenging topic, especially in a country with mixed economy like Romania, a concept which, according to Mankiw (2020: 132), is an economic system that incorporates aspects of both capitalism (where private enterprise operates with minimal government intervention) and socialism (where the government regulates or owns key sectors to promote social welfare and address market failures). According to Rădulescu and Vasile (2014), the concept of *mixed economy* in Romania reflects a blend of both private and public sector involvement, aiming to balance market efficiency with social welfare. In Romania, the mixed economy emerged prominently post-1989, following the fall of the communist regime and the country's transition towards a market-oriented economy. Rădulescu and Vasile (2014) argue that Romania's mixed economy has facilitated the development of a diverse industrial base, balancing traditional sectors such as agriculture with growing manufacturing and service industries. However, this diversification is not without its challenges. The authors highlight persistent issues even in the translation field. Moreover, the unique economic and political context of Romania's mixed economy, coupled with the limitations of Google Translate's resources and the complexity of the terms involved, contributed to lexical errors during the pandemic.

# 2. Data analysis

## 2.1. The translation of keywords in Covid-19 terminology

The selected keywords displayed below were part of Covid-19 terminology. They were translated from English into Romanian by Google Translate, Chat GPT and Google Gemini. First, they were identified as keywords using Google Trends, based on their frequency of occurrence in documents, articles, and news during the pandemic. Secondly, the translations were included in the chart below in order to emphasize the performance of Google Translate.

*Table (7) Key terminology related to Covid-19*[24]

| English terms | Romanian equivalent given by GT | Romanian terms used by authorities | Romanian equivalent given by Chat GPT | Romanian equivalent given by Google Gemini |
|---|---|---|---|---|
| Acute respiratory stress syndrome (ARDS) | Sindromul de stres respirator acut (ARDS) | Sistemul respirator acut sever (SRAS) | Sindromul de stres respirator acut (ARDS) | Sindromul de detresă respiratorie acută (SDRA) |
| Antibody test | Testul anticorpilor | Test anticorpi | Testul anticorpilor | Test de anticorpi |
| Antigen test | Testul antigen | Test antigen | Test de testare | Test de antigen |
| Asymptomatic | Asimptomatic | Asimptomatic | asimptomatic | asimptomatic |

---

| Background | Fundal | Context general | context | Context |
|---|---|---|---|---|
| Case fatality rate (CFR) | Rata fatalității (CFR) | Evoluția săptâmânala a *cazurilor* confi rmate si *fatalitatea* | Rata mortalității cazurilor (CFR) | Rată de fatalitate a cazurilor (RFC) |
| Color-coding | Cod de culoare | Cod de culori | codificarea culorilor | Cod de culori |
| Community immunity | Imunitatea comunității | Imunitate colectivă(56,200 occurrences) | Imunitatea comunitară | Imunitate de grup |
| Confirmed positive cases | Cazuri pozitive confirmate | Cazuri pozitive confirmate | Cazuri confirmate pozitive | Cazuri confirmate pozitive |
| Conspiracy theories | Teoriile conspirației | Teorie ale conspirației | Teorii ale conspirației | Teorii ale conspirației |
| Curfew | Stare de asediu | Restricții de oră (435 occurrences) | "Curfew" is translated into Romanian as "interdicție de circulație" or "ordonanță de | The most accurate translation of "curfew" into Romanian depends on the context you're using it in, as there are a few |
| | | | restricționare a circulației". | potential options: General term: Restricție de circulație: This is the most general term for "curfew" and means "restriction of movement". More specific terms: Interdicție de circulație: This means "prohibition of movement" and is a stronger term than "restricție". Use it if the curfew prohibits movement entirely. Limită de circulație: Th |

| | | | | is means "limit of movement" and is a less strict term, suggesting restrictions on time or place, rather than a complete ban. |
| | | | | Orare de circulație: This means "schedule of movement" and is useful if the curfew has specific hours during which movement is limited. |
| | | | | Additional notes: |
| | | | | Consider the specific context and intended audience |

| | | | | when choosing your translation. |
| --- | --- | --- | --- | --- |
| | | | | For example, if you're referring to a curfew imposed during a pandemic, "restricție de circulație" or "interdicție de circulație" might be more appropriate. |
| | | | | If you're referring to a curfew for minors, "limită de circulație" or "ore de culcare" (bedtime) might be more suitable |

| Daily deaths | Decese | Decese zilnice | Decese zilnice | Decese zilnice |
|---|---|---|---|---|
| Digital inclusion | Incluziune digitală | Incluziune digitală | Includerea digitală | Incluziune digitală |
| Disaster safety | Siguranța în caz de urgență | Siguranța în caz de urgență | Siguranță în caz de dezastru | Siguranță în caz de dezastru |
| Discharge from the nose | Descărcări din nas | Scurgeri nazale | Scurgeri nazale | Secreții nazale |
| Disease cluster | Cluster de boli | Boli pre-existente | Cluster de boli | Focar de boli |
| Doomscrolling | deranjament | A zapa | doomscrolling | Derularea excesivă a știrilor negative |
| Epidemic | epidemic | Infecție virală | Epidemie | Epidemie |
| False-negative | Fals negativ | Test fals negativ | Fals-negativ | Fals negativ |
| hdroxychloroquin | Hidroxiclorochină | Sulfat de hidroxiclorochină | hidroxiclorochină | Hidroxiclorochină |
| incubation period | incubație | Perioada de incubație | Perioadă de incubație | Perioadă de incubație |
| intensive care medicine | Medicină de terapie intensivă | Anestezie și terapie intensivă (ATI) | Medicină intensivă | Medicină intensivă |
| intubation | intubație | intubare | Intubație | Intubație |
| isolation | izolare | Izolare | Izolare | Izolare |
| lockdown<br><br>"Is lockdown safe?" | carantină<br><br>„Este blocarea sigură?" | Stare de alertă/urgență | Carantină<br><br>Este sigură carantina? | Carantină/stare de urgență/blocare<br><br>1. Este carantina sigură?<br><br>2. Cât de sigură este carantina ? |
| lockdown tiers | Niveluri de blocare | Niveluri de alertă | Niveluri de blocare | Nivelul de închidere |
| lockout | blocare | Șomaj tehnic | Blocare | Blocare |
| mortality rate (MR) | Rata mortalității (MR) | Rata standardizată de mortalitate (RSM) | Rata mortalității (RM) | Rată de mortalitate |
| messenger RNA (mRNA) | ARN Mesager (ARNm) | ARN Mesager (ARNm) | ARN mesager (ARNm) | ARN mesager (ARNm) |
| pandemic | pandemic | pandemie | Pandemie | Pandemie |
| patient zero | Pacient zero | Pacient zero | Pacientul zero | Pacientul zero |

| personal protective equipment (PPE) | Echipament de protecție individuală (EIP) | Echipament de protecție individuală (EIP) | Echipament individual de protecție | Echipament individual de protecție (EIP) |
|---|---|---|---|---|
| physical distancing | Distanțare fizică | Distanțare fizică | Distanțare fizică | Distanțare fizică |
| presumptive positive cases | Cazuri pozitive prezumtive | Cazuri pozitive prezumtive | Cazuri prezumtive pozitive | Cazuri positive prezumtive |
| quarantine | carantină | Carantină/izolare | carantină | carantină |
| remdesivir | Remdesivir | remdesivir | remdesivir | remdesivir |
| reproduction number | Numărul reproducerii | Număr de cazuri | Numărul de reproducere | Număr de reproducere |
| respiratory etiquette | Etichete respiratorii | Coduri de conduită | Etichetarea respiratorie | Etichetarea respiratorie |
| reverse Transcription Polymerase chaine reaction (RT-PCR) | Reacție în lanț a polimerazei cu transcripție inversă (RT-PCR) | Reacția în lanț în timp real a *polimerazei* (PCR în timp real) (36,400 occurrences) | Reacție de polimerizare în lanț cu transcripție inversă (RT-PCR) | Reacția de polimerizare în lanț cu transcriere inversă (RT-PCR) |
| SARS (Severe Acute Respiratory syndrome) | SARS (Sindromul respirator acut sever) | SARS (Sindromul respirator acut sever) | Sindrom respirator acut sever (SARS) | SARS (Sindromul respirator acut sever) |
| shutdown | Închis | carantină | închidere | Închidere completă |

| social bubble | Balon social | Lista de contacte | Bulă socială | Bulă socială |
|---|---|---|---|---|
| social distancing | Distanțare socială | Distanțare socială | Distanțare socială | Distanțare socială |
| social immunity | Imunitatea socială | Imunitate socială | Imunitate socială | Imunitate de grup |
| social tracing | Urmărire socială | Urmărirea contactelor | Urmărirea socială | Urmărirea contacților sociali |
| state of emergency | Stare de urgență | Stare de urgență<br><br>Stare de alertă | Stare de urgență | Stare de urgență |
| supersprea der | Superspread er | un prim contaminat extrem de contagios | superrăspîndi tor | Persoană care transmite virusul mai departe |
| symptomat ic | Simptomatic | simptomati c | simptomatic | Simpomatic |
| total cases per 1 Million population | Total cazuri la 1 milion de locuitori | Total cazuri de infecție la 1 milion de locuitori | Totalul cazurilor la 1 million de locuitor | Total cazuri la 1 million de locuitori |
| underlying medical problems | Probleme medicale subiacente | Co-morbidități | Probleme medicale subiacente | Probleme medicale subiance |
| ventilator | ventilator | Ventilator (VNI) | Ventilator | ventilator |
| videoconference | Conferințe video | Conferințe video | Videoconferință | Videoconferință |
| videofurbishing | Video-amenajare | - | Videodecorare | Modernizare tehnică video |
| work from home | Lucru de acasă | Tele-muncă (98,200 occurrences) | Muncă de acasă | Telemuncă |
| | | Lucru de acasă (20,700,000 occurrences) | | |
| | | Lucru online (62,800,000 occurrences) | | |
| zooming | mărire | Timp petrecut pe Zoom | Zoom-are | Utilizarea aplicației Zoom |

Table (7) illustrates various examples of lexical errors arising from Machine Translation systems, highlighting issues such as literal translations, incorrect alternate meanings of words, mistranslation of idioms, word-by-word translation patterns, and inappropriate lexical selections. These errors often occur when a word is used or understood improperly within a specific context, leading to inaccurate translations. Related literal translations are the errors arising when words or phrases are translated directly from one language into another without considering contextual meanings. This approach often fails to capture the subtleties of the original text, resulting in awkward or inaccurate translations. Wrong Alternate Meanings imply instances where a word is translated based on an inaccurate or less common meaning, leading to misinterpretation. For example, the word "bank" could be inaccurately translated as "a financial institution" instead of "the side of a river," depending on the context. Mona Baker (1992) discusses translation problems that arise due to non-equivalence at the word level. Specifically, she highlights how lexical meanings vary across languages and the importance of understanding the nuances of word usage in context. This is a critical aspect when translators deal with words that have multiple meanings, leading to potential errors in conveying the intended message. Baker's analysis is supported by subsequent research exploring how translators navigate ambiguity, particularly in dealing with languages that express ideas and relationships differently, such as in the case of English-Arabic or English-Spanish translations.

Mistranslated idioms occur when idiomatic expressions are translated literally rather than being interpreted in their intended figurative sense, leading to translations that do not convey the intended meaning or cultural nuance, as Nida (2001: 28) claims: "finding satisfactory equivalents for idioms is one of the most difficult aspects of translating. Literal translation can preserve the surface meaning and form, but often leads to confusion when cultural or contextual differences between languages are not accounted for." In what concerns the word-by-word translation patterns, there are the errors that occur when a translation system translates each word individually without considering the grammatical and contextual relationships between them, often resulting in unnatural or incorrect phrasing. As for the inappropriate lexical selections, the errors are made when the Machine Translation system chooses words that do not fit well within the given context, leading to inaccuracies or confusion. A notable discussion of this issue is presented by Voita et al. (2019: 1203), where they identify "lexical cohesion" as a key area in which Machine Translation systems struggle due to the lack of proper contextual awareness. Their study highlights that such errors can result in misaligned meanings, disrupting the coherence of the translation.

As emphasized by notable figures in lexicography and translation studies, understanding a word's meaning within its context is crucial. Rundell (2020), a prominent lexicographer, notes that lexical errors often result from a superficial grasp of word meanings, highlighting the need for comprehensive context understanding to avoid misinterpretation. Similarly, Pym (2014), a translation studies scholar, discusses how Machine Translation systems can struggle with idiomatic and contextual nuances, leading to errors when translating idioms or culturally specific terms. Both experts stress that words can have meanings that extend beyond their primary definitions, influenced by context, cultural nuances, and idiomatic usage. Inaccurate or incomplete understanding of these factors often leads to errors in translation. This is particularly pertinent in Machine Translation, where algorithms may not fully grasp the subtleties of language use.

The errors illustrated in Table (7) will be methodically categorized and examined in the subsequent section. By analyzing these lexical inaccuracies, we aim to better understand the limitations of current Machine Translation systems and explore strategies for improvement. This involves dissecting how literal translations, misinterpreted meanings, and inappropriate lexical choices contribute to overall translation quality and accuracy. The findings will provide insights into how contextual understanding can be enhanced in Machine Translation systems to reduce errors and improve translation fidelity. This detailed examination will help in identifying specific areas where Machine Translation systems may be improved and guide future developments in the field. Understanding these lexical errors and their underlying causes is essential for advancing Machine Translation technology and achieving more accurate and contextually appropriate translations.

The chart below includes numerous medical terms that can be challenging to translate accurately. Misinterpretations can arise due to lack of medical context, if Google Translate does not recognize the medical context, it might translate terms based on their general meaning, leading to inaccuracies. Moreover, there are subtle differences in meaning between English and Romanian medical terms which can easily be missed by Machine Translation. The chart includes several acronyms (e.g., ICU, Covid-19) that might have been misinterpreted by Google Translate, as we may see in this study. If the system is not aware of the expansion of these acronyms, it could generate incorrect translations. Additionally, there are cultural nuances and linguistic differences that can also contribute to translation errors. As we note in this chapter, Google Translate might have used synonyms that were not entirely accurate in the medical context. For instance, "lockdown" might have been incorrectly translated as "închisoare" (prison) instead of "blocare totală" (total lockdown). There are also terms with similar meanings but different connotations that might have been confused. For example, "quarantine" and "isolation" might have been used interchangeably, leading to confusion.

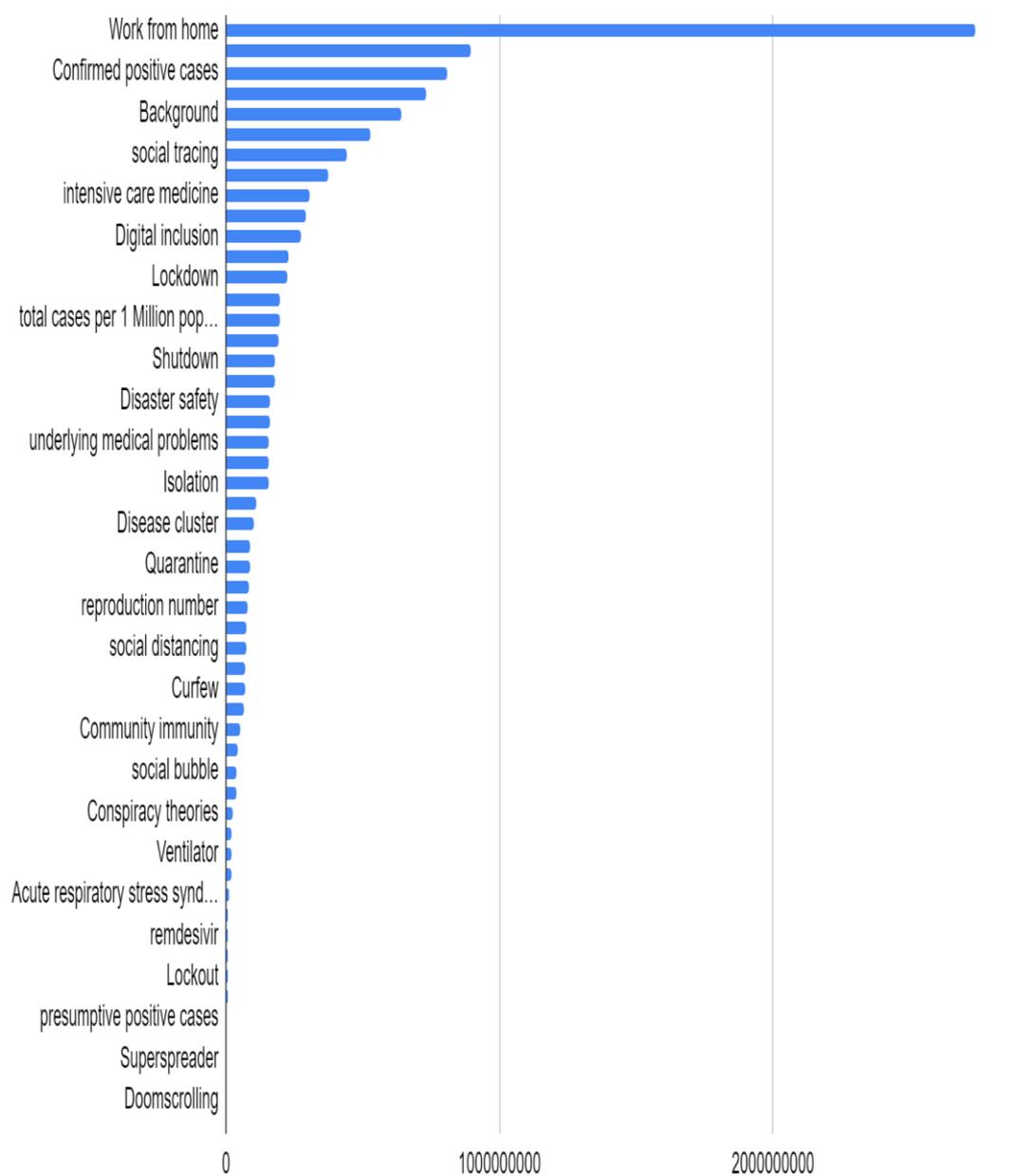

The keywords related to Covid-19

Figure 4. Covid-19 keywords

## 2.2. Formal errors

### 2.2.1. Abbreviations and Acronyms

Abbreviation is considered by Crystal (2003: 120) "a shortened form of a word or phrase, typically created by omitting certain letters or syllables and used to save time or space in writing or speaking. Attila (2022: 379) defines abbreviation as "the umbrella term for all the cases when the original word or phrase is shortened", and acronyms are considered "multi-word sequences starting with initial uppercase letters". Also, abbreviations and acronyms are frequent elements of language in medical texts because they condense complex concepts into symbols. They often carry multiple meanings, adding layers of complexity to Machine Translation. Also underscoring the risks of abbreviation misuse in specialized communication, Brunetti et al. (2007) highlight that such abbreviations can lead to errors in critical contexts, like medication orders, where misunderstanding could have significant consequences.

Moreover, a critical challenge in translation arises when acronyms carry multiple meanings, making context crucial for accurate interpretation. For instance, "CEO" can be translated as "Chief Executive Officer" or "Central European Observatory," requiring a nuanced understanding of the surrounding text for accurate translation. In the data collected, for example, the word CEO was never translated into Romanian as "director general", a term that is commonly used. Despite its advancements, Google Translate can encounter challenges when translating acronyms and abbreviations from English into Romanian in texts related to Covid-19 due to several linguistic and contextual factors, such as lack of contextual understanding, ambiguity, language and cultural nuances, technical complexity, etc. Google Translate operates primarily based on statistical patterns and algorithms rather than true comprehension of language. As a result, it may struggle to interpret the specific context in which an acronym or abbreviation is used, especially in technical or specialized domains like Covid-19. Many acronyms and abbreviations have multiple meanings depending on the context. For instance, "PCR" can stand for "Polymerase Chain Reaction" in a medical context, but it can also represent "put call ratio" in the business field, and Google Translate may not always accurately discern the intended meaning of an acronym or abbreviation. Also, certain acronyms and abbreviations may not have direct equivalents in the target language, or their meanings may vary based on cultural contexts, making accurate translation challenging. Translating technical terms and jargon requires a deep understanding of specialized terminology and domain-specific knowledge. Covid-19-related texts often contain scientific and medical terminology that may be difficult to accurately translate without expertise in the field.

Table (8) categorizes errors related to acronyms and abbreviations, presenting instances encountered during the Google Translate process. Google Translate generally provides accurate translation for standard abbreviations and acronyms, such as the Romanian SDRA (Sindromul de Detresă Respiratorie Acută) for the English equivalent *ARDS* (Acute Respiratory Distress Syndrome). There is a slight difference between the translated result for ACT (Access to Covid-19 Tools). Instead of being translated "Acceleratorul accesului la instrumente penru combaterea Covid-19", Google Translate translates "Acceleratorul de acces la instrument Covid-19". Also, similar to the adequate form is the translation for "Coalition for epidemic preparedness innovations" which is "Coaliția pentru inovații în pregătirea epidemiei (CEPI)" instead of "Coaliția pentru inovații în pregătirea în caz de epidemii."

Several errors in translation arise from a failure to provide an equivalent translation for certain acronyms. Furthermore, the English acronym "CFR" (Case Fatality Rate) is rendered identically in Romanian by Google Translate, rather than using the accurate translation "RM" (rata mortalității), and this might be erroneous and confusing because the well-known acronym "C.F.R" in Romanian means "Căile Ferate Române" (The Romanian Railways – the national railway company of Romania). Similarly, the acronym "ARI" is left untranslated as "ARI" instead of being accurately translated to "SARI" (infecție respiratorie acută severă) in Romanian. Also, Catholic Relief Services (CRS) is not translated at all by Google Translate, as the appropriately equivalent structure "delegația Caritas". These instances underscore the importance of meticulous translation practices to ensure linguistic accuracy and coherence.

Other errors in Google Translate translations can be attributed to certain structures, specifically entities with notable international significance, that remain untranslated. Kuzma et al. (2015: 548) assert that identical abbreviations in the medical field can carry different meanings depending on the specific disease, anatomical context, or procedure being referenced. Also, Muth'im (2018) explores the capabilities and challenges faced by Machine Translation systems when translating abbreviations from one language to another. For instance, in the table above, there is no translation for "Global Polio Eradication Initiative", although Romanian can provide the equivalent: "Inițiativa Globală de Eradicare a Poliomelitei", structure that is used in texts given by the National Institute of Public Health Romania[25]. Similarly, the structure "International Development Bank" is rendered in its original English form, neglecting the adequate translation "Asociația internațională de dezvoltare." Additionally, the acronym "VPD" for "Vaccine-preventable disease" is not linguistically adapted to Romanian, where the adequate translation is "boli prevenibile prin vaccinare", structure that is used by official medical authorities, including on the European Vaccination Information Portal (Portalul European de informații despre vaccin[26]). Furthermore, the untranslated acronym "WHA" contrasts with the accurate Romanian translation "Adunarea mondială a sănătății." This structure is officially used in formal contexts: for example, "Adunarea Mondială a Sănătății este cea mai importantă reuniune globală din domeniul sănătății, la care participă reprezentanții celor 194 de state membre ale OMS. " (The World Health Assembly is the highest decision-making body of the World Health Organization (WHO). It is composed of delegates from all 194 WHO Member States.)[27] These instances underscore the necessity for meticulous translation methodologies, particularly when handling entities with global significance, to ensure linguistic precision and contextual coherence. Another example of confusion arises due to errors of translating the acronyms by Google Translate is the translation of "angajament AMC", which is inaccurately rendered as "angajament de piață în avans" instead of the expected medical term "angajamentul acceleratorului pentru combaterea Covid-19". Similarly, the acronym "APC" (Antigen-presenting cell) is erroneously translated as "angajament de cumpărare anticipată (APC)" instead of the medically accurate term "antigene macrofage." These errors underscore the significance of precise contextual placement of acronyms, particularly within specialized domains like medicine, to ensure the fidelity of translation and prevent potential confusion.

---

| Table (8) Abbreviations and acronyms used during Covid-19 | | |
|---|---|---|
| **Abbreviations and Acronyms** **used in English** | **Abbreviations and Acronyms** **used in Romanian** | **Abbreviations and Acronyms** **given by Google Translate into Romanian** |
| ACT Access to Covid-19 Tools | Acceleratorul accesului la instrumente penru combaterea Covid-19 | Acceleratorul de acces la instrumente Covid-19 |
| Advance Market Commitment (AMC) | Angajament AMC | Angajament de piață în avans |
| AI (Artificial Intelligence) | Inteligență artificială | AI |
| APC Antigen-presenting cell | Antigene macrofage | Angajament de cumpărare anticipată (APC) |
| ARDS Acute Respiratory Distress Syndrome | Sindromul de detresă respiratorie acută | SDRA |
| ARI Acute Respiratory Infection | SARI Infecție respiratorie acută severă | ARI |
| Catholic Relief Services (CRS) | Delegație Caritas | Catholic Relief Services (CRS) |
| CEO | Director general | CEO |
| CDC | Centrul pentru prevenirea și controlul bolilor | Centres for Disease Control and Prevention |
| Centres for Disease Control and Prevention | | CDC |
| CFR<br><br>Case Fatality Rate | RM<br><br>Rata mortalității | CFR |
| Coalition for Epidemic Preparedness Innovations | Coaliția pentru Inovații în Pregătirea în caz de Epidemii | Coaliția pentru Inovații în Pregătirea Epidemiei (CEPI) |
| Covid-19<br><br>Covid-19 is the name of the disease that the novel coronavirus causes. It stands for coronavirus disease 2019 | Covid-19 | Covid-19 |
| FDA (food and drug administration) | Agenția Federală pentru Hrană și Alimente (AFHA) | FDA |
| Global Polio Eradication Initiative (GPEI) | Inițiativa globală de eradicare a poliomielitei | Global Polio Eradication Initiative (GPEI) |
| Intelectual Property (IP) | Proprietate Intelectuală (PI) | Proprietate intelectuală (IP) |
| International Development Association (IDA) | Asociația Internațională de Dezvoltare (AID) | Bank International Development Association (IDA) |
| Intensive Care Unit (ICU) | Unități de terapie intensivă (UATI) | Unități de terapie intensivă (ICU) |
| LMIC<br><br>Covid-19 in low and middle income countries | - | LMIC-uri |

| National Identity Document (NID) | Card de sănătare | NID |
|---|---|---|
| OPV (Olio Polivirus Vaccine) | Poliovirus viu administrat pe cale orală | Poliovirus oral (OPV) |
| PASC (Post-Acute Sequelae of Covid-19) | Sechele post-Covid | PASC |
| PCR Polymerase chain reaction | PCR | PCR |
| PPE Personal protective equipment | EIP Echipament individual de protective | EIP |
| PUI (Patient Under Investigation) | Pacient monitorizat | PUI |
| PCV Pneumococcal conjugate vaccines | VPG Vaccinuri pneumococice conjugate | Vaccinuri pneumococice conjugate (PCV) |
| Pediatric Inflammatory Multisystem (PIMS)colloc | Sidrom Multisistem inflamator pediatric | Sindromul multisistem inflamator pediatric (PIMS-TS) |
| PTSD | Tulburarea de stress post-traumatic | PTSD |
| SARS-CoV-2 Novel coronavirus 2019 is the name of the disease caused by SARS-CoV-2 | SARS-CoV-2 | SARS-CoV-2 |
| SURGE<br><br>Systematic Urgent Review Group Effort | - | Efortului Grupului de Evaluare Urgentă Sistematică COVID-19 (SURGE) |
| UHC<br><br>Universal Health Coverage | - | UHC |
| United Nations Development Progrramme UNDP | Programul Națiunilor Unite pentru Dezvoltare (PNUD) | PNUD |
| Vaccine-preventable disease (VPD) | Boli prevenibile prin vaccinare | VPD |
| WFH<br><br>Working from home | Lucru de acasă<br><br>Muncă la domiciliu | WFH |
| WHA<br><br>World Health Assembly | Adunarea Mondială a Sănătății | WHA |
| WHO<br><br>World Health Organization | OMS<br><br>Organizația mondială a sănătății | CARE |

## 2.2.2. "False friends"

"False friends", also known as "false cognates", are words that appear to be similar in two languages but possess different meanings; they represent a lexical phenomenon worldwide in the process of learning a foreign language, but also in translation studies. According to Lado (1957: 84), they are a "sure-fire trap", by occupying the highest position in the scale of difficulty in translation. These misleading similarities can easily lead to misunderstandings and translation errors, making it essential for translators and language learners to be aware of and attentive to the context in which these words are used. Definitions of "false friends" have included those of Chuquet and Paillard (1989: 224), who define them as "words that are, […] close by form but partially or completely different by meaning". Yaylaci and Argynbayev (2014: 58) define them as "pairs of words which have a similar form and/or pronunciation but different meanings in two languages". Therefore, "false friends" create ambiguity, they are an illusion of familiarity, leading language learners and translators into semantic minefields.

If many researchers argue that "false friends" are typical in situations where two languages with the same typological origins interact, the present context contradicts this view since Romanian and English do not belong to the same language family. Another perspective is supported by Roibu (2014), who suggests considering the fact that the English language has been influenced by Latin, thus explaining the existence of numerous "false friends" based on the interaction between the two languages. In his article titled "False friends – true enemies in language?", Roibu summarizes six causes of this phenomenon: the process of globalization, the lack of systematization in translations in the media (multiple versions of translating the same term), lexicographic simplification, borrowings, semantic changes, formal and sometimes structural and semantic similarities between the two languages.

Errors related to false friends made by MT represent a notable challenge, according to Way et al. (2020), reflecting the difficulties automated systems face in accurately deciphering the intended meaning of words across languages. False friends can lead to substantial inaccuracies in machine-generated translations, generating lexical ambiguity, lack of appropriate cultural context, inaccurate lexical choices, etc. Addressing errors related to false friends in Machine Translation is an ongoing endeavour that requires advancements in natural language processing and a deeper understanding of cross-linguistic nuances. As technology evolves, the quest for more accurate and contextually sensitive Machine Translation continues. Also, the influence of the English language on the lexicon of contemporary Romanian is a significant phenomenon.

Buzea (2017) notes that the current influx of Anglicisms into Romanian represents a compelling and noteworthy subject for linguistic inquiry. Despite numerous languages influencing Romanian over time, none have managed to fundamentally change its Romance essence, a characteristic that also applies to English loanwords in Romanian.

Here are some common errors regarding false friends found in translations related to Covid-19 made by Google Translate:

- *acomodare* – "cazare"

Often, the Romanian word *acomodare* means, according to DEX (2016) "adjustment", but it is used conveying the meaning of the English word, *accomodation*.

- *actual* – "de fapt"

The Romanian word "actual," which means "that exists or happens at present," is used with the sense of "de fapt" from the English paronym "actually."

- *a altera* – "a combina"

The similar form of the Romanian word "a altera," meaning "to deteriorate or change (usually for the worse)," aligns with the English word "to alter," which means "to become different."

- *apropriat*– "adecvat"

The similarity between the English word "appropriate," meaning "suitable, fitting," and the Romanian word "apropiat," which means "close" or "near," leads to confusion.

- *a asista* – "a ajuta"

The Romanian word "a asista" means "to be present." A similar form is found in the English word "to assist," which means "to help someone with something."

- *a asuma* – "a presupune"

The word "a asuma" can create confusion, as the English word "to assume" means "to suppose," while the Romanian word means "to take upon oneself."

- *audiență* – "public"

The Romanian word "audiență" means "audience" in the sense of "meeting" or "interview." However, the English word "audience" has a different meaning, referring to a group of people or spectators. Therefore, the use of the word "audiență" in this sentence is incorrect.

- *a avertiza* – "a promova"

The confusion between "a avertiza" (meaning "to warn" or "to prevent") in Romanian and "to advertise" (meaning "to promote") in English results in errors, as well as between "avertisment" (meaning "warning" or "prevention") and "publicitate" (meaning "the act of making something public" or "advertisement").

- *comun* - "comun"

The false friend error made by Google Translate in translating "common" as "comun" instead of "neutru" or "bine comun" instead of "bine general" reflects a misunderstanding of the nuanced meanings and context-specific usage of the term in Romanian. In English, "common" can convey various meanings, including shared, ordinary, or prevalent. It may refer to something that is widely known, used, or understood by a group of people. However, in specific contexts, "common" can also denote something that is ordinary or unremarkable. In Romanian, "comun" indeed translates to "common" in English. However, the false friend error occurs when the term "common" is used in a specific context where its meaning diverges from the standard interpretation. In the examples provided, "comun" is used to convey the idea of "neutral" or "general" rather than its more common meanings. The translation "comun" by Google Translate fails to capture the context-specific nuances and intended meanings of the term. "Neutru" or "bine general" would be more contextually appropriate translations in Romanian for conveying the idea of "neutral" or "general," respectively.

- *comun* - "banal"

In translating "think of the common cold" as "gândiți-vă la răceala comună," Google Translate commits an error rooted in the use of a false friend. The term "comun(ă)" in Romanian, while sharing a similar form to "common" in English, does not convey the same nuanced meaning in this medical context. "Comună", as an adjective, typically means "shared," "mutual," in Romanian, and while it can denote something that is widespread or frequent, it fails to capture the specific connotation associated with the phrase "common cold." In English, "the common cold" is a fixed expression referring to a mild viral infection affecting the upper respiratory tract, characterized by symptoms like a runny nose, sore throat, and cough. The term "common" here underscores both the ubiquity and the relative mildness of the condition, implying that it is a frequent and typically non-severe illness. Translating "the common cold" directly into "răceala comună" does not preserve this nuanced meaning. Instead, it may imply a general or ordinary type of cold, which is not a conventional way to refer to this illness in Romanian medical or colloquial contexts. The suggested translation, "răceală banală," addresses this issue more effectively. It is more accurate and it avoids the false friend "comună," which does not carry the same medical specificity and common usage as "banală" in the context of describing a minor, widespread illness.

- *drum provocator* – "drum dificil"

The structure "challenging road" presents a false friend error in the translation process, notably when interpreted as "drum provocator" by Google Translate. False friends occur when words or phrases in different languages appear similar but have different meanings or connotations. In this case, "provocator" in Romanian typically conveys the sense of "provoking" or "provocative," which diverges from the intended meaning of "challenging" in English. In English, "challenging road" suggests a road or pathway that presents difficulties or obstacles, requiring effort, skill, or determination to navigate successfully. The term "challenging" denotes a situation or environment that tests one's abilities or resilience, often implying complexity, adversity, or hardship. Conversely, the Romanian term "provocator" carries a different semantic scope, primarily associated with provocation or incitement. While it can denote something that stimulates thought or reaction, it primarily conveys the idea of inciting a response or eliciting a particular emotion or behaviour. The translation "drum provocator" reflects a literal rendering of "challenging road" that aligns with the surface meaning of the English phrase. However, it overlooks the nuanced distinctions in meaning and connotation between the two languages, resulting in a false interpretation of the intended message. A more linguistically accurate and contextually appropriate translation in Romanian would be "un drum dificil" or "un drum care prezintă provocări." These expressions capture the essence of the original phrase by emphasizing the difficulty or adversity inherent in navigating the road, without introducing the connotations of provocation or incitement associated with the term "provocator."

- *eventual* – "în final"

The word "eventual" appears in inappropriate contexts (e.g., "I waited for her impatiently, and eventually, Maria arrived home and told me everything"), being used with the meaning of "finally" that "eventually" has in English.

- *evidență* – "dovada"

The similarity between the Romanian word "evidență," meaning "clarity in perceiving something," and the English word "evidence," which means "proof," leads to errors in automatic translation.

- *guvernanță* – "bună guvernare"

In translating "This also must include improved governance, and ensuring attention to equity and those marginalized and left behind" as "Aceasta trebuie să includă, de asemenea, o guvernanță îmbunătățită și asigurarea atenției acordate echității și celor marginalizați și lăsați în urmă," Google Translate introduces a significant error by rendering "governance" as "guvernanță". In English, the term "governance" pertains to the processes, systems, and practices used to manage and direct an organization or society. It encompasses the broader concept of how decisions are made, implemented, and monitored, ensuring that the organization's or society's objectives are met effectively and equitably. "Governance" is a multifaceted concept used extensively in discussions about public administration, corporate management, and institutional oversight. In Romanian, however, "guvernanță" is not a widely accepted or recognized term to describe this concept. Instead, it misleadingly resembles "guvernantă," which is an entirely different word meaning "governess" – a term used to describe a woman employed to educate and care for children in a private household. This false friend introduces confusion, as it significantly distorts the intended meaning of the original sentence. Using "guvernanță" can mislead the reader into thinking about a role associated with childcare rather than the complex systems of governance. The appropriate Romanian term for "governance" is "guvernanță" itself, but with a proper contextual understanding, as it is relatively new and gaining acceptance primarily in academic and professional circles. However, to avoid ambiguity, a more precise and widely understood alternative would be "buna guvernare" or "managementul guvernării."

- *locație* – "loc"

The misuse of the term "locație" to denote "place" is a prevalent error observed among Romanian speakers, a pattern also replicated by foreign students learning Romanian and relying on Google Translate. This phenomenon can be attributed to the resemblance to the English noun "location," which indeed signifies "place." Additionally, the presence of the word "locație" in Romanian, meaning "rented space," further contributes to this confusion.

- *rezoluție* –"obiectiv"

The false friend in this case refers to the translation of the English word "resolution" into Romanian as "rezoluție" by Google Translate, when the more accurate translation would be "obiectiv". In English, "resolution" can have multiple meanings, but in the context displayed in the corpus, it had the meaning of "setting goals, objectives", "resolution" is often used to express "determination" or "commitment to achieving something specific". On the other hand, in Romanian, "rezoluție" is typically associated with visual display resolution, such as the resolution of a photograph or a computer screen. The more appropriate translation for the goal-setting context would be "obiectiv" in Romanian. This highlights the importance of considering the context and nuances of words, as well as understanding the specific connotations they carry in different languages, to avoid potential misunderstandings or inaccuracies in translation.

- *suport* –"sprijin"

The error regarding false friends in the translation of "support" into Romanian as "suport" rather than "sprijin" found in the texts translated by Google Translate reflects a lexical discrepancy rooted in linguistic interferences between languages. Such errors highlight the complexity of translation processes and underscore the significance of understanding contextual semantics and cultural connotations. In English, "support" encompasses multifaceted meanings ranging from emotional and moral backing to tangible assistance and structural reinforcement. Within the context of Covid-19 discourse, "support" often implies the provision of aid, encouragement, or resources to individuals, communities, or healthcare systems affected by the pandemic. This semantic breadth underscores the intricacies inherent in translating the term into a target language like Romanian. Conversely, "suport" in Romanian primarily denotes a structural or mechanical device designed to uphold or stabilize an object or system. Its connotations revolve around physical reinforcement or assistance, typically within technical or mechanical contexts. While "suport" may align with the notion of providing physical assistance, its semantic scope remains distinct from the broader sociocultural and psychological dimensions associated with the English concept of "support" in the context of Covid-19. The more appropriate translation for "support" in Romanian is "sprijin," which encapsulates the broader spectrum of meanings attributed to the English term. "Sprijin" denotes not only physical reinforcement but also moral encouragement, solidarity, and assistance extended to individuals or communities navigating challenges such as the Covid-19 pandemic. Its semantic flexibility aligns more closely with the diverse dimensions of "support" as understood in English discourse.

### 2.2.3. Borrowings

In the analysis conducted on the corpus, comprising medical texts related to Covid-19, which were translated from English into Romanian, more than 25 instances of non-translation were identified. Almahasees (2022: 124) highlights the challenges of untranslated errors in Machine Translation, particularly between English and Arabic. These errors often arise when the translation systems fail to process certain terms, such as proper names, technical terminology, or cultural references, leaving them untranslated, exploring these issues within the context of widely-used Machine Translation platforms like Google Translate, Microsoft Translator, and Sakhr, emphasizing the need for improved mechanisms to address such deficiencies for more accurate and context-sensitive translation results. Our findings include: "advocacy" instead of "susținere", "bandaid" instead of "plasture", "big data" instead of "volume mari de date", "bioterror attack" instead of "atac biologic", "briefing" for "prezentare succintă", "T helper cells" instead of "ajutătoare", "end-to-end solution" instead of "soluția completă", "cluster" inaccurately rendered for "mulțime", "curfew" instead of "restricții de oră", "hotspots" for "punctele-cheie", "infodemic" for "dezinformare", "know-how" for "soluția", "long Covid" instead of "Covid de lungă durată", "Vaccines pillar" instead of "pilonul de vaccinuri", "pre-print" for "înainte de imprimare", "randmoizat" for "la întâmplare", "proxy protection" for "protejarea proximității", "spike protein" instead of "proteina de vârf", "screening" for "testare și examinare", "service" (in the context of "service workers"), erroneously translated instead of "persoane care lucrează cu publicul", "shibboleths" for "o anumită credință", "smartphone" inaccurately rendered for "telefon smart", and "station" instead of "ordin de a sta în casă."

When Google Translate renders "superspreader" into Romanian as "superspreader" instead of "un prim contaminat extrem de contagios[28]," this leads to a translation error with potential linguistic and contextual implications. The term "superspreader" is a specialized term in epidemiology referring to individuals who are highly infectious and capable of transmitting a disease to a large number of people. In Romanian, translating it as "superspreader" maintains the English term without considering its linguistic adaptation to the target language. The term "un prim contaminat extrem de contagios" conveys the concept of increased risk of disease transmission, aligning with the nuanced understanding of epidemiological terminology in Romanian. Translation accuracy influences the trustworthiness and reliability of information, especially in critical domains like public health. Inaccurate translations may undermine confidence in the source and hinder effective communication strategies aimed at combating Covid-19.

A significant term, overused during the pandemic was "screening". The failure of Google Translate to translate the term "screening" into Romanian, opting instead to leave it untranslated, represents a semantic gap and a missed opportunity for linguistic clarity and precision. This phenomenon can be attributed to several linguistic and contextual factors: "Screening" is a term commonly used in medical contexts to refer to the process of systematically testing and examining individuals for the presence of certain diseases, conditions, or risk factors. It is formally defined as "Screening is the systematic application of a test or enquiry to identify individuals at sufficient risk of a specific disorder to warrant further investigation or direct preventive action, amongst persons who have not sought medical attention on account of symptoms of that disorder."[29] While English speakers may readily understand the term, its meaning may not be immediately apparent to Romanian speakers who are unfamiliar with medical terminology or English loanwords. Therefore, translating "screening" into Romanian as "testarea şi examinarea" would provide a clearer and more precise understanding of the concept for Romanian readers. In Romanian medical discourse, terms like "testarea" and "examinarea" are commonly used to describe medical testing and examination processes. Leaving "screening" untranslated may result in ambiguity and reduce comprehension for Romanian readers, particularly those who are not proficient in English or familiar with medical terminology. Moreover, this borrowing is not limited to the context of the Covid-19 pandemic; rather, it is pervasive across various medical domains. Romanian medical journals, websites, and professional discourse frequently utilize "screening" to describe procedures and tests aimed at the early detection of diseases. For instance, in the official journal of European Union, there are 145 occurrences of this term in the report about the "Council Recommendation of 9 December 2022 on strengthening prevention through early detection: A new EU approach to cancer *screening*, replacing Council Recommendation"[30]. Such usage underscores the integration of global medical practices and terminologies into the Romanian healthcare system, facilitating a standardized approach to medical procedures and enhancing communication among healthcare professionals. The widespread acceptance and usage of "screening" in Romanian medical contexts signify the ongoing interplay between global medical advancements and local medical practices. The translation error made by Google Translate for "plug and play vaccines" from English into Romanian reflects a failure to grasp the nuanced meaning and context of the term within the specific domain of Covid-related discussions. The term "plug and play vaccines" carries a specific technical meaning within the domain of vaccine development, particularly in the context of mRNA vaccines and their adaptability to new viral strains. However, Google Translate may not have the capability to fully comprehend the intricate semantics and technical nuances associated with specialized terminology. Given the complexity and specificity of the term "plug and play vaccines," a more accurate translation would involve conveying the concept of mRNA vaccines that can rapidly adapt to new viral strains, akin to updating software on the same operating system. The suggested translation, "vaccinurile ARNm care pot construi o versiune rapidă pentru o potențială nouă tulpină," captures the essence of this concept more effectively and provides a clearer understanding of the innovative nature of these vaccines in Romanian. Translating the English term "plug and play vaccines" into "vaccinurile ARNm care pot construi o versiune rapidă pentru o potențială nouă tulpină" effectively captures the essence of this concept and provides a clearer understanding of the innovative nature of these vaccines in Romanian. The term "plug and play vaccines" conveys the idea of a versatile, easily adaptable technology that can be quickly modified to address new strains of a virus. However, this phrase might not be immediately understood by a Romanian-speaking audience without a background in technology or biotechnology. Also, the official terms used in Romanian, "vaccinurile plug and play" or "vaccinurile de tip ARNm" do not provide a more accessible and informative explanation that highlights the advanced capabilities of mRNA vaccines, making the concept clearer and more impactful for Romanian readers. Whereas, the description suggested emphasizes that these vaccines can rapidly construct a new version tailored to emerging strains, which is the core innovative feature of the "plug and play" concept. This more detailed translation ensures that the audience comprehends not only the flexibility and speed of these vaccines, but also the groundbreaking nature of mRNA technology, which is crucial in the context of rapidly evolving viral threats.

---

[28] term used in news articles: https://spotmedia.ro/stiri/sanatate/virusul-s-a-transmis-si-la-o-distanta-de-opt-metri-la-abatorul-tonnies-din-germania-studiu.
[29] Department of Health (2000).

---

[30] "Recomandare a Consiliului din 9 decembrie 2022 privind consolidarea prevenirii prin depistare timpurie: O nouă abordare a UE privind *screeningul* pentru depistarea cancerului", In *Jurnalul oficial al Uniunii Europene* (2022).

Another error made by Google Translate is rendering "disease cluster" into Romanian as "cluster de boli" instead of "focar de boli". In translation, a borrowing error occurs when a term or phrase from one language is directly imported into another without appropriate adaptation or translation. In this case, "cluster" is not translated into Romanian, which results in a literal borrowing that does not capture the intended meaning accurately. "Disease cluster" refers to the occurrence of a higher number of disease cases than expected in a particular geographical area or among a specific group of people. It implies a concentration or grouping of disease cases. While "cluster de boli" may seem like a straightforward translation, it lacks the nuanced understanding of the term within the context of epidemiology and public health. "Focar de boli" is a more appropriate translation in Romanian, as it captures the idea of a localized outbreak or concentration of diseases. Translating specialized terminology, especially in fields like epidemiology and public health, requires a deep understanding of the concepts and terminology used in both languages. "Disease cluster" is a term specific to epidemiological investigations and requires a translation that accurately reflects its meaning within that domain. "Focar de boli" not only captures the meaning of "disease cluster" more accurately, but also aligns with the terminology commonly used by authorities and experts in Romanian-speaking regions.

Google Translate may sometimes provide untranslated words in a translation from English into Romanian for several reasons such as language complexity or a non-existent equivalent in the target language. Certain words or phrases may lack direct equivalents or close approximations in Romanian. In such cases, Google Translate might choose to leave the word untranslated rather than provide an inaccurate or misleading translation.

Overall, the decision of Google Translate to leave certain words untranslated in a translation from English into Romanian may reflect an effort to maintain fidelity to the source text while providing users with the most accurate and contextually appropriate translations possible, given the complexities of language and translation. This approach acknowledges that some terms, especially those with no direct equivalents or those heavily loaded with cultural or technical nuances, might be best left in their original form to preserve meaning. For example, as Specia et al. (2010) discuss, the challenge of preserving semantic integrity across languages often necessitates leaving certain terms untranslated to ensure that the core message is conveyed accurately. This practice is particularly relevant in specialized fields such as technology, medicine, and legal contexts, where precise terminology is crucial. By opting to leave some words untranslated, Google Translate acknowledges the limitations of automated systems in fully capturing the intricacies of human language and strives to offer translations that are both faithful to the source and meaningful in the target language.

Our analysis of lexical errors produced by Google Translate revealed over 49 instances of untranslated elements. Of these, 26 were lexical untranslated errors, reflecting the system's inability to effectively interpret or adapt specific terms within the pandemic context. Additionally, 23 out of 35 pandemic-related acronyms were left untranslated, indicating significant limitations in handling abbreviations and acronyms that demand precise contextual comprehension. These untranslated terms underscore the significance of precise terminology and context-aware translation practices, particularly in the domain of medical discourse, where accuracy and clarity are paramount. As Almahasees (2022) notes, Machine Translation systems like Google Translate often struggle with specialized terminology, particularly when handling culturally specific references or technical terms that lack direct equivalents in the target language. This challenge is compounded by the complexity of the source language or the absence of a corresponding concept in Romanian, prompting the system to leave certain terms untranslated rather than risk an inaccurate or misleading rendering (Almahasees, 2022). Such limitations highlight the importance of refining translation algorithms to better account for contextual and domain-specific nuances.

## 2.2.4. Distortions

Some errors such as "nehospitalizați" instead of "nespitalizați", "să mă autoizol" instead of "să mă autoizolez", "mesaje clarate" instead of "mesaje clare" or "proceisem" instead of "poreclite", "păunurile ascunse" for "hidden harms" instead of "răni acute" made by Google Translate when translating Covid-19 related texts from English into Romanian can be examined from linguistic and contextual perspectives. Google Translate's choice of "nehospitalizați" instead of "nespitalizați" reflects a morphological error. While both terms convey the idea of "unhospitalized" or "not hospitalized," "nespitalizați" adheres to the correct morphological structure in Romanian.

In "să mă autoizol," Google Translate omits the verbal morphology, resulting in a grammatical error. The correct form should be "să mă autoizolez," where "autoizolez" agrees with the first-person singular.

"Mesaje clarate" is an inaccurate translation of "clear messages." While "clarate" is a form that does not exist in Romanian, "clare" is the appropriate adjective, conveying the notion of clear. The error here lies in semantic accuracy and idiomatic usage.

In translating the phrase "we're hopeful" as "suntem speranți," Google Translate introduces a distortion error that significantly alters the intended meaning. The term "speranți" is an invented word in Romanian, not recognized in standard usage, which leads to confusion and lacks any established meaning in the language. This fabricated term does not convey the sentiment of "hopefulness" in the same way that the original English phrase does. In English, "hopeful" expresses a state of optimism or expectation about future events or outcomes. Since "speranți" does not exist in Romanian, its use fails to communicate any meaningful sentiment and disrupts the clarity of the translation. The appropriate translation of "we're hopeful" into Romanian would be "avem speranță" or "suntem plini de speranță." "Avem speranță" translates to "we have hope," directly capturing the sense of possessing a hopeful outlook. "Suntem plini de speranță," meaning "we are full of hope," also effectively conveys the emotional state of being hopeful. Both phrases accurately reflect the optimistic anticipation implied by the original English phrase. In conclusion, "suntem speranți" is a distortion error because it relies on a non-existent word that fails to convey the intended hopeful sentiment. Instead, using "avem speranță" or "suntem plini de speranță" provides a clear, accurate translation that faithfully represents the emotional state described by "we're hopeful." This ensures that the translated text remains both meaningful and culturally appropriate.

The translation of words such as *getting, apart, remote (chance), besides, trial, severity, robust* is the result of wrong lexical selections, indicating that the NMT system could not provide an accurate equivalent. A higher level of unintelligibility is represented by words such as *911, v-safe, spike, test vaccine* that are not translated at all. The emergency number *911* is exclusively an American number that has as an equivalent *112* in Europe. In the last decade it has been decided that in films, documents and all sorts of texts the emergency number should be adjusted to the one that is available in the target language.

Overall, the distortions generated by Google Translate in translating Covid-19-related texts from English into Romanian emphasize the intricate challenges of Machine Translation. They highlight the system's struggles with capturing morphological, lexical, and contextual nuances, often leading to significant misrepresentations of meaning.

## 2.3. Semantic errors

## 2.3.1. Confusion of sense relations

This section delves into a critical examination of the errors stemming from confusions made by Google Translate in Covid-related texts translated from English into Romanian. Amidst the urgency and demand for rapid translations, Google Translate's algorithms encounter complexities inherent in language, often resulting in errors that compromise the integrity and clarity of the translated content. One recurring source of error lies in the confusion of meanings, where the system misinterprets the contextual nuances and semantic subtleties of the source text, leading to inaccurate or misleading translations.

- *detecție* – "detectare"

The translation of "detection" as "detecție" instead of "detectare" in a Covid-related text from English to Romanian by Google Translate can be attributed to several linguistic factors, including semantic nuances and domain-specific terminology. In English, "detection" refers to the act or process of identifying or discovering something, often in the context of identifying the presence of a particular substance, condition, or phenomenon. In the context of Covid-19, "detection" specifically pertains to identifying the presence of the virus in individuals through testing or diagnostic methods. In the field of medical and epidemiological discourse, precision in terminology is crucial for accurately conveying scientific concepts and procedures. While "detecție" in Romanian may indeed encompass the general idea of identifying the presence of a signal in electromagnetic waves, its usage may not align precisely with the nuanced meaning of "detection" in the context of virus identification or diagnostic testing. "Detectare" is the more appropriate term in Romanian for expressing the act of identifying or detecting something, particularly in scientific or medical contexts. It is commonly used in discussions related to diagnostic procedures, surveillance, and monitoring of infectious diseases, including Covid-19.

- *distribuție* – "distribuire"

Another error made by Google Translate was for "distribution" as "distribuție" instead of "distribuire" in Romanian within English texts related to Covid-19, that reveals a nuanced discrepancy in the translation process, particularly concerning lexical selection and domain-specific terminology. In the context of epidemiology and public health discourse, "distribution" pertains to the dissemination or allocation of resources, cases, or factors related to the spread of the virus. It involves the systematic arrangement or allocation of elements within a particular context. The term "distribuire" in Romanian encapsulates this sense more accurately, aligning with the specialized terminology used in epidemiological literature. While "distribuție" can convey the idea of distribution in a general sense, its usage in the context of Covid-19 transmission and management might not fully capture the intended meaning. "Distribuire," on the other hand, is more closely associated with the action of distributing or allocating resources, information, or cases in a specific context, which is more appropriate in discussions related to the pandemic. The error underscores the importance of translators possessing a deep understanding of both the source and target languages, as well as familiarity with domain-specific terminology and collocations. Without this expertise, Machine Translation systems like Google Translate may struggle to accurately capture the nuances of specialized discourse, leading to potential mistranslations or misinterpretations.

- *informal* – "nedeclarat"

The error made by Google Translate for "informal" primarily arises from the ambiguity and polysemy of the term, which encompasses multiple meanings and contexts across languages. While "informal" in English generally denotes a relaxed, casual, or unofficial nature, its translation into Romanian as "informal" may convey a similar connotation in certain contexts. However, the error emerges when "informal" is used in the specific context of "informal workers". In English, "informal workers" refers to individuals engaged in employment or economic activities outside the formal sector, often characterized by irregularity, lack of legal protection, and absence of official documentation. Similarly, "informal economy" denotes economic activities that operate outside the formal regulatory framework, often involving unregistered businesses, cash transactions, and informal employment arrangements. In Romanian, the translation "lucrători informali" does not fully capture the nuanced meaning and implications of "informal workers" or "informal economy." While "lucrători informali" suggests individuals engaged in informal or non-official activities, it may not encompass the broader socio-economic context and legal implications associated with "informal workers" in English. A more accurate and contextually appropriate translation in Romanian for "informal workers" could include phrases like "muncitori neangajați oficial" (unofficially employed workers) or "nedeclarați" (undeclared workers). These alternatives better capture the essence of individuals working outside the formal employment sector and convey the notion of unofficial, unregulated employment arrangements.

- *a lista* – "a afişa"

The Google Translate's mistranslation of the verb "to list" into "a lista" in Romanian instead of "a afişa" reflects a discrepancy in the interpretation of the semantic nuances of the context. In English, "to list" can have multiple meanings, including "to enumerate" or "to display in an organized manner." For example, in the context of the sentence "WHO listed Moderna vaccine," the term "to list" implies the action of officially acknowledging or including the Moderna vaccine in a published record or registry. This sense aligns more closely with the notion of acknowledgment or recognition rather than simple enumeration. While "a lista" in Romanian generally corresponds to "to list" in English and can convey the idea of enumeration or creating a list, its usage may not capture the precise connotations of acknowledgment or recognition in the context of official records or publications. "A afişa" might be more appropriate translations in this specific context, emphasizing the act of formally displaying or incorporating the Moderna vaccine into an official listing or registry. The choice between "a lista" and "a afişa" depends on the pragmatic considerations and collocational patterns within the target language. While "a lista" is a valid translation for "to list" in certain contexts, its usage might not fully capture the intended meaning of acknowledgment or recognition in official documentation or listings. In conclusion, the mistranslation by Google Translate highlights the importance of considering semantic precision, collocational accuracy, and linguistic pragmatics in the translation process. Enhancing Machine Translation systems to account for such nuances necessitates ongoing refinement and integration of linguistic knowledge and context-specific understanding to ensure accurate and contextually appropriate translations.

- *lockout* – "blocare" (medical)

In translating the phrase "to go from the laboratory to being delivered to the population" as "vaccinurile să treacă de la laborator până la livrarea populației," Google Translate introduces a semantic error that results in ambiguity and potential misunderstanding. The phrase "livrarea populației" in Romanian is grammatically ambiguous and could be interpreted in multiple ways, leading to confusion about the intended meaning. "Livrarea populației" can semantically suggest "the delivery of the population," which is nonsensical in this context and could imply transporting or distributing the population itself. This interpretation is clearly incorrect and does not align with the intended meaning, which is the distribution of vaccines to the people. The use of "populație" in the genitive case without clear contextual markers can lead to this misinterpretation, as the genitive case can denote possession or a relationship between two nouns, making it unclear whether the population is the recipient or the entity being delivered. The intended meaning of the original phrase is to describe the process by which vaccines move from the development phase in the laboratory to their distribution to the public. A more accurate and less ambiguous translation would be "livrarea către populație," meaning "delivery to the population." This translation clarifies that the population is the recipient of the vaccines, not the object being delivered. The use of "către" (towards) establishes a clear direction of delivery, eliminating any possible confusion regarding the role of the population in this context. Additionally, another appropriate translation could be "distribuția către populație," which translates to "distribution to the population." This term emphasizes the logistical and systematic process of getting the vaccines to the public, fitting well within the context of public health and vaccination campaigns. It removes the ambiguity present in "livrarea populației" by specifying that the action is directed towards the population as the end recipient.

- *lockdown* – "blocare" (medical)

The error in Google Translate's rendering of "lockout" into Romanian as "blocare" rather than "izolare"[31] underscores a fundamental challenge in Machine Translation: the complexity of semantic and contextual nuances embedded within language. While "lockout" can indeed refer to a form of isolation or confinement, particularly in the context of pandemic-related restrictions, its primary meaning pertains to a work stoppage initiated by an employer during a labour dispute. This primary sense of the term might not be readily apparent to Machine Translation systems, especially if they rely solely on statistical patterns without deeper semantic analysis.

---

[31] This term was officially used by the Ministry of Health, and the media (Ministerul Sănătății. 2020. Ordonanța Militară nr. 2 privind măsuri de prevenire a răspândirii Covid-19. Publicată în Monitorul Oficial, Partea I nr. 232 din 21 martie 2020.)

- *(a asuma)mantaua* – " (a-şi asuma) responsabilitatea"

The phrase "un grup de oameni de știință și-a asumat mantaua de a ne oferi un mod neoficial de a descrie aceste noi linii," translated by Google Translate from "a group of scientists have taken up the mantle of giving us an unofficial way of describing these new lineages," contains a semantic error due to the literal translation of the word "mantaua." In Romanian, "mantaua" literally translates to "the mantle" or "cloak," a term primarily used to describe a physical garment or, metaphorically, to denote something covering or enveloping. In this context, "mantaua" does not appropriately convey the figurative meaning intended by the English phrase, which implies taking on a role, duty, or responsibility. The English phrase "to take up the mantle" signifies assuming a position of authority or duty, often one that involves continuing the work of a predecessor or taking on a significant role.[32] The literal translation to "mantaua" fails to capture this nuanced responsibility and instead suggests a misleading image of a cloak or covering. The suggested translation, "responsabilitatea," more accurately reflects the intended meaning. Using "responsabilitatea" aligns with the original context, where the scientists are assuming the duty of providing an unofficial framework for describing new lineages. This translation avoids the confusion caused by the literal interpretation and ensures that the reader understands that the scientists have accepted a significant task, emphasizing their role and commitment in providing a new descriptive method. Thus, "responsabilitatea" is the appropriate term to convey the abstract and figurative nature of the "mantle" in this context, maintaining the semantic integrity of the original phrase.

- *mascare* – "a purta mască"

The discrepancy between the intended meaning of "masking" in English and its translation to "mascare" in Romanian by Google Translate can be attributed to several linguistic factors, including semantic nuances, contextual understanding, and collocational usage. In English, "masking" in the context of Covid-19 refers to the practice of wearing a protective face mask to prevent the spread of the virus. It conveys the idea of using a mask as a barrier or shield against airborne particles. However, in Romanian, "mascare" primarily denotes the act of disguising or concealing one's identity, being typically associated with disguising or camouflaging one's appearance for various purposes, such as theatrical performances, costume parties, or espionage. This carries a distinct semantic connotation unrelated to the intended meaning of wearing a mask for protection against disease transmission. The mistranslation of "masking" to "mascare" highlights the importance of cultural and pragmatic competence in translation, as well as the need for translators to possess a deep understanding of both the source and target languages to convey meaning accurately.

---

[32] Concise Oxford Lingua English Romanian Dictionary (2009), p. 668.

- *sensation* – "sentiment"

The translation error involving the word "sensation" in the phrase "Exista deja senzația că oamenii se pierd" for "Already there was a sense that people were getting lost" exemplifies the pitfalls of false friends in language translation. In this context, "senzația" is incorrectly translated as "sensation" due to its superficial similarity to the English word. In English, "sensation" generally refers to a physical feeling or perception resulting from something that happens to or comes into contact with the body, such as "the sensation of cold" or "a tingling sensation." It is typically used to describe immediate, concrete physical experiences. In contrast, the Romanian "senzația" can also encompass a broader, more abstract sense, including an intuitive or emotional perception of a situation, akin to the English "sense" or "feeling." In the phrase "Exista deja senzația că oamenii se pierd," the intended meaning is an abstract, intuitive perception that people were metaphorically getting lost or feeling disoriented. The false friend error arises because "sensation" in English does not typically convey this broader, intuitive sense and instead suggests a physical experience. Thus, translating "senzația" as "sensation" is misleading and semantically inappropriate. The correct translation would use "sense" to accurately reflect the abstract, emotional context intended in the original Romanian phrase. This ensures that the meaning of an intuitive understanding or perception is preserved, maintaining the integrity and clarity of the original message.

- *suplimentari* – "în plus"

In the given text, the translation of "additional" by "suplimentari" can be identified as a collocation error made by Google Translate, leading to a lack of naturalness or clarity in the translation. In the context of the sentence, "additional" implies an increase or augmentation in the number of children receiving vaccinations beyond a previously mentioned figure or expectation. The term "suplimentari" in Romanian, while related to the concept of supplementation or addition, does not align naturally with the intended meaning in this context.

It may introduce confusion or ambiguity regarding the exact nature of the increase in vaccination coverage. A more suitable translation for "additional" in this context would be "În plus" or "În plus față de aceasta," which effectively conveys the notion of an increase or addition in vaccination coverage without introducing any potential misunderstanding. Therefore, the suggested translation "În plus, peste 760 de milioane de copii au fost vaccinați" captures the intended meaning more accurately and maintains clarity and coherence in the Romanian translation.

- *transmisie* – "transmitere" (medical)

An error made by Google Translate that has four occurrences in 230 texts is in translating "(virus) transmission" as "transmisie" instead of "transmitere" in Romanian. In medical contexts, especially when discussing the spread or transmission of diseases like Covid-19, the term "transmission" refers to the process of the disease being passed from one person to another or from one source to another. In Romanian, "transmitere" more accurately captures this sense of active transfer or conveyance. "Transmisie" is not necessarily inappropriate in Romanian, but it is less commonly used in medical contexts to convey the specific idea of disease transmission. It is not commonly associated with broadcasting or signal transmission in other contexts. "Transmitere" is the more precise term when discussing the spread of diseases. Machine Translation systems like Google Translate often rely on statistical patterns and large corpora of text to generate translations. While they can be effective for general language translations, they may lack the ability to understand context-specific nuances and collocations, especially in specialized fields like medicine. In summary, the error made by Google Translate reflects a lack of sensitivity to the specialized terminology and collocations used in medical discourse. Understanding the subtle differences between terms like "transmisie" and "transmitere" requires a deeper understanding of the domain-specific language and context, which Machine Translation systems may struggle to achieve accurately.

- *zooming* – "mărire"

The error made by Google Translate in translating "zooming" as "mărire" instead of "a petrece timp pe Zoom" stems from a misunderstanding of the context and the intended meaning of the term. The term "zooming" exhibits polysemy, meaning it has multiple meanings or interpretations depending on the context. In English, "to zoom" can refer to various actions such as enlarging an image or rapidly moving or progressing. In the context of the Covid-19 pandemic, "zooming" commonly refers to spending time on the video conferencing platform

Zoom. Translating words and phrases requires more than just matching individual words; it involves capturing the intended meaning and contextual nuances. "Zooming" in the context of the pandemic conveys the idea of engaging in virtual meetings, social gatherings, or educational activities using the Zoom application. The translation "a petrece timp pe Zoom" accurately reflects this specific semantic interpretation. The language used in discussions about Covid-19 often includes colloquial expressions and domain-specific terminology. Translating such language requires familiarity with the subject matter and an understanding of its linguistic nuances. "Zooming" as a colloquial term associated with virtual interactions during the pandemic necessitates a translation that reflects its usage and connotations.

## 2.3.2. Collocation errors

Within the realm of linguistics, collocation errors denote inaccuracies or mismatches in the pairing or combination of words within a given language; they are lexical combinations of words that tend to co-occur frequently and exhibit strong associations based on semantic, syntactic, or pragmatic considerations. According to Hill (2000: 64), collocations should be categorized by their strength into four groups: unique, strong, medium-strength and weak. Strong collocations consist of word pairings that are highly conventional and rarely vary, such as "make an effort," while weak collocations involve combinations that are less fixed and more context-dependent, such as "big decision" or "nice day." By identifying which category an error falls into, you can determine whether the issue stems from a failure to recognize rigid lexical patterns (strong collocations) or from the inability to process broader contextual cues (weak collocations). This insight is particularly valuable when evaluating the performance of a translation system like Google Translate, as it highlights specific linguistic challenges and areas for improvement.

In linguistic analysis, collocation errors manifest when the selection or arrangement of words in a particular context deviate from the expected or idiomatic patterns of usage. They can occur at various linguistic levels, including morphological, syntactic, and semantic dimensions. Semantic collocation errors involve discrepancies in the semantic compatibility or appropriateness of paired words within a given context. According to Zinsmeister and Heid (2004: 311), collocational knowledge is crucial for improving translation quality. Additionally, Melamed (1997) emphasizes the importance of semantic compatibility in translation accuracy. In computational linguistics and natural language processing, collocation errors pose challenges for Machine Translation systems and automated language processing algorithms. According to Zinsmeister and Heid (2004), collocations pose unique challenges because their usage often involves language-specific restrictions and nuances that cannot be easily captured through general syntactic or semantic rules. They for the need to encode collocational constraints lexically, as failing to account for these constraints can lead to unnatural or incorrect translations. *aches* and *pain* - "durere" (medical)

The translation error made by Google Translate, where the words "aches" and "pain" were both translated as "durere" in the general and main presentation of Covid-19 by WHO is a collocation errors. The translation fails to differentiate between the nuanced meanings of "aches" and "pain" in English, resulting in a loss of specificity and clarity in the translated text.

From a linguistic perspective, "While both "ache" and "pain" denote forms of bodily discomfort, they differ in terms of the nature, intensity, and duration of the sensation. "Ache" typically conveys a dull, persistent discomfort, while "pain" encompasses a broader spectrum of sensations, ranging from mild discomfort to intense suffering. The semantic differences between "ache" and "pain" have been explored by various authors in the field of linguistics and medical terminology. Gaston-Johansson (1984) explores the semantic differences between the words "pain," "ache," and "hurt" in the context of pain assessment. His findings indicate that "pain" is often associated with more intense and severe sensations, while "ache" denotes a milder, more persistent discomfort. "Hurt" is used to describe a broader range of unpleasant sensations. By translating both "aches" and "pain" as "durere," Google Translate overlooks the subtle distinctions between these terms, potentially leading to confusion or misunderstanding for readers relying on the translated text.

The lack of differentiation undermines the accuracy and effectiveness of the translation, particularly in conveying the nuances of symptoms associated with Covid-19. A a suggestion would be to use "durere" for "ache" and " suferință" for "pain", so that the translation could capture the nuanced meanings of the source text more accurately. The term "cod de conduită" referring to "respiratory etiquette" in the context of Covid-19 has seen widespread usage in the Romanian language. This term is prominently employed by authoritative entities such as the National Health Insurance House, various official news websites, the General Directorate of Social Assistance and Child Protection, as well as institutions subordinate to the Romanian County Councils. The frequent usage of "cod de conduită" by these institutions highlights its importance and acceptance in public health communication during the pandemic.

- *adverse consequences* – "consecințele adverse"

The structure "consecințele adverse" used by Google Translate as a translation for "adverse consequences" represents a deviation from the more commonly used and idiomatic expression "efectele adverse" in Romanian to convey the meaning of "adverse consequences." In English, it is a concept frequently encountered in medical discourse, while in Romanian, the more established and widely used collocation for this concept is "efectele adverse." A quick Google search illustrates this preference in usage. The term "efecte adverse" yields approximately 1,840,000 results in 0.28 seconds, indicating its widespread acceptance and common usage in Romanian medical and general contexts. In contrast, "consecințe adverse" returns about 2,110 results in 0.38 seconds, demonstrating a significantly lower frequency of occurrence. This disparity underscores the importance of selecting terminology that aligns with prevalent linguistic practices to ensure clarity and comprehensibility in professional and public discourse. The preference for "efecte adverse" is also evident in Romanian medical literature and official healthcare communications, where precision and standardization are paramount. The choice of "consecințele adverse" by Google Translate may stem from its algorithmic processing of language data and corpus analysis, which might not fully capture the nuanced collocational preferences and idiomatic expressions specific to Romanian.

- *after a steep fall-off in prenatal visits – "după o scădere abruptă a vizitelor prenatale"*

The translation "După o scădere abruptă a vizitelor prenatale" provided by Google Translate for "After a steep fall-off in prenatal visits" exhibits a collocation error that arises from both semantic and pragmatic considerations. In this case, "steep fall-off" in English denotes a sudden and significant decline or reduction in prenatal visits, emphasizing the rapidity and magnitude of the decrease. However, the term "scădere abruptă" used in the translation may not fully capture the nuanced connotations of "steep fall-off" and it is not accurate, as the adjective "abruptă" implies a sudden decline associated with hills, slopes, climbing, etc. The phrase "scădere abruptă" may convey a sense of suddenness but may not fully convey the severity or extent of the decline in prenatal visits. Moreover, the choice of words in translations should consider the target audience's linguistic expectations and comprehension levels. The term "vizitelor prenatale" is appropriate for a medical context but may not resonate as effectively with broader audiences or laypersons. While "scădere abruptă" accurately reflects the concept of a sudden decrease, alternative expressions such as "reducere drastică" or "diminuare semnificativă" may better capture the severity and impact of the decline in prenatal visits. Furthermore, considering the broader context of the Covid-19 pandemic, translations should aim to convey the urgency and implications of reduced prenatal care access for pregnant individuals and healthcare systems. In conclusion, while "După o scădere abruptă a vizitelor prenatale" may convey the general idea of a decline in prenatal visits, it may not fully capture the nuances and impact of a "steep fall-off" as intended in the original English expression.

- *background* – "fundal"

The mistranslation of the word "background" into Romanian as "fundal" by Google Translate instead of "context general" can be attributed to several linguistic and contextual factors, including collocation errors and semantic nuances. "Background" is often collocated with "context" in English to denote the broader circumstances or setting surrounding a particular issue or topic. Another suggested and used version of translation would be "istoricul Covid-19" to focus on the history related to the disease. In the context of Covid-19, "background" implies the overall context or general situation related to the pandemic. However, "fundal" in Romanian is more commonly associated with the background in a visual or environmental sense rather than the broader context of a situation. The word "background" carries nuanced meanings depending on the context in which it is used. While it can refer to the visual or auditory backdrop in a scene, it also conveys the underlying or general circumstances that shape a particular situation. In the context of Covid-19, "background" pertains to the overarching context, including factors such as epidemiological data, societal impact, and public health measures.

- *battered governments* – "guverne  bătute"

In translating "battered governments" as "guverne bătute" in the phrase "Dar ce se întâmplă dacă acele guverne bătute se angajează să plătească pentru acest vaccin esențial, care salvează vieți timp de mulți ani ?" Google Translate commits a collocation error that misrepresents the intended figurative meaning of the original term. The English phrase "battered governments" metaphorically refers to governments that were under significant strain or have faced substantial challenges, due to crises such as economic difficulties, social unrest, or public health emergencies like the Covid-19 pandemic. The literal translation "guverne bătute" directly translates to "beaten governments" and implies a physical or literal beating. This term is both unnatural and inappropriate in the context of describing governments. It fails to convey the metaphorical nuance that these governments have been severely tested or pressured by adverse circumstances. The use of "bătute" also introduces a sense of defeat or violence, which is not suitable for conveying the intended meaning of resilience under pressure. A more accurate and contextually appropriate translation would be "guverne puse la încercare." This phrase translates to "governments put to the test" and effectively captures the idea of governments being subjected to severe trials or challenges. "Puse la încercare" is a widely understood expression in Romanian that conveys the notion of undergoing difficult situations or facing significant pressure, without the literal and physical connotations of being beaten. Using "guverne puse la încercare" maintains the figurative sense of the original phrase, indicating that these governments are enduring and navigating through substantial hardships. It reflects the concept of resilience and the ability to commit to long-term initiatives, such as funding a life-saving vaccine, despite being under considerable stress.

- *bioterror attack* – "*atac bioteror*"

In translating "bioterror attack" as "atac bioteror," Google Translate introduces a collocation error that fails to align with the established terminology used in Romanian, particularly in the context of discussions about public health and security during the Covid-19 pandemic. The phrase "bioterror attack" refers specifically to an attack involving the deliberate release of viruses, bacteria, or other biological agents to cause illness or death in people, animals, or plants, typically as a means of terrorism. The term "atac bioteror" is a non-standard and awkward construction in Romanian. The direct translation merges "bio" (biological) and "teror" (terror) into a single phrase, which is not a conventional or recognized term in Romanian language. This translation does not effectively communicate the gravity and specific nature of a biological attack associated with terrorism. During the Covid-19 pandemic and in the broader context of biological threats, the accepted term in Romanian is "atac biologic." This term translates to "biological attack" and clearly conveys the idea of an attack involving harmful biological agents. "Atac biologic" is widely understood and used in Romanian discourse to describe scenarios where pathogens or biological toxins are intentionally used to inflict harm. It succinctly captures the essence of a bioterror attack without ambiguity. Moreover, the use of "atac biologic" is consistent with the terminology employed in Romanian public health and security literature. It accurately reflects the seriousness and specific nature of the threat posed by biological agents used in a terroristic manner. This term is preferred in official communications and aligns with international terminology, facilitating clearer understanding and effective communication in discussions about bioterrorism and public safety.

- *to break the pattern* – "a sparge tiparul"

The collocation error made by Google Translate in translating the phrase "to break the pattern" from English into Romanian as "a sparge tiparele" instead of "a ieși din tipare" reflects a mismatch in collocational preferences and idiomatic usage between the two languages. In English, "to break the pattern" is an idiomatic expression used to describe the act of deviating from established norms, routines, or expectations. It conveys the idea of disrupting or departing from a predictable sequence of events or behaviors. However, in Romanian, the idiomatic equivalent of "to break the pattern" is "a ieși din tipare." This phrase encapsulates the notion of stepping outside conventional boundaries or norms, thus aligning more closely with the idiomatic usage and collocational preferences in Romanian. The translation provided by Google Translate, "a sparge tiparele," while grammatical, lacks the idiomatic resonance and contextual appropriateness of "a ieși din tipare" in Romanian. The verb "a sparge" conveys the idea of physically breaking or shattering, which does not fully capture the intended metaphorical sense of departing from established patterns or norms in this context. To address such errors, Machine Translation systems like Google Translate can benefit from incorporating comprehensive linguistic databases, semantic analysis tools, and contextual disambiguation algorithms to better capture idiomatic expressions and collocational preferences specific to each language.

- *calls* – "apeluri" (puternice pentru acoperirea feței*)*

The phrase "Apeluri puternice pentru acoperirea feței," translated by Google Translate from "Loud calls for face coverings," exemplifies a semantic error that undermines the clarity and accuracy of the original message, particularly in the context of the Covid-19 pandemic. Firstly, the noun "apeluri" for "calls" is not the most appropriate choice. While "apeluri" can mean appeals or calls for action, it often lacks the intensity and urgency conveyed by "loud calls" in English, especially in the context of public health directives. "Solicitări intense," translating to "strong requests" or "intense solicitations," captures the sense of urgent public demand or official directives urging people to take specific actions. This translation better reflects the authoritative and pressing nature of public health messages during the pandemic. Secondly, the term "acoperirea feței" translates literally to "face covering" and is insufficiently informative and contextually inappropriate in the pandemic setting. "Acoperirea feței" might refer broadly to any form of covering for the face, which can be vague and non-specific. In the context of the Covid-19 pandemic, the specific recommendation was for wearing masks to prevent the spread of the virus. Therefore, translating "face coverings" to "purtarea măștii," which means "wearing a mask," provides precise and contextually relevant information. "Purtarea măștii" clearly indicates the recommended action of wearing protective face masks, aligning with public health guidelines and ensuring that the translation conveys the specific and intended preventive measure. Overall, the suggested translation "Solicitări intense pentru purtarea măștii" effectively communicates the urgency and specificity of the original phrase. It aligns with the context of public health messaging during the pandemic, making it clear that the strong appeals were for wearing masks as a protective measure.

- *to catch Covid-19* – "a prinde Covid-19" (medical)

The phrase "pot prinde Covid-19" is a collocation error made by Google Translate in the translation from English to Romanian of the sentence "can I catch Covid-19." Collocation errors occur when words are combined in a way that is not idiomatic or natural in the target language. In this case, "pot prinde Covid-19" is not a common collocation in Romanian to express the idea of contracting or getting infected with Covid-19. While "prinde" can be translated as "catch," its usage in the context of infectious diseases like Covid-19 is not typical in Romanian. The more natural and commonly used collocation in Romanian to express the idea of contracting Covid-19 is "pot lua Covid-19" or simply "pot contracta Covid-19." "a lua" or "a contracta" both convey the meaning of "catch" or "get" in the context of acquiring an illness or infection. Therefore, in the context of discussing the transmission of Covid-19, "pot lua Covid-19" or "pot contracta Covid-19" are more appropriate and linguistically accurate translations in Romanian, conveying the intended meaning in a manner that aligns with native speakers' linguistic expectations and usage patterns.

- *cold chain* – "lanț de frig" (medical)

The collocation error made by Google Translate in translating "cold chain" as "lanț de frig" instead of "lanțul rece" in a text related to Covid-19 reveals a discrepancy in the interpretation of specialized terminology in the medical and public health domain. In English, "cold chain" refers to a logistical system used for transporting and storing temperature-sensitive products, such as vaccines and pharmaceuticals, at low temperatures to maintain their efficacy and integrity. The term encompasses the entire process, from manufacturing and distribution to storage and delivery, emphasizing the importance of temperature control throughout the supply chain. The translation "lanț de frig" accurately conveys the literal meaning of "cold chain" by combining the noun "lanț" (chain) with the noun "frig" (coldness) in Romanian that lacks the idiomatic fluency and specific connotations associated with the term "cold chain" in the medical and logistical context. Conversely, "lanțul rece" is the idiomatic expression commonly used in Romanian to denote the concept of the "cold chain" in the context of pharmaceutical logistics and vaccine distribution, by encapsulating the notion of a controlled and regulated supply chain designed to maintain product integrity through refrigeration and temperature management. The term "cold chain" is officially translated into Romanian as "lanțul rece." This translation is consistently used in various official and media contexts. For instance, a news headline illustrates the usage: "Un stat și-a pus hackerii să atace "lanțul rece" de aprovizionare cu vaccin anti-Covid" ("A state has deployed its hackers to attack the 'cold chain' for Covid-19 vaccine supply")[33]. This example highlights the term's application in discussions related to the logistics and security of vaccine distribution, emphasizing its importance in maintaining the integrity of temperature-sensitive medical supplies. The consistent use of "lanțul rece" in Romanian media and official documents underscores its acceptance and integration into the medical and logistical vocabulary within the context of the Covid-19 pandemic.

- *contact tracers* - "urmăritorii de contact"

The translation "urmăritorii de contact" by Google Translate for "contact tracers" presents a collocation error arising from the misappropriation of terms and the failure to accurately convey the semantic nuances inherent in the original English phrase. The suggested translation "personal de investigare epidemiologică" offers a more contextually relevant and semantically precise rendition, aligning with the terminology utilized by medical authorities during the pandemic. The term "contact tracers" refers to individuals tasked with identifying and tracking individuals who may have been exposed to a contagious disease, particularly during epidemiological investigations. In the context of public health crises such as the Covid-19 pandemic, contact tracing plays a pivotal role in controlling the spread of the virus by identifying and isolating potentially infected individuals. While "urmăritorii" conveys the notion of "followers" or "pursuers," it fails to capture the specific connotation of identifying and tracking individuals within the context of contact tracing efforts. Consequently, the translated expression lacks the precision and clarity required to effectively communicate the intended concept. On the other hand, the suggested translation "personal de investigare epidemiologică" offers a more linguistically and contextually appropriate rendition of "contact tracers." The term "personal" conveys the notion of individuals or personnel, while "investigare epidemiologică" specifically denotes epidemiological investigation, aligning more closely with the operational activities and objectives of contact tracing efforts undertaken by medical authorities during public health crises. Moreover, the suggested translation "personal de investigare epidemiologică" reflects the terminology commonly employed by medical authorities and public health agencies during the Covid-19-pandemic to describe individuals engaged in contact tracing activities. By utilizing terminology consistent with established medical practices and protocols, the suggested translation enhances clarity, accuracy, and comprehensibility in conveying the concept of contact tracing within the target language.

---

[33] https://www.digi24.ro/stiri/externe/mapamond/un-stat-si-a-pus-hackerii-sa-atace-lantul-rece-de-aprovizionare-cu-vaccin-anti-covid-1412007 [Accessed on the 3rd of May 2024].

- *curfew* – "stare de asediu"

The error made by Google Translate in translating "curfew" as "stare de asediu" instead of "restricții de ore" stems from a misunderstanding of the context and the intended meaning of the term. "Curfew" refers to a specific restriction imposed by authorities on the movement of people during certain hours, typically during emergencies or times of crisis such as the Covid-19 pandemic. In this context, "curfew" is a collocation with distinct semantic specificity, implying restrictions on movement during specific hours of the day or night. Effective translation requires a deep understanding of the cultural and contextual nuances associated with the source language and the target language. In the context of pandemic measures, "restricții de ore" accurately captures the essence of "curfew" by conveying the idea of restrictions on movement during specific hours, aligning with the terminology commonly used by authorities and understood by the Romanian-speaking population. Translating "curfew" as "stare de asediu" fails to capture the idiomatic and contextual appropriateness of the term within the Covid-19 context. "Stare de asediu" typically refers to a state of siege or emergency, which may not accurately convey the specific restrictions on movement during the pandemic.

- *depth* – "adâncime"

The collocation error made by Google Translate in translating "depth" as "adâncime" instead of "profunzime" reflects a misunderstanding of idiomatic usage and contextual appropriateness in Romanian. In English, "depth" can refer to various aspects such as the distance from the top surface to the bottom surface of something, the intensity or complexity of a situation, or the profoundness of thought or emotion. It is a versatile term used in a variety of contexts to convey depth in a literal or figurative sense. The translation "adâncime" attempts to capture the literal meaning of "depth" by referring to physical distance or extent from the surface. While "adâncime" is a valid translation in certain contexts, it may not fully capture the nuanced meanings of "depth" in all contexts, especially when used in a metaphorical or abstract sense. On the other hand, "profunzime" is a more contextually appropriate and accurate term in Romanian to convey the figurative or abstract sense of "depth." "Profunzime" implies a deeper level of understanding, insight, or significance, aligning more closely with the nuanced meanings of "depth" in various contexts.

- *discharge from the nose* - "descărcări din nas" (medical)

Google Translate made a collocation errory by translating "discharge from the nose" into Romanian by "descărcări din nas" instead of "scurgeri nazale". In this case, "descărcare" in Romanian typically refers to the discharge or release of energy or a burden, not bodily fluids like mucus from the nose. The term "descărcări din nas" does not accurately convey the intended meaning of "discharge from the nose" in the medical or biological context. "Discharge from the nose" refers to the flow or expulsion of mucus from the nasal passages, which is more accurately captured by "scurgeri nazale" in Romanian, having a specific medical meaning that may not directly align with the typical usage of "descărcare" in Romanian.

- *droplets* - "stropi"

In the context of Covid-19, accurate translation is crucial for conveying the precise nature of virus transmission and the corresponding public health guidelines. The term "droplets" in English, referring to the small particles of liquid expelled when an infected person coughs, sneezes, talks, or breathes, is pivotal in discussions about how the virus spreads. Translating "droplets" as "stropi" in Romanian, as done by Google Translate, introduces a collocation error that can obscure the scientific and medical accuracy of the text. "Stropi" in Romanian generally refers to "splashes" or "sprays," typically used to describe larger, more forceful ejections of liquid, such as splashes of water. This term conveys an image of sizable droplets typically seen in contexts like gardening or spilling liquids, which does not align with the technical understanding of "droplets" in epidemiology and infectious disease control. The use of "stropi" might suggest a misleading notion of size and force, implying larger and more visible particles than those typically responsible for the aerosol and droplet transmission of respiratory viruses like SARS-CoV-2. In the realm of public health and scientific discourse, the term "droplets" often refers to "micro-droplets" or "aerosols," which are significantly smaller particles that can remain suspended in the air and facilitate the spread of the virus over varying distances. The appropriate translation in Romanian would be "picături" or "micro-picături," terms that specifically denote small drops, often microscopic, aligning with the terminology used in medical and epidemiological contexts. "Picături" conveys the correct scale and nature of these particles, emphasizing their small size and relevance in the transmission of the virus.

- *early doses* – "doze timpurii"

The translation of "early doses" as "doze timpurii" in the phrase "Aceste două oferte pot ajuta la asigurarea accesului la doze timpurii pentru cei mai vulnerabili la o scară cu adevărat globală" by Google Translate introduces a collocation error that distorts the intended meaning of the original phrase. In English, "early doses" refers to the initial quantities of a vaccine or medication that become available for administration at the beginning of a distribution timeline. This term emphasizes the timing and prioritization of these doses to reach recipients as soon as possible. In Romanian, the phrase "doze timpurii" directly translates to "early doses," but it fails to convey the specific context of availability and administration intended in the original text. "Timpurii" generally means "early" in a temporal sense but does not adequately express the notion of doses being among the first to be administered or made available. The term lacks the clarity needed to communicate the priority and urgency implied by "early doses" in a public health context. A more precise translation would be "doze care se pot administra mai devreme," which translates to "doses that can be administered earlier." This suggested translation accurately captures the idea of doses being prioritized for early distribution and administration. It clarifies that the emphasis is on the timing of when these doses are given, particularly to the most vulnerable populations, as soon as they are available. This phrase effectively conveys the urgency and prioritization inherent in the concept of "early doses," ensuring that the recipients can benefit from early access to critical interventions. It aligns well with the original intent of ensuring rapid availability and administration of these doses to those in greatest need on a global scale. In addition, another possible translation could be "doze disponibile în faza inițială" or "doze din primele tranșe." "Doze disponibile în faza inițială" translates to "doses available in the initial phase," which clearly denotes the initial availability of these doses. "Doze din primele tranșe" translates to "doses from the first batches," highlighting the prioritization of these early batches for administration.

- *fact checkers* – "verificatorii de fapte"

The translation "verificatorii de fapte" by Google Translate for "fact checkers" represents a collocation error. It is a misappropriation of terms and the failure to accurately capture the nuanced semantic nuances of the original English phrase. The suggested translations "verificator de adevăruri" or "analist de presă" offer more contextually accurate renditions, thereby rectifying the collocation error and aligning with the intended meaning, particularly within the context of addressing fake news during the pandemic. The phrase "verificatorii de fapte" introduces a collocation error primarily due to the literal translation of "fact checkers" into Romanian without considering the specific semantic and functional aspects of the term. While "verificatorii" pertains to the notion of verification or checking, "fapte" more narrowly conveys the concept of events or occurrences rather than the broader notion of factual accuracy or verification as implied by "fact checkers" in the original English phrase.

- *financial expenses* – "cheltuielile financiare"

The translation "cheltuielile financiare" provided by Google Translate for "financial expenses" illustrates a collocation error that can be attributed to both semantic redundancy and lack of idiomatic accuracy. Collocation errors often occur when two or more words with overlapping meanings are used together, resulting in semantic redundancy. In this case, "cheltuielile financiare" combines "cheltuielile" (expenses) with "financiare" (financial), resulting in a redundant expression that conveys the idea of financial expenditures twice. While "cheltuielile" already implies financial transactions or outlays, adding "financiare" adds little semantic value and may even detract from clarity and conciseness. Native Romanian speakers would more commonly use expressions like "cheltuieli" or "costuri" to refer to expenses in general, without the need for the redundant qualifier "financiare." In conclusion, "cheltuielile financiare" may technically convey the idea of financial expenses, but it suffers from semantic redundancy and lacks idiomatic accuracy compared to more concise and idiomatic expressions like "cheltuieli" or "costuri."

- *a firm believer* – "un credincios ferm"

The phrase "a firm believer" in the context "As a firm believer in multilateralism, I was heartened that the United Nations General Assembly adopted a political declaration on pandemic prevention," was inadequately translated by Google Translate as "un credincios ferm."  This translation introduces a collocation error that can significantly alter the intended meaning in Romanian. The term "credincios" in Romanian predominantly refers to someone who is religiously faithful or devout, which is inappropriate in the context of belief in a concept like multilateralism. "Credincios" typically implies a religious or spiritual allegiance, and using "ferm" (firm) to modify it doesn't adapt well to non-religious contexts, making the phrase sound awkward and misleading. The more suitable translation, "un adept convins," better captures the nuance of the English phrase. "Adept" translates to "follower" or "proponent" and is commonly used in Romanian to denote someone who supports or advocates for a particular idea or system, without the religious connotations. The word "convins" translates to "convinced" and conveys a strong, unwavering commitment to the belief or ideology, which aligns with the term "firm" in the original English phrase. Together, "un adept convins" effectively communicates a strong, resolute belief in multilateralism, accurately reflecting the author's intent. This translation avoids the pitfalls of misinterpretation and ensures that the reader understands the unwavering support for multilateralism in a secular and ideological context, as intended in the original English statement.

- *the flow of goods and merchandise* – "fluxul de mărfuri şi mărfuri"

In translating "the flow of goods and merchandise" as "fluxul de mărfuri și mărfuri," Google Translate introduces a collocation error that both repeats and fails to accurately convey the meaning of the structure from the source text. Firstly, there is a redundancy of "mărfuri" (which means "merchandise" or "goods"). This repetition is not only stylistically awkward but also obscures the distinct meaning that each term is intended to convey in the original English phrase. In English, "goods" refers to all types of physical items that are bought and sold, encompassing both consumer products and industrial materials. "Merchandise," on the other hand, often refers more specifically to goods that are bought and sold in the context of trade, particularly retail. In Romanian, a more accurate and nuanced translation of "goods" is "bunuri," which encompasses all types of assets and products available for trade. "Mărfuri," while often used interchangeably with "bunuri," typically refers specifically to commercial goods intended for sale, especially in a business or trade context. Therefore, translating "goods" as "bunuri" and "merchandise" as "mărfuri" maintains the subtle distinction between these two concepts, avoiding redundancy and accurately reflecting the comprehensive scope of the term "goods" alongside the commercial specificity of "merchandise." Another suggestion of translation is would be "fluxul comercial," which translates to "the commercial flow." This phrase effectively emphasizes the overall movement and exchange of goods within a commercial context, encapsulating the buying and selling activities without the need to specify the types of goods involved.

- *gender lens* – "lentilă de gen"

The translation of "gender lens" as "lentilă de gen" in the phrase "De ce este nevoie de o lentilă de gen pentru răspunsul la Covid-19" by Google Translate introduces a significant collocation error. The term "gender lens" in English is a metaphorical expression used in the context of policy-making, analysis, and response strategies to emphasize the importance of considering gender-specific impacts and perspectives. It suggests a way of viewing issues, decisions, or actions that takes into account the different needs, roles, and experiences of various genders. Translating "gender lens" literally as "lentilă de gen" in Romanian creates an awkward and confusing phrase that fails to convey the metaphorical and analytical nuance of the original term. In Romanian, "lentilă" typically refers to an optical lens used for vision or photography, and its use in this context does not make immediate sense. It does not evoke the intended metaphor of examining or addressing issues through the specific considerations and impacts of gender differences. A more appropriate translation for "gender lens" in Romanian is "perspectivă de gen." The phrase "perspectivă de gen" translates to "gender perspective" and effectively captures the idea of analyzing or responding to issues by considering gender-related factors. This term is widely understood and used in Romanian to describe approaches or viewpoints that take into account gender-specific differences and inequalities. Using "perspectivă de gen" aligns with the metaphorical intent of "gender lens," which is to look at and address challenges, such as the Covid-19 pandemic, by incorporating gender-sensitive analysis and responses. This term ensures clarity and precision, allowing the reader to understand that the discussion is about the necessity of considering gender impacts in the response to the pandemic, rather than any literal use of a physical lens. Furthermore, "perspectivă de gen" is consistent with the language used in Romanian academic, policy-making, and advocacy contexts when addressing gender issues. It resonates with established discourse on gender equality and the incorporation of gender analysis in various fields, making it the suitable choice for translating "gender lens."

- *to get around the issue* – "a ocoli o problemă"

The phrase "Pentru a ocoli această problemă" appeared in a text related to Covid-19 and was generated by Google Translate. While grammatical, its usage reflects a collocation error stemming from a lack of idiomatic fluency and contextual appropriateness within the medical discourse. In Romanian, "pentru a ocoli această problemă" translates to "to get around this issue" in English. Nevertheless, for expressing the notion of avoiding or circumventing a problem more aptly, alternatives such as "a evita această problemă" or "a rezolva această problemă" would be more contextually fitting.

- *global public* – "publicul global"

The error made by Google Translate in translating "global public" as "publicul global" instead of "publicul larg" reflects a misunderstanding of idiomatic usage and contextual appropriateness in Romanian. In English, "global public" refers to the worldwide audience or the collective population of the globe. It denotes people from various countries and regions who are interconnected by global issues, events, or phenomena. The term "global public" emphasizes the breadth and inclusivity of the audience, encompassing individuals from diverse backgrounds and locations. The translation "publicul global" attempts to capture the literal meaning of "global public" by combining the adjective "global" with the noun "public." While grammatical, it lacks the idiomatic fluency and contextual appropriateness required to convey the intended meaning effectively in Romanian. On the other hand, "publicul larg" is a more contextually appropriate rendering in Romanian. This expression translates to "the general public" or "the wider public," conveying the notion of a broad and inclusive audience without specifying the global scope explicitly. "Publicul larg" is commonly used in Romanian to refer to the general population or audience, encompassing people from various demographics and backgrounds.

- *good* – "bun"

The translation of the word "good" as "bun" in the phrase "Dar, așa cum stau lucrurile, nu avem date bune pentru a fi siguri despre acest lucru" for "But as things stand, we don't have good data to be certain about this" exemplifies a semantic error that disrupts the intended meaning. In this context, the term "good" in English is used to describe the quality and reliability of data, implying that the data is of sufficient accuracy, comprehensiveness, and trustworthiness to form a conclusive opinion. In Romanian, while "bun" can be translated to "good" and is used in many contexts to denote something of satisfactory quality or favourable condition, it does not capture the full nuance of "good" in relation to data quality. "Bun" is often associated with general positive attributes and may suggest something being pleasant or of decent quality without necessarily implying thoroughness or correctness. Therefore, using "bun" to describe data might lead to ambiguity, suggesting that the data is only favourable or agreeable rather than methodically sound and reliable. A more precise translation in this context would be "corect" or "complet." "Corect" translates to "accurate" or "correct," which directly addresses the necessity for data to be precise and reliable in order to draw certain conclusions. It emphasizes the validity and correctness of the information. On the other hand, "complet" translates to "complete," underscoring the need for data to be thorough and comprehensive, covering all necessary aspects to support certainty in conclusions.

Using "corect" or "complet" in place of "bun" aligns more closely with the intended meaning in the original English sentence. These alternatives accurately convey the requirement for data to be of high quality in terms of accuracy and comprehensiveness, which is essential for certainty and sound judgment. Therefore, translating "good" as "bun" fails to adequately reflect the specific qualities of data being discussed, leading to a semantic misinterpretation. The correct usage of "corect" or "complet" preserves the nuance and clarity, ensuring the message's integrity is maintained in the Romanian translation.

- *the hammer approach* - "abordarea ciocanului"

The error noted in the translation of "the hammer approach" from English into Romanian as "abordarea ciocanului" by Google Translate represents an instance of collocationerror. In the original English phrase, "the hammer approach" metaphorically refers to a forceful or aggressive method of tackling a problem or situation. The term "hammer" symbolizes strength, directness, and decisiveness in this context. However, the translation "abordarea ciocanului" in Romanian lacks the idiomatic coherence and semantic precision of the original phrase. While "ciocan" translates to "hammer" in Romanian, the combination "abordarea ciocanului" fails to resonate with native speakers and lacks meaningful usage in Romanian discourse. A more contextually appropriate and linguistically accurate translation in Romanian could involve expressions like "abordarea lipsită de tact" (the approach lacking tact) or "abordarea agresivă" (the aggressive approach). These alternatives capture the essence of forcefulness and directness conveyed by "the hammer approach" while aligning more closely with idiomatic usage and semantic conventions in Romanian.

- *health facility* - "unitate de sănătate"

The translation of "health facility" as "unitate de sănătate" in the phrase "Acum mă tem că am putea primi Covid-19 și de la o unitate de sănătate" by Google Translate introduces a collocation error that misrepresents the specific context intended by the original term. The phrase "health facility" in English typically refers to any establishment that provides medical care, such as hospitals, clinics, or outpatient centers. It encompasses a wide range of institutions where healthcare services are delivered. In Romanian, the term "unitate de sănătate" translates literally to "health unit," which does not adequately convey the breadth and specificity of "health facility.". Also, it is not commonly used in Romanian and may be interpreted as a generic or ambiguous term, lacking the precision required to describe a place where medical services are provided. This can lead to confusion, as it does not immediately evoke the image of a medical establishment where one might receive treatment for illnesses such as Covid-19. A more accurate translation would be "instituție medicală," emphasizing the formal and established nature of the facility providing healthcare services.

- *information, misinformation, and disinformation* – "informare, dezinformare și dezinformare"

The translation provided by Google Translate, "informare, dezinformare și dezinformare," is indeed inaccurate and fails to capture the nuances and distinctions present in the original English phrase "Information, misinformation, and disinformation.". Linguistically, the error arises due to the semantic imprecision and syntactic mismatch between the source language (English) and the target language (Romanian). While "misinformation" and "disinformation" are distinct terms in English, denoting different types of false or misleading information, the Romanian translation fails to differentiate between them adequately. The term "dezinformare" is used twice in the translation, implying a lack of differentiation between "misinformation" and "disinformation," which are conceptually distinct. The failure of Google Translate to accurately render the original phrase reflects a deficiency in understanding the specific terminology and context associated with the topic of the spread of information during the Covid-19 pandemic. In scholarly discourse and public health communication, precise terminology is crucial for distinguishing between different types of false information and understanding their impact on public perception, behaviour, and policy responses. To rectify this translation error and provide a more accurate rendition of the original phrase, alternative translations such as "Informație, dezinformare și dezinformare intenționată" or "Informație, informație greșită și informație falsă răspândită intenționat" would be more appropriate. These translations incorporate the distinction between "misinformation" ("informație greșită") and "disinformation" ("informație falsă răspândită intenționat"), thereby aligning more closely with the intended meaning and contextual nuances of the original English phrase. Furthermore, the suggested translations include additional clarifications such as "greșită" (incorrect) and "falsă" (false), which serve to enhance precision and clarity in conveying the difference between misinformation and disinformation.

- *to keep the baby* – "a ține copilul"

The error made by Google Translate in translating "to keep the baby" from English into Romanian as "a ține copilul" instead of "a păstra copilul" lies the complexity of lexical semantics and collocational preferences across languages. In English, the phrase "to keep the baby" conveys the idea of caring for, nurturing, or maintaining the well-being of a child, often within a familial or caregiving context. The word "keep" in this context carries connotations of providing care, protection, and support for the child's welfare. In Romanian, however, the verb "a ține" primarily denotes the act of holding or grasping, with nuances extending to retaining or keeping possession of something physically. While "a ține" can convey the idea of physically holding onto an object or person, its usage in the context of childcare or parental responsibility is less common and may not fully capture the intended meaning of providing ongoing care and support for the child's welfare. On the other hand, "a păstra" in Romanian aligns more closely with the semantic nuances of "to keep" in English within the context of childcare. "A păstra" encompasses the notions of preserving, safeguarding, and maintaining something in a state of security or well-being. Thus, "a păstra copilul" better captures the intended meaning of nurturing and safeguarding the child's welfare over time. Furthermore, the lexical choice of "a păstra" over "a ține" reflects collocational preferences and idiomatic usage within the Romanian language. "A păstra copilul" is a more idiomatic and contextually appropriate expression for conveying the idea of caring for and nurturing a child, whereas "a ține copilul" may evoke more literal or physical interpretations of holding onto the child.

- *let's flatten the infodemic curve* – "Să aplatizăm curba infodemică"

The phrase "Să aplatizăm curba infodemică" in Romanian represents an instance of collocation error stemming from a literal translation approach, particularly attributable to the tool Google Translate. This collocation error arises due to the incongruity between the semantic nuances of the source phrase "Let's flatten the infodemic curve" and its resultant Romanian counterpart. The term "infodemic," a portmanteau of "information" and "epidemic," denotes the rapid dissemination of large volumes of information, often inaccurate or misleading, during a crisis or pandemic. In the context of public health, the phrase "flatten the curve" has gained prominence, particularly during the Covid-19 pandemic, signifying the mitigation of the virus's spread to prevent overwhelming healthcare systems. When analysing the term "flatten the infodemic curve," it encapsulates the notion of mitigating the spread and impact of misinformation or disinformation, akin to the epidemiological concept of flattening the curve for infectious diseases. However, the Romanian phrase "Să aplatizăm curba infodemică" lacks the specific connotation of addressing misinformation and inaccuracies within the information ecosystem. In light of the nuanced objective of combating misinformation, the suggested translation "Haideți să reducem valul de dezinformare" embodies a more linguistically and semantically accurate rendition. The term "reducem" (reduce) encapsulates the notion of mitigating or minimizing the propagation of misinformation, aligning more closely with the intended objective of flattening the infodemic curve. Moreover, "valul de dezinformare" (the wave of misinformation) effectively emphasizes the pivotal role of combating misinformation in ameliorating the impact of the infodemic.

- *long grass* – "abordarea cu iarbă lungă"

The collocation error made by Google Translate in translating the expression "the long grass approach" as "abordarea cu iarbă lungă" instead of "abordare la firul ierbii" reflects a misunderstanding of idiomatic usage and contextual appropriateness in the target language. In English, "the long grass approach" is an idiomatic expression that metaphorically refers to a strategy or method characterized by avoidance, delay, or indirectness in dealing with an issue or problem. The phrase alludes to the idea of navigating through long grass, which can be challenging, time-consuming, and fraught with uncertainty. The translation "abordarea cu iarbă lungă" attempts to capture the literal meaning of "long grass" but fails to convey the intended metaphorical sense of the expression. While grammatical, it lacks idiomatic fluency and does not effectively communicate the nuanced connotations associated with the original phrase. On the other hand, "abordare la firul ierbii is a more contextually appropriate rendering in Romanian. This expression translates to "approaching the grass blade by blade (in detail)," suggesting a meticulous or thorough examination of the situation, akin to scrutinizing each blade of grass individually. It captures the essence of the original expression while conveying the notion of detailed analysis or scrutiny.

- *(a asuma)mantaua* – " (a-şi asuma) responsabilitatea"

The phrase "un grup de oameni de ştiinţă şi-a asumat mantaua de a ne oferi un mod neoficial de a descrie aceste noi linii," translated by Google Translate from "a group of scientists have taken up the mantle of giving us an unofficial way of describing these new lineages," contains a semantic error due to the literal translation of the word "mantaua." In Romanian, "mantaua" literally translates to "the mantle" or "cloak," a term primarily used to describe a physical garment or, metaphorically, to denote something covering or enveloping. In this context, "mantaua" does not appropriately convey the figurative meaning intended by the English phrase, which implies taking on a role, duty, or responsibility. The English phrase "to take up the mantle" signifies assuming a position of authority or duty, often one that involves continuing the work of a predecessor or taking on a significant role.[34] The literal translation to "mantaua" fails to capture this nuanced responsibility and instead suggests a misleading image of a cloak or covering. The suggested translation, "responsabilitatea," more accurately reflects the intended meaning. Using "responsabilitatea" aligns with the original context, where the scientists are assuming the duty of providing an unofficial framework for describing new lineages. This translation avoids the confusion caused by the literal interpretation and ensures that the reader understands that the scientists have accepted a significant task, emphasizing their role and commitment in providing a new descriptive method. Thus, "responsabilitatea" is the appropriate term to convey the abstract and figurative nature of the "mantle" in this context, maintaining the semantic integrity of the original phrase.

- *medical worker* – "lucrător de birou/ medical"

The mistranslation of "office worker" or "medical worker" by Google Translate into "lucrător de birou" or "lucrător medical" reflects a discrepancy in semantic understanding and collocational accuracy. In English, "office worker" typically refers to an individual employed in an office setting, performing administrative, clerical, or managerial tasks. Similarly, "medical worker" denotes someone engaged in healthcare-related professions, including doctors, nurses, technicians, and other healthcare personnel. The mistranslation occurs because the terms "lucrător de birou" and "lucrător medical" in Romanian do not precisely convey the intended meaning of "office worker" or "medical worker." While "lucrător" translates to "worker" or "employee," it lacks specificity regarding the nature of the work or the professional context. A more appropriate translation for "office worker" in Romanian would be "angajat de birou" or "funcționar de birou," which explicitly denote someone employed in an office environment and engaged in administrative or clerical duties. Similarly, for "medical worker," a more accurate translation could include terms like "personal medical" or "cadru medical," which encompass a broader range of healthcare professionals and convey the specialized nature of their work within the medical field.

---

[34] Concise Oxford Lingua English Romanian Dictionary (2009), p. 668.

- *a nose or throat swab* – "tampon pentru nas și gât" (medical)

The translation of the English structure "a nose or throat swab" made by Google Translate into Romanian as "un tampon pentru nas și gât" for a text about tests for coronavirus represents a collocation error due to the inadequacy in capturing the specific medical terminology and nuanced usage within the context of Covid-19 testing procedures. In English, "a nose or throat swab" refers to the process of collecting samples from the nasal cavity or throat using a specialized instrument or device, typically a cotton swab or similar implement. The term "swab" denotes the action of taking a sample or specimen for diagnostic purposes, commonly used in medical and laboratory settings. The translation "un tampon pentru nas și gât" attempts to convey the idea of collecting samples from the nose and throat, which aligns with the intended meaning of the English phrase. However, the term "tampon" in Romanian, particularly in the medical field, typically refers to a piece of cotton material used to absorb bodily fluids, such as menstrual blood or other types of bleeding, rather than an instrument for sample collection. The collocation error becomes apparent when considering the specific terminology used in Covid-19 testing protocols. In the context of coronavirus testing, the correct term for the device used to collect samples from the nose and throat is not "tampon," but rather "tampon nazofaringial." This term accurately reflects the instrument's design and function, emphasizing the use of a cotton-tipped swab for sample collection. Furthermore, the term it is used in official texts, such as the prospect of Covid-19 antigen rapid test "Dispozitiv de testare rapid antigen Coronavirus este un test de diagnostic in vitro pentru detectarea calitativa a antigenilor noului coronavirus in *tampon nazofaringian*, utilizand metoda imunocromatografica rapida."[35]

---

[35] The Coronavirus Rapid Antigen Testing Device is an in vitro diagnostic test for the qualitative detection of new coronavirus antigens in swab specimens, using the rapid immunochromatographic method.

- *oncology pipe* – "conductă de oncologie" (medical)

The translation provided by Google Translate, "conductă de oncologie," for the English phrase "oncology pipeline" is considered a collocation error due to semantic incongruence and incorrect collocational usage. In Romanian, "conductă" typically denotes a conduit, pipe, or channel used in plumbing or engineering contexts, conveying the idea of a physical pathway for fluids or materials. In the bilingual dictionary, "pipeline" is translated by "TECH. conductă de petrol; to be in the ~ FIG. a fi în curs de elaborare."[36] It belongs to a distinct lexical field associated with infrastructure, engineering, or transport. Conversely, "oncology pipeline" in English refers metaphorically to the developmental process or progression of new drugs, treatments, or therapies within the field of oncology. It denotes the sequence of research, development, testing, and approval stages for oncological interventions, often within pharmaceutical or medical contexts. The collocation error arises from the mismatch between the intended meaning of "oncology pipeline" and the literal interpretation of "conductă de oncologie." While "conductă" may convey the notion of a pathway or channel, its association with plumbing or engineering does not align with the abstract, metaphorical sense of "pipeline" in the context of medical research and development. A more appropriate translation for "oncology pipeline" in Romanian would involve conveying the concept of the developmental program or process within the oncological domain. A term such as "programul de dezvoltare din domeniul oncologiei" or "fluxul de cercetare în oncologie" would better capture the intended meaning and align with the collocational preferences within the medical and pharmaceutical discourse. In Romanian medical terminology, the word "conducta" is commonly used on official medical websites to refer to various types of conduits or channels within the body. However, the specific formation "conducta de oncologie" is not a standard term in medical language. Instead, more precise terms are preferred to describe specific oncology-related pathways or procedures. This distinction highlights the importance of using contextually appropriate and medically accurate terminology in professional healthcare communication. For instance, the term "conducta biliară" (bile duct)[37] is widely recognized and used, whereas "conducta de oncologie" lacks the specificity required for effective medical communication.

---

[36] Concise Oxford Lingua English Romanian Dictionary. (2009). p.485.
[37] "Enzimele se eliberează în tuburi mici, numite *conducte*, care în cele din urmă se golescîin canalul pancreatic." https://www.sfatulmedicului.ro/Cancer-hepatobiliar/cancerul-pancreatic-cea-mai-agresiva-forma-de-cancer_17573 [Accessed on the 20th of May 2024].

- *overwhelming* – "copleșirea" (sistemelor de sănătate)

The translation of the phrase "meaning avoiding cases from spiking and overwhelming health systems" to "ceea ce înseamnă evitarea cazurilor de la creșterea și copleșirea sistemelor de sănătate" involves collocation error with the word "copleșirea." In Romanian, "copleșirea" generally translates to "overwhelming" in the sense of being emotionally or mentally overpowering, which can convey a sense of being flooded or overcome by emotions, pressures, or responsibilities. However, in the context of public health and system capacities, this translation does not fully capture the intended meaning of a system becoming overburdened or excessively strained. The phrase "overwhelming health systems" refers to the situation where healthcare systems are unable to cope with the surge in demand due to a spike in cases, leading to their functional capacity being surpassed. This situation necessitates a term that conveys a systemic overload rather than an emotional or subjective sense of being overwhelmed. The suggested term "suprasolicitarea," meaning "overloading" or "overburdening," is a more precise and contextually appropriate choice. "Suprasolicitarea" directly refers to the state of systems or organizations being excessively stressed or burdened beyond their operational limits. It aligns with the idea of healthcare systems being pushed beyond their functional capacity due to a sudden increase in cases. This term is often used in technical and organizational contexts to describe scenarios where the demand exceeds the available resources or capabilities, making it particularly apt for describing the strain on healthcare systems during a public health crisis. Using "suprasolicitarea" in place of "copleșirea" ensures that the translation accurately reflects the nature of the pressure on health systems, emphasizing their potential to be overloaded and unable to meet the demand. This choice maintains the precision and clarity of the original English phrase, effectively communicating the critical concept of healthcare system capacity and the risks of it being exceeded during spikes in cases. Therefore, "ceea ce înseamnă evitarea cazurilor de la creșterea și suprasolicitarea sistemelor de sănătate" is the more appropriate translation, capturing both the urgency and the specific systemic challenges described.

- poor spelling and grammar – "*ortografie și gramatică slabe*"

The translation "ortografie și gramatică slabe" by Google Translate for "poor spelling and grammar" represents a collocation error resulting from the misalignment of terms and the failure to convey the nuanced semantic and syntactic structure of the original English phrase. The phrase "ortografie și gramatică slabe" suffers from a collocation error primarily due to the use of "slabe" (weak) in conjunction with "ortografie și gramatică." While "slabe" conveys the notion of weakness or deficiency, its application in this context lacks the specificity required to encapsulate the concept of errors in spelling and grammar. The term "slabe" is more broadly applicable and may not distinctly convey the linguistic deficiencies associated with poor spelling and grammar. The suggested translation "ortografie și gramatică defectuoase" offers a more accurate and contextually fitting rendition, rectifying the collocation error and providing a linguistically precise translation.

- *to relax their lockdowns* – "să relaxeze blocajele"

The translation of the phrase "With a number of countries in Asia, such as South Korea, China and Singapore, now starting to relax their lockdowns" to "Cu un număr de țări din Asia, cum ar fi Coreea de Sud, China și Singapore, acum încep să-și relaxeze blocajele" by Google Translate exhibits a notable collocation error with the structure "să-și relaxeze blocajele." The term "blocajele" translates literally to "blockages" or "obstructions," which is a non-standard and awkward way to refer to the systematic lockdown measures implemented to control the spread of Covid-19. This choice does not accurately reflect the organized and structured nature of lockdown policies and is not commonly used in the context of public health regulations. In Romanian, the appropriate terms for "lockdowns" in the context of public health measures are "restricții" or " izolare." "Restricții" translates to "restrictions," which more accurately conveys the regulatory actions taken to limit movement and social interactions during a pandemic. "Restricții de izolare" specifically refers to "isolation restrictions," highlighting the aspect of limiting people's movement and interaction to contain the virus spread. Moreover, the verb "a relaxa" in the context of "blocajele" (blockages) is misleading because "relaxing blockages" does not convey the process of systematically reducing restrictions. The suggested translations "să ridice restricțiile" or "să relaxeze restricțiile de izolare" offer more precise and idiomatically correct alternatives. "Să ridice restricțiile" translates to "to lift the restrictions," indicating a formal reduction or removal of the constraints. "Să relaxeze restricțiile de izolare" translates to "to ease the isolation restrictions," which captures the gradual and controlled reduction of lockdown measures. These suggested translations align with the intended meaning of the original phrase, emphasizing the methodical and phased approach to reducing the lockdown measures. They provide clarity and maintain the formal and technical nature of the discussion about public health policies. Hence, the translation "să-și relaxeze blocajele" fails to convey the structured process of easing restrictions and could lead to confusion. In contrast, "să ridice restricțiile" or "să relaxeze restricțiile de izolare" are precise, contextually appropriate, and enhance the clarity and accuracy of the translated text.

- *respiratory droplets* – "picături respiratorii" (medical)

A similar error to the previous one is in translating "respiratory droplets" as "picături respiratorii" instead of "picături nazale". It reflects a discrepancy in the interpretation of medical terminology and the nuanced understanding of the anatomical contexts associated with respiratory transmission of viruses and medical usage. In English, "respiratory droplets" denotes tiny particles of moisture expelled from the respiratory system when an individual breathes, speaks, coughs, or sneezes. These droplets may contain infectious agents such as viruses and bacteria, contributing to the transmission of respiratory infections, including Covid-19. The translation "picături respiratorii" accurately captures the concept of droplets associated with respiratory processes, aligning with the literal meaning of "respiratory droplets" by combining the noun "picături" (droplets) with the adjective "respiratorii" (respiratory). However, the collocation error arises from the fact that the term "picături nazale" would be a more contextually accurate translation in Romanian, as it specifies the nasal usage of the droplets. This distinction is crucial in the context of Covid-19 transmission, where respiratory droplets expelled from the nose play a significant role in virus dissemination and infection spread.

- *respiratory etiquette* – "etichete respiratorii" (medical)

The translation error made by Google Translate in rendering "respiratory etiquette" into Romanian as "etichete respiratorii" instead of "coduri de conduită" can be analysed through linguistic and contextual lenses. "False friends" occur when words in two languages look similar but have different meanings. In this case, "etichetă" in Romanian corresponds to "label" in English, not "etiquette" in the sense of social behaviour. Additionally, a collocation error" pertains to the inappropriate pairing of words within a specific context. "Etichete de conduită" doesn't capture the intended meaning of "respiratory etiquette" accurately. "Respiratory etiquette" refers to a set of behaviours related to preventing the spread of respiratory illnesses like Covid-19. In this context, "coduri de conduită" would be more appropriate, as it captures the notion of behavioural norms or codes related to respiratory health during the pandemic.

- respiratory medicine – "*medicină respiratorie*"

The collocation error made by Google Translate in translating "respiratory medicine" as "medicină respiratorie" instead of "pneumologie" in a Covid-19 text from English into Romanian reflects a discrepancy in the idiomatic and professional terminologies between the source and target languages. In English, "respiratory medicine" is a specialized medical term referring to the branch of medicine that focuses on the diagnosis, treatment, and management of respiratory conditions and diseases affecting the lungs, bronchi, and airways. It encompasses various disciplines such as pulmonology, allergology, and respiratory physiology. The translation "medicină respiratorie" captures the literal meaning of "respiratory medicine" by combining the noun "medicine" with the adjective "respiratory" in Romanian. While grammatical, "medicină respiratorie" may not fully convey the specialized nature and scope of the medical field encompassed by "respiratory medicine." On the other hand, "pneumologie" is the term commonly used in Romanian to denote the medical specialty concerned with the study and treatment of diseases related to the respiratory system, particularly focusing on the lungs and respiratory tract. It encompasses the diagnostic and therapeutic aspects of respiratory conditions, including asthma, chronic obstructive pulmonary disease (COPD), and pneumonia.

- *shortages of beds* – "lipsuri de paturi"

The collocation error in the translation of "there are shortages of hospital beds" into "există lipsuri de paturi de spital" by Google Translate represents a discrepancy in the natural pairing of words within the target language. In the original phrase, "shortages of hospital beds," the term "shortages" denotes a deficit or insufficiency of available hospital beds relative to demand or need. This phrase emphasizes the scarcity or inadequacy of beds within the healthcare system, conveying a sense of urgency or concern regarding the capacity to accommodate patients. However, the translation "există lipsuri de paturi de spital" utilizes the noun "lipsuri" to convey the idea of shortages or deficiencies. Even if "lipsuri" does convey a sense of lacking or shortage, it may not collocate naturally with "paturi de spital" in Romanian, further contributing to the collocation error. A more appropriate and contextually accurate translation would involve using terms like "lipsă" or "deficit" to convey the concept of a shortage of hospital beds. These terms are more commonly used in Romanian to denote a lack or insufficiency and better align with idiomatic usage and linguistic conventions. Therefore, a more precise translation would be "există lipsă de paturi de spital" or "există deficit de paturi de spital," which effectively communicates the notion of insufficient hospital bed availability in a manner consistent with natural language usage in Romanian.

- *shutdown* – "închis"

The translation made by Google Translate of "shutdown" into Romanian as "închis" instead of "închidere totală" is a collocation error. "Shutdown" in the context of the pandemic refers to a complete or partial closure of businesses or government services, encompassing non-essential businesses, schools, and public transportation. The term "închidere totală" accurately captures the concept of a comprehensive shutdown, whereas "închis" does not convey the same level of closure or comprehensiveness. In the context of Covid-19, "shutdown" specifically refers to measures aimed at restricting the operation of various establishments and services to mitigate the spread of the virus. "Închidere totală" better conveys the intended meaning in Romanian, emphasizing the comprehensive nature of the closure. While "lockdown" (izolare) and "shutdown" (închidere totală) are related concepts, they have distinct implications and applications. "Lockdown" typically refers to strict government-imposed restrictions on movement and public gatherings aimed at controlling the spread of a contagious disease. On the other hand, "shutdown" specifically pertains to the closure of businesses and services. It is essential to maintain this differentiation in translation to accurately convey the specific measures being implemented.

- *the situation is cloudier* – "situația este tulbure"

The translation error observed in the rendering of "the situation is cloudier" from English into Romanian as "situația este tulbure" by Google Translate exemplifies a collocation error. In the original English expression, "the situation is cloudier" conveys a metaphorical sense of increasing uncertainty or complexity. The term "cloudier" suggests a growing lack of clarity or transparency, implying that the situation has become more ambiguous or difficult to discern. However, the translation "situația este tulbure" in Romanian does not effectively capture the intended meaning of the original phrase. While "tulbure" translates to "cloudy" in Romanian, the collocation "situația este tulbure" lacks idiomatic resonance and is not commonly used in Romanian discourse to denote increased uncertainty or complexity. A more contextually appropriate and linguistically accurate translation in Romanian could involve expressions like "situația este neclară" (the situation is unclear) or "situația este incertă" (the situation is uncertain). These alternatives better convey the sense of ambiguity and lack of clarity implied by "cloudier" in English while aligning more closely with natural language usage and idiomatic conventions in Romanian.

- *to smuggle the instructions* – "a introduce contrabandă instrucțiunile"

The translation error made by Google Translate, rendering "to smuggle the instructions" as "a introduce contrabandă instrucțiunile" reflects a misunderstanding of the intended meaning and a collocation error. Let's dissect the reasons behind this linguistic discrepancy. "To smuggle" in English does not typically connote the act of introducing something surreptitiously, as the term "contrabandă" suggests. Instead, "to smuggle" often implies a cozy or intimate action, unrelated to the act of introducing instructions. The use of "contrabandă" (contraband) in the translation is a collocation error. "Contrabandă" typically refers to smuggling illegal goods or items, which is unrelated to the context of "smuggling instructions." Google Translate may have lacked the contextual understanding necessary to accurately interpret the phrase "to smugle the instructions." Without a deeper understanding of the intended meaning within the context of Covid-19-related instructions, the translation may result in nonsensical or inaccurate renderings. Languages contain unique cultural and linguistic nuances that may not always have direct equivalents in other languages. The concept of "snuggling instructions" may not have a direct counterpart in Romanian, requiring a more nuanced understanding of the intended meaning. The correct translation you provided, "a introduce în celule instrucțiunile," offers a more appropriate interpretation of the intended meaning. It conveys the idea of inserting or placing the instructions in cells, which aligns more closely with the context of Covid-19-related instructions. This translation captures the intended action without introducing extraneous or inaccurate connotations like "contrabandă."

- *to speed up* – "a accela studiile clinice" (medical)

The error exhibited in the translation from English into Romanian, where "clinical vaccine trials are speeding up" is translated as "se accelează studiile clinice" by Google Translate, represents a collocation error and a distortion. In the original English phrase, "speeding up" is a phrasal verb indicating an increase in pace or acceleration. It conveys the idea that clinical vaccine trials are progressing more quickly than before. However, the translation "se accelează studiile clinice" does not accurately capture the intended meaning. While "accelerare" (acceleration) and "a accelera" (to accelerate) are appropriate translations for "speeding up," they are not the most natural or idiomatic choices in Romanian for describing the acceleration of a process like clinical trials. A more appropriate translation could include phrases like "se grăbesc studiile clinice" (clinical trials are being hurried) or "studiile clinice sunt făcute rapid" (clinical trials are being conducted quickly). These alternatives better convey the sense of increased speed and urgency in the clinical trial process. The choice of "accelează" in the translation demonstrates a lack of understanding of collocational patterns in Romanian, as "acceleration" is not commonly associated with "studiile clinice" in natural discourse. This highlights the importance of linguistic nuance and contextual understanding in translation, particularly when dealing with idiomatic expressions and specialized terminology.

- *stagnant water* – "apă stătută"

The translation error of "apă stagnată" for "stagnant water" by Google Translate reflects confusion of sense relations and highlights the importance of semantic precision and contextual understanding in translation. The suggested translation of "apă stătută" would align more closely with the intended meaning and usage in Romanian. "Stagnant water" in English refers specifically to water that is not flowing or moving, often found in ponds, puddles, or other bodies of water with limited circulation. The term "stagnant" emphasizes the lack of movement or progress in the water. However, the Romanian word "stagnată" may carry connotations related to lack of progress or development in a broader sense, rather than specifically referring to water that is not flowing. The term "apă stagnată" conveys the idea of water that has become still or motionless, but it does not fully capture the specific quality of water that is stagnant. "Stagnată" emphasizes a lack of development or evolution, which may not accurately reflect the intended meaning of the original term. "Apă stătută," on the other hand, more precisely conveys the concept of water that is stagnant or not moving, aligning better with the intended meaning of "stagnant water."

- *talk* – "vorbirea"

The direct translation "Vorbirea despre mutații" into Romanian from "Talk of mutations" results in a collocation error. In Romanian, "vorbirea" refers more specifically to the act or process of speaking, as in the physical or cognitive ability to produce spoken language (similar to "speech"). It is rarely used in the context of discussing or talking about a subject in an informal or conversational manner. Conversely, "discuția" is the more appropriate choice for "talk" when referring to a discussion or conversation about a particular topic. In the bilingual dictionary, "talk" is translated by "1.discuții; bârfe", "2. conversație/discuție", "3. discurs".

- *this is all well and good* – "toate acestea sunt bune și bune"

The translation provided by Google Translate, "Toate acestea sunt bune și bune," for the English phrase "This is all well and good" into Romanian is inaccurate and fails to capture the meaning and nuances present in the original text. "This is all well and good" is an idiomatic expression used to convey approval or acceptance of a situation or course of action. It implies that everything is satisfactory or acceptable in a given context. However, the Romanian translation provided by Google Translate, "Toate acestea sunt bune și bune," lacks the idiomatic equivalence and fails to convey the intended meaning effectively. The use of "și bune" twice in the translation does not accurately capture the idiomatic sense of the original phrase. To rectify this translation error and provide a more accurate rendition of the original phrase, alternative translations such as "asta e foarte bine" or "Asta e bine și de bine" would be more appropriate. These translations capture the idiomatic sense of approval or acceptance conveyed by the English expression while aligning more closely with the nuances of the Romanian language and cultural context.

- *tight chest* – "piept strâns" (medical)

The structure "piept strâns" provided by Google Translate as a translation for "tight chest" in the context of Covid-19 symptoms is an example of a collocation error that stems from both semantic and cultural considerations. "Tight chest" is a common expression used to describe the sensation of pressure, constriction, or discomfort in the chest area. This feeling is often associated with respiratory issues and is frequently reported by individuals experiencing Covid-19 symptoms. In the context of Romanian medical discourse, the structures "opresiune","senzație de apăsare în piept", and "constricție toracică" are more commonly used to convey the sensation of tightness or constriction in the chest. The structure that aligns more closely with the nuances of medical terminology and the experiences reported by individuals with Covid-19 symptoms is "opresiune". The last term is described in a specialised dictionary as "the impression or sensation of respiratory distress, dyspnea", as Titirică stated (2019: 460).

- *tracking* and *a game-changer* – "urmărirea persoanelor" and "schimbător de joc"

In the translation of the phrase "Tracking who is infected is essential to controlling the transmission of contagious diseases. Could digital technology prove to be a game-changer for the current pandemic?" to "Urmărirea persoanelor infectate este esențială pentru controlul transmiterii bolilor contagioase. Tehnologia digitală s-ar putea dovedi a fi un schimbător de joc pentru pandemia actuală," Google Translate introduces two significant collocation errors that compromise the clarity and appropriateness of the translation. Firstly, the term "tracking" is translated as "urmărirea." While "urmărirea" literally translates to "following" or "pursuing," and can be used to describe the act of keeping track of something, it tends to carry a connotation of surveillance or chasing, which might be misleading in the context of public health. The focus in the original context is on the systematic observation and collection of data regarding the spread of infections. More precise alternatives would be "monitorizarea" or "depistarea." In order ti imply an ongoing process of observing and recording data over time, and is commonly used in healthcare to describe the observation of disease trends, and to empasize the identification of cases, aligning closely with efforts to track the emergence and spread of infectious diseases. Secondly, the phrase "game-changer" is inadequately translated as "schimbător de joc." This literal translation does not capture the idiomatic nuance of the English term, which refers to something that significantly alters the dynamics or outcomes in a given situation. "Schimbător de joc" is a direct translation that is not natural in Romanian. A more fitting translation would be "un factor care schimbă datele problemei ", in order to effectively communicate the concept of something fundamentally altering the circumstances or expectations in a significant way. It reflects the notion of a decisive influence that can substantially impact the approach or outcomes of managing the pandemic.

- *the virus swabbed in each test* – "virusul tamponat" (medical)

A similar collocation error is in the Romanian translation "virusul tamponat în fiecare test" for the English phrase "the virus swabbed in each test," as generated by Google Translate. It reflects a misunderstanding of both the context and the specific terminology related to coronavirus testing procedures. The erroneous translation demonstrates a failure to accurately convey the intended meaning due to issues with word choice, syntactic structure, and semantic interpretation. The English phrase "the virus swabbed in each test" suggests the action of using a swab to collect samples containing the virus during the testing process. "Swabbed" serves as a past participle verb, indicating the method by which the virus is obtained for testing. The term "each test" emphasizes the repetition of this procedure across multiple instances of testing, underscoring the ubiquity and systematic nature of the testing regimen. In contrast, the translation "virusul tamponat în fiecare test" presents several challenges. Firstly, the term "tamponat" suggests the application of a swab or tampon, but it does not accurately convey the action of collecting samples from the nasal or throat passages, as implied by the English phrase "swabbed." Secondly, the preposition "în" (in) used in the translation introduces ambiguity regarding the specific context of the testing procedure and the relationship between the virus and the test itself. Moreover, the translation overlooks the crucial aspect of antigen testing, which involves the detection of specific viral proteins indicative of a coronavirus infection. The absence of this essential detail further detracts from the accuracy and comprehensibility of the translation. The correct translation, "virusul depistat prin fiecare test antigen cu bețișor nazal sau laringial," addresses these shortcomings by incorporating the necessary technical terminology and clarifying the testing methodology. The phrase "depistat prin fiecare test" (detected through each test) accurately conveys the process of identifying the virus within the sample collected during antigen testing. Additionally, the inclusion of "bețișor nazal sau laringial" specifies the type of swab used for sample collection, enhancing the precision and specificity of the translation.

- *vital lifeline* – "colac vital"

The mistranslation of "vaccination could prove a vital lifeline" by Google Translate into "vaccinarea s-ar putea dovedi un colac vital" highlights a discrepancy in both semantic and contextual accuracy. Such errors underscore the challenges inherent in translating nuanced expressions that carry specific connotations and cultural significance. In the original English phrase, "vaccination could prove a vital lifeline," the term "vital lifeline" metaphorically suggests that vaccination serves as a crucial means of protection or support, potentially saving lives in the context of disease prevention. The term "lifeline" conveys the notion of a critical or indispensable resource that provides assistance or relief during challenging circumstances. However, the translation "vaccinarea s-ar putea dovedi un colac vital" is imprecise and lacks clarity in conveying the intended meaning. The term "colac" in Romanian has multiple meanings, including "lifebuoy," "inner tube," or "ring", or "pretzel." While "colac" could be interpreted metaphorically as a form of support, its association with flotation devices or swimming aids may not effectively capture the gravity and significance of the intended message. A more accurate and contextually appropriate translation could involve phrases like

"vaccinarea s-ar putea dovedi un colac de salvare" (a lifebuoy) or "vaccinarea s-ar putea dovedi o soluție vitală" (a vital solution). These alternatives better convey the metaphorical sense of vaccination serving as a critical means of protection and support in safeguarding individuals against disease.

- *wear a properly fitted mask* – "a purta mască corect montată" (medical)

The phrase "purtați mască corect montată" as rendered by Google Translate exemplifies a collocation error stemming from the misappropriation of terms within the context of the target language, Romanian, while translating the English expression "wear a properly fitted mask." This translation inadequacy arises primarily due to a lack of nuanced understanding of both the English source phrase and the nuances of Romanian linguistic conventions. Firstly, the phrase "mască corect montată" reflects a fundamental mismatch between the intended meaning of "properly fitted mask" and the translated expression. In Romanian, "montată" conveys the act of mounting or installing something, typically in a physical or structural sense. Thus, the term "montată" is linguistically incongruent when applied to the act of wearing a mask, as it implies a physical attachment or installation rather than the proper adjustment of the mask to fit one's face. Secondly, the phrase "mască corect" also exhibits a cacophony in its structure. While "corect" appropriately denotes correctness or accuracy, its juxtaposition with "mască" lacks the semantic precision required to convey the notion of a properly fitted mask. The term "corect" fails to capture the specific connotation of proper fitting or adjustment inherent in the original English expression. In light of these linguistic shortcomings, the suggested translation "purtați corect masca" offers a more linguistically coherent and contextually appropriate rendition. The term "pusă corect" effectively encapsulates the notion of correctly placing or positioning the mask, thereby ensuring a proper fit to the wearer's face.

- *we all have biases* – "cu toții avem părtiniri"

The translation "cu toții avem părtiniri" by Google Translate for "we all have biases" represents a collocation error due to the misalignment of terms and the failure to capture the intended semantic nuances of the original English phrase. The suggested translations "cu toții avem prejudecăți" or "toți suntem părtinitori" offer more contextually accurate renditions, thereby rectifying the collocation error and aligning with the intended meaning, especially within the context of an article discussing fake news and skepticism during the pandemic. The phrase "cu toții avem părtiniri" introduces a collocation error primarily due to the use of "părtiniri" (partisanship) instead of "biases" to convey predispositions or inclinations. While "părtiniri" denotes a bias or preference towards a particular party or viewpoint, it does not effectively capture the broader spectrum of cognitive biases and predispositions implied by the term "biases" in the original English phrase. "Părtiniri" carries connotations of political or ideological alignment, which may not accurately reflect the cognitive biases discussed in the context of fake news and skepticism during the pandemic.

- *what is the difference between efficacy and effectiveness* – "care este diferența dintre eficacitate și eficacitate*?* "

A similar translation error rendered the phrase "What is the difference between efficacy and effectiveness" as "Care este diferența dintre eficacitate și eficacitate?" in Romanian, stems from a failure to accurately capture the nuanced distinctions between the terms "efficacy" and "effectiveness" in the context of the original English phrase. "Efficacy" and "effectiveness" are distinct terms in English that carry specific meanings in the domain of healthcare and research. "Efficacy" refers to the ability of a treatment or intervention to produce a desired effect under ideal or controlled conditions, typically in clinical trials or experimental settings. On the other hand, "effectiveness" pertains to the extent to which a treatment or intervention achieves its intended outcomes in the real-world or in practical settings, considering factors such as patient adherence, healthcare delivery systems, and contextual variables. In the provided translation, "eficacitate" is used twice, which fails to differentiate between the concepts of "efficacy" and "effectiveness." This lack of distinction results in a loss of precision and clarity regarding the specific meanings conveyed by the original English terms. Consequently, the translated phrase fails to accurately capture the nuanced differences between "efficacy" and "effectiveness" within the context of Covid-19-related discussions, where the effectiveness of interventions and public health measures is of paramount importance. To address the translation error and provide a more accurate rendition of the original phrase, alternative translations such as "Care este diferența dintre eficacitate și eficiență?" would be more appropriate. By replacing "eficacitate" with "eficiență," the translated phrase accurately reflects the distinction between "efficacy" and "effectiveness" in Romanian, aligning more closely with the intended meanings conveyed by the original English terms.

## 2.3.2.1. Arbitrary combinations

- *breakthrough infection* – "infecție episcopală" (medical)

The error of an arbitrary combination made by Google Translate in rendering "breakthrough infection" as "infecție episcopală" in Romanian instead of "infecție care a recidivat" reflects a failure in understanding the specific medical context and terminology associated with the term "breakthrough infection." It represents an instance of an arbitrary combination error. Such errors occur when the translation algorithm generates combinations of words or phrases that lack idiomatic or semantic congruence in the target language and it refers to an infection that occurs in an individual who has been vaccinated against the disease, typically with a focus on cases where the infection breaks through the protective barrier provided by vaccination. In this context, "breakthrough" implies an unexpected or uncommon event where the infection surpasses the protective effects of vaccination. The translation "infecție episcopală" is an arbitrary combination that does not accurately capture the intended medical meaning of "breakthrough infection." "Episcopală" pertains to the term "episcopal," which is associated with bishops or the church hierarchy, and is completely unrelated to the medical context of breakthrough infections. A more appropriate translation for "breakthrough infection" in Romanian would involve conveying the specific medical concept of an infection that occurs despite prior vaccination. "Infecție care a recidivat" captures the essence of an infection recurring or relapsing after an initial vaccination or treatment, aligning more closely with the intended medical meaning.

- *hidden harms* – "păunurile ascunse"

The translation error in the phrase "hidden harms" from "Meanwhile, 'hidden harms' such as neglect and domestic abuse were harder to spot and address while social workers were required to work remotely," into "păunurile ascunse" by Google Translate represents an instance of arbitrary combination. In English, the term "hidden harms" refers to negative impacts or damages that are not immediately visible or easily detectable, often due to their subtle or covert nature. These can include issues like neglect or domestic abuse, which are typically concealed within private settings and become more difficult to identify and address, especially in contexts where social workers are constrained to remote operations. Here, "păunurile" translates to "peacocks," which is nonsensical in this context, as it bears no relation to the concept of harm or injury. This erroneous word choice likely stems from a confusion with similarly spelled words, reflecting an arbitrary and inaccurate lexical selection. The more accurate translation for "hidden harms" in this context would be "daune ascunse". "Ascunse" aptly conveys the idea of being hidden or not immediately visible.

In summary, arbitrary combination errors highlight the importance of linguistic sensitivity, cultural understanding, and contextual awareness in translation processes. Achieving accurate and effective translations requires a nuanced understanding of collocational patterns, idiomatic expressions, and semantic nuances specific to both the source and target languages.

## 2.3.2.2. Preposition partners

Some lexical errors made by Google Translate in texts related to Covid-19 from English into Romanian represent instances where the choice of prepositions does not align with the idiomatic usage or syntactic conventions of the Romanian language. Such errors can lead to inaccuracies, syntactic ambiguities, and semantic distortions in the translated text. Here are the examples of lexical errors and their correct counterparts:

- "a se îmbolnăvi cu Covid" instead of "a se îmbolnăvi de Covid": The preposition "cu" suggests a sense of accompaniment or association, which is not idiomatic in Romanian when discussing contracting an illness. The correct preposition is "de," which conveys the idea of becoming ill with Covid-19.

- "ne ajută să ne protejăm în a ne îmbolnăvi de acel germen în viitor" instead of "ne ajută să ne protejăm de îmbolnăvirea cu acel germen în viitor": Here, the preposition "în" does not appropriately convey the intended meaning of protection from illness. "De" is the correct preposition to express protection from illness caused by a specific germ.

- "la 27 august" instead of "pe 27 august": While "la" can denote a sense of "at" or "on" in certain contexts, "pe" is the appropriate preposition to indicate a specific date or point in time, such as "on August 27[th]."

- "a fi implicat cu" instead of "a fi implicat în": The preposition "cu" suggests a sense of being together or accompanying, which does not accurately convey the intended meaning of involvement or association. "În" is the correct preposition to denote involvement or participation in a particular situation or activity.

These lexical errors reflect the challenges inherent in Machine Translation systems like Google Translate. To address such errors, it is essential to enhance the system's understanding of idiomatic expressions, syntactic structures, and semantic nuances specific to the target language. By improving the accuracy and appropriateness of preposition partners in translations, Machine Translation systems can produce more fluent, natural, and contextually appropriate renditions of text in Covid-19 related contexts and beyond.

## 2.4. Stylistic errors

Google Translate relies on existing data, which might be limited for these specific terms. Stylistic errors in translation refer to deviations that impact the tone, register, or stylistic consistency of a text, resulting in a translation that fails to align with the intended style of the original. These errors often arise from an over-reliance on literal translation, insufficient cultural adaptation, or the inability of Machine Translation systems to capture subtleties of tone and register. Stylistic errors in Machine Translation often arise from the system's inability to fully capture the nuances of tone, register, or natural language flow. Such errors include inappropriate formality or informality, awkward word choices, redundancy, and unnatural sentence structures. Zinsmeister and Heid (2004:311-314) have noted that these issues frequently stem from limitations in aligning linguistic style across languages, as well as from the focus of many Machine Translation systems on meaning rather than stylistic appropriateness.

- a cui agendă aș putea sprijini prin împărtășirea ei

The phrase "A cui agendă aș putea sprijini prin împărtășirea ei?" includes a word order error and a lexical choice that deviate from the intended meaning of the original English sentence "Whose agenda might I be supporting by sharing it?" This translation issue made by Google Translate demonstrates the challenges inherent in Machine Translation, particularly when dealing with nuanced linguistic structures and context-specific terminology. In the Romanian translation, the word order is inverted compared to the original English sentence. The use of the word "agendă" in Romanian does not accurately convey the intended meaning of "agenda" in the context of the original English sentence. While "agendă" can refer to a list of items or scheduled events, it does not capture the brofader connotations of "agenda" in English, particularly in the context of discussing information related to Covid-19. "Agenda" in this context implies a set of topics or issues to be addressed or promoted, which may include facts, data, or specific viewpoints. Additionally, "împărtășirea ei" (its sharing) is a somewhat awkward phrasing in Romanian. While it conveys the idea of sharing, it lacks precision and may not be the most natural way to express the concept in this context. A revised translation that addresses these concerns could be: " "Pe cine aș putea ajuta prin distribuirea acestei informații?" This translation maintains the essence of the original question while addressing the lexical and structural issues present in the initial Machine Translation and respecting the natural word order in the target language.

- anticorpii durează timp să apară (medical)

The translation "Anticorpii durează timp să apară" for the English structure "Antibodies take time to appear" is erroneous due to its failure to adhere to the syntactic structure and idiomatic usage of Romanian. This discrepancy reflects a collocation error and a misinterpretation of the grammatical conventions governing the Romanian language. In English, the phrase "Antibodies take time to appear" is syntactically sound and grammatical. It employs a subject-verb-object (SVO) structure, where "antibodies" (subject) take action (verb) to appear (object). The phrase conveys the idea that the process of antibody production requires a certain amount of time. However, the translation "Anticorpii durează timp să apară" presents several issues. Firstly, the word order is unnatural in Romanian syntax. While "anticorpii" (antibodies) is appropriately positioned as the subject, the verb "durează" (take) is misplaced, as it should not precede the object "timp" (time). In Romanian, the verb "durează" typically functions as an intransitive verb, and it requires a different syntactic structure to convey the intended meaning accurately. A more idiomatic and grammatical translation in Romanian would be "Durează până apar anticorpii." This structure follows the standard Romanian syntax, where the verb "durează" appropriately precedes the adverbial phrase "până apar" (until they appear), and "anticorpii" serves as the object of the sentence. Furthermore, the original English phrase emphasizes the temporal aspect of antibody production, suggesting that it takes time for antibodies to develop. The translation should accurately reflect this temporal relationship, which the revised Romanian version achieves by incorporating the adverbial phrase "până apar" (until they appear).

- curbă netezită

Google Translate may misinterpret idiomatic expressions and translate them word-for-word, resulting in nonsensical or misleading translations. For instance, the phrase "flatten the curve" could be translated into Romanian as "netezirea curbei," but the translation was "curbă netezită," the meaning may be lost or distorted. The placement of prepositional phrases and adverbial clauses can significantly impact the meaning of a sentence. Errors in placing these phrases can lead to confusion or misinterpretation in the translated text. For example, the phrase "during the pandemic" should be translated as "în timpul pandemiei" in Romanian, but if the translation places the prepositional phrase incorrectly, it might result in "pandemia în timpul," which is syntactically ill-formed.

- locuri cu oameni înghesuiți

The word order error made by Google Translate in translating from English "places with people crowded together" into Romanian as "locuri cu oameni înghesuiți" reveals a discrepancy in understanding syntactic structures and idiomatic usage between the two languages. The erroneous translation reflects a lack of contextual and semantic sensitivity, resulting in a deviation from the intended meaning and linguistic conventions. In English, the phrase "places with people crowded together" conveys the idea of locations where individuals are densely packed or gathered closely in a confined space. The term "crowded together" serves as a compound modifier, describing the spatial arrangement of people within these places. The syntactic structure follows the pattern of attributive modification, where the modifier ("crowded together") precedes the noun it modifies ("people"). Conversely, the translation "locuri cu oameni înghesuiți" employs a different word order and syntactic structure. While grammatical, the translation places the adjective "înghesuiți" (crowded) after the noun "oameni" (people), resulting in a less idiomatic and semantically ambiguous expression. The adjective "înghesuit" typically describes a state of being compressed or confined, conveying the sense of physical constraint or tightness rather than spatial density or crowding. A more linguistically accurate and contextually appropriate translation in Romanian would be "locuri înghesuite" or "locuri aglomerate." These expressions maintain the syntactic integrity and idiomatic consistency of the original English phrase while effectively conveying the intended meaning of densely populated places or crowded areas. Furthermore, the use of the adjective "înghesuit" after the noun "oameni" introduces semantic ambiguity and detracts from the clarity and precision of the translation. The term "înghesuit" conveys a sense of physical constraint or confinement, which may lead to misinterpretation or confusion regarding the intended message.

- noua variantă răspândire rapidă

There are literal translations. Translating each word individually without considering sentence structure, leading to unintelligible sentences. For example, "The new variant is spreading quickly" was translated to "Noua variantă răspândire rapidă," which lacks a verb.

- poartă oamenii măști

An illustrative example of a stylistic error produced by Google Translate can be observed in the translation of the English sentence "*people wear masks*" into Romanian as "*poartă oamenii* *măşti.*" While the lexical choices are largely accurate, the syntactic arrangement reflects an unnatural word order for declarative sentences in Romanian. The translation inverts the standard subject-verb-object structure, placing the verb "*poartă*" (wear) before the subject "*oamenii*" (people), which in Romanian would typically signal an interrogative or emphatic construction rather than a neutral statement. The more appropriate and stylistically correct rendering would be "*oamenii poartă măşti,*" which maintains the expected syntactic and communicative norms of Romanian.

- a oferi assistenţă critică

The phrase "să ofere asistenţă critică," translated from "provide critical assistance" by Google Translate, contains a stylistic error that can lead to confusion in Romanian. The word "critic" in Romanian, while it does share roots with the English term "critical," often carries connotations more closely aligned with a state of crisis or something being in a dire, crucial condition, such as "situaţie critică" (critical situation). This usage can imply that the assistance is in response to an urgent, life-threatening crisis rather than emphasizing its importance or necessity. In contrast, the suggested translation, " A oferi ajutor esenţial," more accurately conveys the intended meaning of "critical" in the context of assistance. The term "esenţial" translates to "essential," which underscores the indispensable and foundational nature of the help being provided. This term is more appropriate for conveying the vital importance of the assistance without implying an immediate crisis, thus maintaining clarity and aligning better with the intended nuance of the English phrase. Moreover, "ajutor" (help or aid) is a more fitting and commonly used term in Romanian for assistance in this context, as opposed to "asistenţă," which is often reserved for more formal or technical contexts, such as medical or technical assistance. Hence, "oferi ajutor esenţial" accurately captures the essence of providing crucial and significant support, which is central to the original English phrase.

- o erupție cutanată rău

The English phrase "a bad rash" is a common medical term used to describe a severe or uncomfortable skin condition. The word "bad" in this context functions as an adjective modifying the noun "rash." Google Translate's translation of "a bad rash" to "o erupție cutanată rău" is problematic for several reasons. The translation is a literal one, which does not consider the stylistic and semantic differences between English and Romanian. While "rău" can indeed mean "bad" in Romanian, its use in this context is inappropriate. In Romanian, "rău" may be used as a quantifying adverb, similar to "very" or "quite" in English, especially in informal contexts. Machine Translation systems often struggle with understanding context. In a medical context, it is crucial to maintain the accuracy and specificity of the translation, which Google Translate failed to achieve in this instance.

- petrecerea mai puțin timp la locul de muncă

The translation provided by Google Translate, "Petrecerea mai puțin timp la locul de muncă," in response to the English phrase "Spending less time in the workplace" exemplifies a structural error, chiefly manifesting in the misplacement of linguistic elements and the misinterpretation of semantic nuances inherent in both languages, English and Romanian. Primarily, the phrase "Petrecerea mai puțin timp la locul de muncă" introduces a structural dissonance by placing the noun "petrecerea" (party) at the beginning of the sentence, thus conferring an unintended connotation to the translation. In Romanian, "petrecerea" commonly refers to a social gathering or celebration, conveying a semantic divergence from the intended meaning of "spending" in the context of time allocation within the workplace. Furthermore, the phrase "mai puțin timp" effectively encapsulates the concept of spending less time, aligning with the intended meaning of the source phrase. However, the positioning of "mai puțin timp" within the sentence structure, particularly following "petrecerea," contributes to syntactic incongruity, resulting in a disjointed and linguistically discordant expression. In contrast, the suggested translation "dacă petreci mai puțin timp la locul de muncă" offers a structurally coherent and semantically accurate rendition of the original English phrase. By placing the verb "petreci" (to spend) at the beginning of the sentence and introducing the conditional conjunction "dacă" (if), the translation effectively communicates the conditional aspect inherent in the source phrase. Moreover, the utilization of the infinitive "petrece" maintains linguistic fidelity while conveying the intended action of spending less time in the workplace.

- (oamenii se infectau din cauză) relativă uşurinţă

The error in the translation provided by Google Translate, where "relative ease" is translated as "relativă uşurinţă" instead of "uşurinţei relative", reflects a word order error that impacts the clarity and precision of the target text. Such errors occur when the translator misplaces modifiers or adjectives, resulting in ambiguity or confusion regarding the intended meaning. In the original English phrase, "relative ease" modifies the noun "ease" to indicate that the ease with which the virus could spread is in comparison to some standard or reference point. The term "relative" serves as an adjective emphasizing the comparative aspect of the ease of virus transmission. However, the translation "cauza relativă uşurinţă" does not effectively convey this comparative relationship and results in ambiguity in the target text. Placing "relativă" before "uşurinţă" disrupts the intended meaning and clarity of the phrase, as it suggests that "relativă" modifies "cauza" rather than "uşurinţă". A more linguistically accurate and contextually appropriate translation would involve placing "relative" after "uşurinţă" to maintain the intended meaning and clarity of the original phrase. Thus, the corrected version would be: "un număr mult mai mare de oameni se infectau din cauza uşurinţei relative cu care virusul s-ar putea răspândi." This revised version ensures that "relative" appropriately modifies "ease" (uşurinţă) and conveys the intended comparative aspect of the ease of virus transmission.

- urmează să înceapă să-şi uşureze blocarea

The translation of the phrase "is due to start easing its lockdown" by Google Translate as "urmează să înceapă să-şi uşureze blocarea" (from "South Africa, meanwhile, which has had one of the strictest lockdowns in the world, banning the sale of cigarettes and alcohol as well as dog walking, is due to start easing its lockdown from 30 April, with some businesses and schools reopening, although with limits on class-size" - "Între timp, Africa de Sud, care a avut una dintre cele mai stricte blocaje din lume, interzicând vânzarea de ţigări şi alcool, precum şi plimbarea câinilor, urmează să înceapă să-şi uşureze blocarea din 30 aprilie") introduces a significant stylistic error that disrupts the natural flow and clarity of the text. Firstly, the verb "a uşura" in Romanian translates to "to ease" or "to lighten" in a very literal sense, typically used to describe the act of making a physical or emotional burden lighter. This usage is inappropriate for describing the complex and systematic process of easing lockdown restrictions during the Covid-19 pandemic. In the context of public health measures, "lockdown" refers to a set of strict regulations aimed at restricting movement and activities to control the spread of the virus. The literal translation "blocare" is technically correct for "lockdown," but it lacks the nuance and specificity needed to describe the phased reduction of these restrictions. The phrase "să-și ușureze blocarea" is awkward and unclear, suggesting a simplistic or personal lightening of an unspecified block rather than a carefully managed, government-led easing of societal restrictions. A more precise and stylistically appropriate translation would be "se preconizează o relaxare graduală a izolării." This phrase captures several important elements. "Se preconizează" translates to "is planned" or "is anticipated," suggesting a structured and forward-looking approach to policy changes. "Relaxare graduală" means "gradual easing", which better reflects the step-by-step approach often used in lifting lockdown measures. Finally, "izolării" translates to "the lockdown" or "the isolation," which is a more fitting term in the context of the pandemic than "blocare." This term conveys the broader societal and public health implications of lockdowns. By using "se preconizează o relaxare graduală a izolării," the translation aligns with the notion of a planned and phased approach to reducing restrictions, suggesting a methodical process that involves careful consideration and monitoring.

- vaccinul protejează de boli grave (medical)

Some Romanian sentences can be ambiguous due to SOV order. Google Translate might misinterpret the intended meaning. For example, "The vaccine protects against severe illness" could be translate by "Vaccinul protejează împotriva îmbolnăvirii grave," emphasizing the vaccine instead of protection. There are examples of mistranslation of Technical Terms: Lacking specific data for Covid-19 terms, Google Translate might rely on general translations leading to inaccuracies.

- virus transmitere rată (medical)

In Romanian, compound nouns are formed differently from English, and word order errors can lead to ambiguity or confusion in the translation. For example, "virus transmission rate" could be translated as "rata de transmitere a virusului," but if the translation places the words without taking the target syntactic structure into account, it may result in "virus transmitere rată," which is not a grammatical structure.

- Repetitions

In the translated texts related to Covid-19, the occurrences of repetitive structures not only disrupt the flow and coherence of the text but also reflect potential challenges in capturing the nuances of the Romanian language and its syntactic patterns. Firstly, the phrase "le-am inoculat conform directivei guvernului de a inocula tuturor" contains repetition with "de a inocula" appearing twice in close proximity. Such redundancy could be perceived as a deviation from standard Romanian expression, where a more concise construction might be preferred. Similarly, in "câștigând timp în timp ce se dezvoltă un vaccin eficient," the repetition of "în timp ce" within the same sentence could be considered stylistically cumbersome. Romanian offers alternative structures to convey temporal relationships more succinctly. Furthermore, the sentence "când a început blocajul. Ridicarea blocajelor și a altor măsuri, cum ar fi staționarea" exhibits redundancy through the repeated use of "blocaj" and its derivatives. This repetition might impede the text's fluency and could be refined for stylistic variety and clarity of expression. Or "mementourile prin SMS au crescut acoperirea imunizării și oportunitatea vaccinării, 93% dintre cei intervievați spunând că ar fi dispuși să plătească pentru mementouri text" features repetition with "mementourile" and "SMS," which may lead to ambiguity or verbosity. A more concise formulation could enhance the text's readability and impact. The repetitive structures observed in the translated texts highlight potential limitations in the sophistication of Machine Translation systems when handling nuanced linguistic features, syntactic constructions, and stylistic conventions specific to Romanian. Addressing these stylistic errors is essential to ensure the clarity, coherence, and cultural appropriateness of translated texts in the context of Covid-19 communication and beyond.

In summary, lexical errors in Google Translate from English into Romanian related to Covid-19, caused by word order issues, can result in inaccurate, unidiomatic, or confusing translations. Addressing these errors requires improvements in the system's understanding of Romanian syntax, idiomatic expressions, and grammatical structures to ensure more accurate and natural-sounding translations. The word order error in the translation underscores the importance of syntactic accuracy and idiomatic consistency in cross-linguistic communication. Achieving linguistically faithful and contextually relevant translations requires a deep understanding of both the source and target languages, as well as sensitivity to cultural nuances and semantic nuances, in order to facilitate effective communication and convey meaning accurately across linguistic boundaries.

# 3. Result evaluation

This study employed a descriptive analysis method in order to analyze the lexical errors generated by Google Translate when translating Covid-19-related texts from English into Romanian. This method allowed for a detailed examination of specific terms, collocations, and lexical structures, which allow us to achieve a better understanding of the nature of the errors produced by the Machine Translation system. The descriptive qualitative approach above focuses on identifying and categorizing errors. The approach also takes into account relevant quantitative dimensions, paying particular attention to the frequency and occurrence patterns of these errors. The findings, derived from a systematic analysis of numerical data, include the frequencies and percentages of lexical errors, offering insight into Google Translate's performance in this specialized domain. In the following section, the values and percentages for each error category are presented in tables and charts that illustrate the patterns and types of lexical errors that were noted. Machine Translation evaluation has long been a topic of extensive research, with several metrics developed to assess translation quality. The BLEU (Bilingual Evaluation Understudy) score remains one of the most widely used methods for evaluating MT output by comparing n-grams between the machine-generated text and one or more human reference translations. Despite its popularity, BLEU has limitations, especially in the context of lexical errors in domain-specific translations like Covid-19-related content. Research by Callison-Burch et al. (2006) and Koehn (2009) highlights that BLEU focuses on surface-level accuracy, often overlooking deeper semantic and contextual errors, which are crucial when translating technical and medical information.

In this study, it became evident that using BLEU to evaluate lexical errors in Covid-19-related texts was not fully appropriate due to its inherent limitations. BLEU measures n-gram precision but fails to account for lexical nuances, contextual appropriateness, and domain-specific terminology – factors that are critical in ensuring the accuracy and reliability of translations in the medical and public health domains. As Zhang (2021) points out, while BLEU can provide a rough estimate of overall translation quality, it often falls short in evaluating translations that involve specialized terminology or complex linguistic structures. Given the specialized nature of the Covid-19 texts analyzed, alternative methodologies were considered. These include human evaluation, where domain experts assess both accuracy and fluency based on their specialized knowledge, and error analysis methodologies, which categorize and quantify specific types of errors such as mistranslations of medical terms, incorrect collocations, and semantic shifts. Lommel et al. (2014) introduced a hierarchical error taxonomy, the Multidimensional Quality Metrics (MQM), which differentiates between fluency errors and accuracy errors – a distinction that is particularly significant for this study. Fluency errors pertain to the internal linguistic quality of the target text, focusing on issues such as grammar, syntax, punctuation, and overall naturalness. These errors affect the readability and coherence of the output, even when the meaning is preserved. On the other hand, accuracy errors involve deviations from the intended meaning of the source text, such as mistranslations, omissions, or additions. These errors directly compromise the semantic fidelity and reliability of the translation. By distinguishing these two categories, the MQM framework allows for a more nuanced evaluation of Machine Translation output, highlighting the interplay between producing translations that are linguistically fluent while remaining semantically accurate. This distinction is particularly relevant in contexts where stylistic fluency is critical, but not at the expense of preserving the source text's meaning. MQM allows for a more nuanced analysis of errors, taking into account the impact of lexical inaccuracies on both the intelligibility and the informational content of the translated text.

Furthermore, human-in-the-loop evaluation, which incorporates expert feedback into the evaluation process, offers a more robust method for assessing MT performance in specialized fields. According to UCLA Medical Centre (2021), a study conducted by the UCLA Medical Centre found that Google Translate preserved the overall meaning for 82.5% of the translations, with accuracy between languages ranging from 55% to 94%. In their study, Toral and Way (2018) discuss translation evaluation, particularly focusing on the challenges of assessing literary translations. They highlight that while Neural Machine Translation (NMT) systems have made significant strides in improving translation quality, evaluation remains a complex task due to the interplay between fluency and fidelity. Literary texts, in particular, demand not only semantic accuracy, but also stylistic fluency, making standard evaluation metrics like BLEU insufficient for capturing the nuances of literary translation quality. Toral and Way (2018: 316) claim that " the evaluation of literary translations is especially challenging, as it involves both fluency and fidelity, which are not always adequately captured by automatic metrics such as BLEU." While the BLEU score offers a broad measure of translation quality, its limitations in handling lexical precision and domain-specific content necessitate the adoption of more granular and context-aware methodologies. These methods provide a clearer picture of lexical errors in the performance of Google Translate and suggest pathways for improving Machine Translation systems in specialized fields such as medicine and public health.

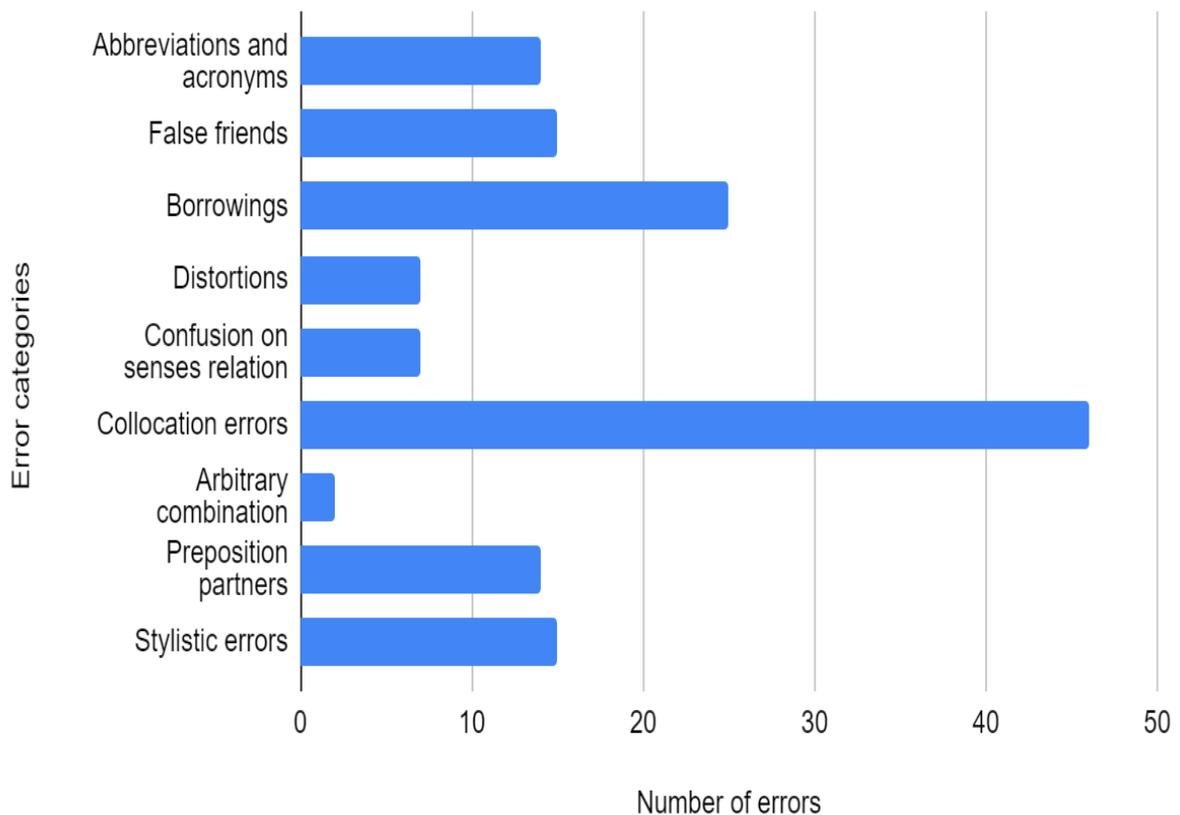

Figure 5. The number of errors by category

The graph above presents different categories of errors on the y-axis and the corresponding number of errors on the x-axis. The collocation errors show the highest frequency of errors, with nearly 50 occurrences. These errors arise when words that are often used together (collocations) are misused. Borrowings, which refer to the inappropriate or incorrect usage of words from other languages, represent the second most frequent category, with around 20 errors. "False friends" also show a significant number of errors, likely around 15. Regarding the stylistic errors, involving misuse of tone, style, or formality, they have about 12 occurrences. Preposition partners, namely those errors in selecting the correct prepositions, also contribute a fair number of errors, slightly above 10. The abbreviation and acronym errors are relatively few in number, just under 10. Distortions and confusions regarding sense relations (where related meanings of words are confused) have around 5 and 7 errors respectively.

Concerning arbitrary combination, likely referring to random or nonsensical word pairings, the figure above shows the least errors, with under 5 occurrences. In summary, collocation errors dominate the chart, followed by borrowings and false friends, indicating a focus on lexical combinations and language interference.

The pie chart in Figure 6 shows the distribution of different types of errors in the context of analyzing how Google Translate handles Covid-related terms. Collocation errors represent the largest proportion, accounting for 44.8% of all errors. This suggests that Google Translate struggles the most with correctly pairing words that typically occur together in specialized Covid-related terms. It highlights the importance of proper word combinations in medical and technical translation, which seems to be a significant challenge for Machine Translation in this context. The borrowings make up 24.3% of errors. This indicates that Google Translate incorrectly uses loanwords or words borrowed from other languages when translating Covid-related terms, possibly due to lexical overlap between languages or inadequate adaptation to the target language's conventions. The "false friends" and stylistic errors each account for 14.5% of errors. Their significant proportion suggests that Google Translate may often mistake such words, leading to misinterpretations of Covid-related concepts. The stylistic errors imply that Google Translate has trouble maintaining the appropriate formality, tone, or style in the translation of Covid terms. Given the seriousness of the pandemic, the incorrect tone or style could lead to misleading or inappropriate translations.

Regarding the arbitrary combinations that account for only 1.8% of the total errors, their presence indicates that random or nonsensical word pairings are relatively rare. However, they still pose a problem in maintaining clarity and coherence when discussing specialized topics like the pandemics. The chart highlights that the main issues in Google Translate's handling of Covid-related terms are collocation errors and borrowings. Addressing collocation issues could significantly improve translation accuracy in critical contexts like public health communication.

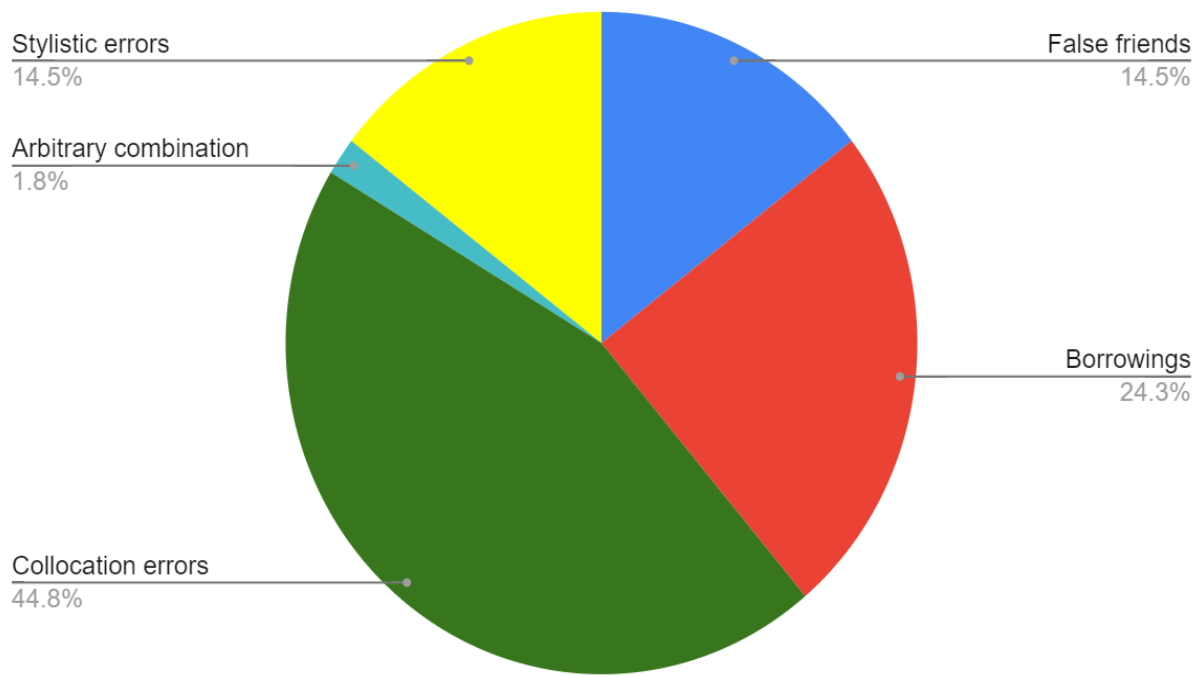

Figure 6. The representations of the values

## 3.1. Word Error Rate (WER)

First of all, it is essential to understand the scale and context of the WER. The Word Error Rate typically ranges from 0 to 1, where 0 indicates a perfect translation (no errors), and 1 implies that the entire translation is incorrect.

For the errors made in the abbreviations and acronyms displayed above, the WER value is 0.98, a result that suggests a high rate of error in translation. A WER of 0.98 is extremely high, but one has to take into consideration that the value of WER was calculated for individual terms or structures of terms, not for an entire text. However, it's highly probable that the translation is inaccurate strictly referring to the terminology of Covid-19 and may not effectively convey the intended meaning. Abbreviations and acronyms, especially related to technical fields like Covid-19, pose challenges for Machine Translation systems. They require accurate translation and context preservation, which might be difficult for machine translators to handle effectively. High WER highlights the need for improvements in Machine Translation systems, especially for specialized fields such as healthcare and public health. Continuous development and refinement of Machine Translation algorithms can enhance accuracy and usability. This underscores the importance of human review, context awareness, and ongoing advancements in Machine Translation technology.

The WER employed to assess the quality of translations of Covid-19 terminology from English into Romanian produced by Google Translate obtained the value of 0.89, which corresponds to an error rate of 89%. Out of a total of 55 Covid-19 specific terms translated, 26 were found to be inadequate by insertion, deletion and substitution. The errors varied in severity, ranging from minor lexical discrepancies to major mistranslations. For instance, "decese" was used instead of the more accurate "decese zilnice" for "daily deaths", constituting a minor error. On the other hand, a major lexical error was noted in the translation of "curfew" as "stare de asediu" instead of the correct "restricţii de oră". Also, "lockout" was translated by "blocare" instead of "şomaj tehnic", substitution that is semantically not related with the term from the target language. Another relevant term for the pandemic period is "social tracing" that was translated by "urmărire socială" instead of "urmărirea contactelor". This literal translation ("socială") given by Google Translate made the translation confusing and irrelevant to the context. Instead of translating "intensive care medicine" by the Romanian acronym "ATI" (Anestezie şi Terapie Intensivă), that has over 42,000,000 occurrences on Google search, Google Translate gave the translation "medicină de terapie intensivă", a collocation that is not usually used in Romanian related to Covid-19.

The high Word Error Rate (WER) of 0.89 (89%) highlights significant inadequacies in the performance of Google Translate when translating Covid-19 terminology from English into Romanian. This strikingly high error rate means that out of every 100 words translated, 89 are likely to contain errors, which signals serious concerns about the tool's reliability in handling highly specialized content. The fact that 26 erroneous translations out of 55 terms were identified underscores the unreliability of Google Translate as a standalone tool for translating specialized medical and public health terminology with accuracy and precision.

Similar findings regarding the performance of Google Translate with specialized terminology, especially in medical contexts such as Covid-19-related terms, have been reported in other language pairs. Several studies have observed that Google Translate's accuracy tends to be lower when translating highly specialized content like medical terminology, which includes technical terms, medical jargon, and context-specific expressions. According to Garcia-Serrano and Cruz (2020), a study evaluating Google Translate's performance in translating medical terminology between English and Spanish found that the translation accuracy for medical terms was notably lower than for general text. The error rate for medical terms in some instances exceeded 30%, similar to the high error rates observed in your study for English to Romanian. Before the pandemic, in German, a study examining Google Translate's medical translations reported similarly high error rates for specialized medical terms, especially for technical terms and complex healthcare-related phrases. Misinterpretations were common, pointing to significant challenges in ensuring reliable and accurate medical translation, according to Delfani et al. (2024).

These findings are particularly alarming in the context of medical and public health communication, where accuracy is paramount. Major lexical errors, such as incorrect word choices, improper collocations, and borrowings, have the potential to distort the intended message, leading to misunderstandings and even misinformation. Such inaccuracies in translations can result in confusion among healthcare professionals and the public, potentially leading to detrimental consequences, especially in the context of a global health crisis like Covid-19, where clear and accurate communication is critical to managing the pandemic effectively.

The types of errors observed in the Google Translate output, such as collocation errors (the largest error category at 44.8%), borrowing errors (24.3%), false friends, and stylistic issues, reveal that the tool struggles significantly with the nuances and complexity of specialized terminology. For example, collocation errors—the incorrect pairing of words that are frequently used together—are particularly concerning in medical translations, as they can alter the meaning of crucial information about treatments, symptoms, or health protocols. Similarly, borrowing errors suggest that Google Translate might over-rely on direct word transfers from the source language, leading to inaccurate or culturally inappropriate translations in the target language, further contributing to confusion.

Based on the WER result of 0.89 and the detailed error analysis, it is evident that while Google Translate can be useful for general, everyday translations, it falls drastically short in translating specialized Covid-19-related terms from English to Romanian with the required accuracy. Relying on Google Translate alone for medical or public health communications poses risks that could affect both the quality of care provided and the public's understanding of important health information. As such, professionals working with Covid-19 content – whether in medical, scientific, or public health fields – should exercise great caution when using Machine Translation tools. It is strongly recommended that translations of specialized terms, particularly in high-stakes environments like healthcare, be reviewed and verified by human experts or professional translators to ensure clarity, accuracy, and appropriateness for the target audience.

In conclusion, the high error rate and types of lexical inaccuracies demonstrated by Google Translate in handling Covid-19 terminology highlight the tool's limitations in translating specialized and technical content. To avoid the risk of miscommunication and its potentially harmful effects, Google Translate should be used with caution, and all critical translations should be subject to expert verification – especially in fields as sensitive and impactful as healthcare and public health communication.

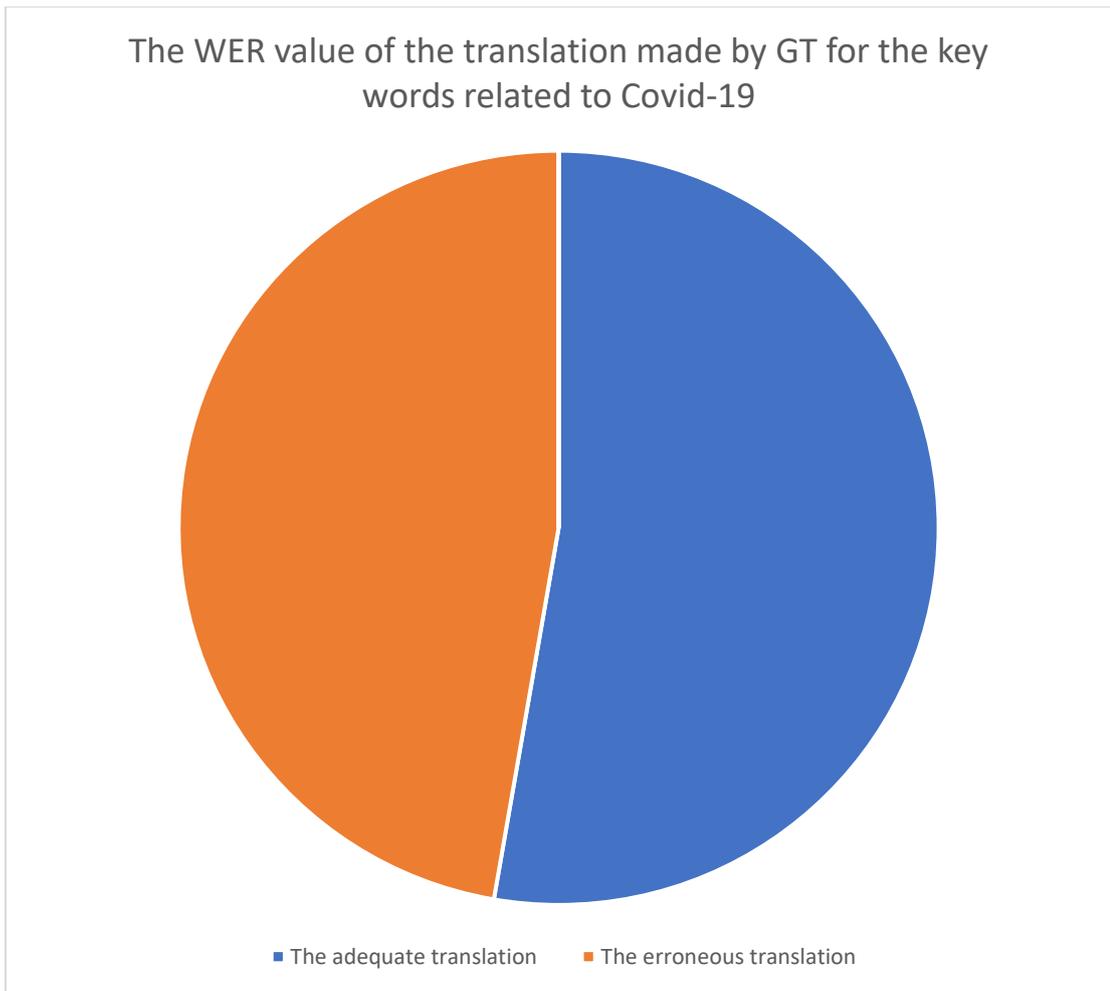

Figure 7. The WER value

## 3.2. BLEU score

BLEU score is designed to evaluate the overall quality of translations by comparing the machine-generated translations with human reference translations. It calculates the precision of n-grams (sequences of n words) in the machine-generated translation compared to the reference translation. However, in specialized domains such as Covid-19, the context and semantic accuracy of translations are paramount. BLEU score does not account for the contextual appropriateness and semantic accuracy of translations, focusing instead on the surface-level similarity. It provides a holistic evaluation of the entire translated text and does not differentiate between various types of errors, including lexical, syntactic, and semantic errors. In the context of this study, which aims to evaluate lexical errors specifically, BLEU score does not offer the granularity required to identify and analyse specific lexical errors made by Google Translate. Given that the texts selected for this study are related to Covid-19, a specialized medical and public health domain, the lexical accuracy and appropriateness of translations are crucial. BLEU score, being a general metric, may not be sensitive enough to capture the domain-specific nuances and lexical intricacies, thus rendering it less suitable for evaluating translations in specialized domains.

To achieve a detailed and accurate evaluation of lexical errors, a manual error annotation methodology was employed. This approach involves experts annotating and categorizing the lexical errors made by Google Translate in the translated texts. This manual annotation allows for a nuanced analysis of the types and frequencies of lexical errors, offering valuable insights into the translation quality. The manually annotated lexical errors were systematically classified and analysed to identify patterns, trends, and recurrent types of lexical errors made by Google Translate. While BLEU score remains a valuable metric for evaluating Machine Translation quality, its limitations in capturing specific lexical errors, especially in specialized domains like Covid-19, necessitate the adoption of alternative evaluation methodologies. The manual error annotation and systematic error classification and analysis methodologies employed in this study offer a detailed and nuanced evaluation of Google Translate's lexical errors, thus providing valuable insights into the areas requiring improvement in Machine Translation quality.

Interpreting a BLEU (Bilingual Evaluation Understudy) score for translations carried out by Google Translate in a corpus of 230 texts related to Covid-19 involves understanding the quality and accuracy of the translations in comparison to reference translations. A BLEU score ranges from 0 to 1, where 1 indicates a perfect match between the candidate translation and the reference translations. Higher BLEU scores generally indicate better translation quality, with scores closer to 1 implying higher accuracy and fluency in translation. For Covid-19-related texts, accuracy and clarity are paramount due to the critical nature of the information being conveyed. Therefore, a higher BLEU score is desirable to ensure that translations effectively communicate important health-related information. A high BLEU score in Covid-19-related translations indicates that Google Translate effectively captures the meaning and context of the texts, providing accurate and coherent translations. Conversely, a lower BLEU score suggests that translations may contain errors, inaccuracies, or inconsistencies compared to the reference translations. This could potentially lead to misunderstandings or misinterpretations of critical information related to Covid-19.

**BLEU:** $\qquad$ 0.44

**Precision x brevity:** 0.54 x 81.58

|  | 1-gram | 2-gram | 3-gram | 4-gram |
|---|---|---|---|---|
| • **Type** | | | | |
| • **Individual** | 2.40 | 0.45 | 0.33 | 0.24 |
| • **Cumulative** | 1.95 | 0.85 | 0.58 | 0.44 |

Figure 8. BLEU score result

In the figure above, the BLEU score of 0.44 for the Covid-19-related keywords translated by Google Translate reveals a relatively low level of accuracy when compared to human reference translations. This BLEU score reflects the Machine Translation 's inability to precisely replicate the human translation, suggesting that Google Translate is struggling to maintain high-quality translations in the specialized domain of public health and medical information. The BLEU score is based on n-gram precision, which evaluates how well the machine-generated n-grams (sequences of words) align with the human reference. In this case, the individual n-gram scores demonstrate a sharp decline as the n-grams become longer: from 2.40 for 1-grams to 0.24 for 4-grams. This significant drop across 2-grams, 3-grams, and 4-grams indicates that while Google Translate may succeed in translating isolated words or short phrases (1-gram), it struggles with longer, more complex sequences of words. The cumulative n-gram scores similarly show a pattern of diminishing quality from 1.95 for 1-grams to 0.44 for 4-grams, reinforcing the notion that translation quality decreases as the length and complexity of phrases increase.

The precision x brevity score of 0.54 x 81.58 suggests that while the Machine Translation attempts to be precise in its output, the brevity penalty reflects that the machine-generated translations are shorter than the reference translations, which further contributes to the low BLEU score. In contexts where maintaining complete and nuanced meaning is crucial—such as translating pandemic-related terms and guidelines—the brevity of translations may lead to omissions of critical information, which can cause miscommunication and misunderstandings. In the case of Covid-19, the specialized terminology and context-sensitive language require high levels of accuracy and semantic understanding, areas where Machine Translation systems like Google Translate often struggle. The low BLEU score suggests that the Machine Translation introduces lexical, syntactic, and semantic errors that hinder the precise communication of pandemic-related concepts. For instance, terms like "social distancing" or "droplet transmission" might not only require accurate word-for-word translation but also an understanding of the cultural and medical implications of these terms, which a Machine Translation might miss.

Furthermore, the low BLEU score could also indicate that Google Translate has difficulty handling collocations, idiomatic expressions, and domain-specific medical terms that are essential for public health communication. Given that the medical field relies heavily on precise language to convey critical information, even minor translation errors could lead to misunderstandings or misapplication of health guidelines, thereby posing significant public health risks. In conclusion, the BLEU score of 0.44 highlights the limitations of Machine Translation when tasked with translating specialized content, such as Covid-19-related terms from English to Romanian. The drop in n-gram precision and the brevity penalty underline the need for more advanced and context-aware translation systems, especially in domains where accuracy and semantic fidelity are paramount. This finding underscores the importance of human oversight in ensuring the quality of translations in sensitive areas such as healthcare and public safety.

# Chapter V

# Conclusions

## 1. Conclusions

The analysis of lexical errors made by Google Translate highlights both the potential benefits and the risks associated with translating documents from English into Romanian, particularly in the context of Covid-19-related information. While Machine Translation offers an efficient and accessible means of disseminating critical information to diverse audiences, it also poses significant risks due to the occurrence of (lexical) errors. This research is grounded in a comprehensive corpus of translation errors systematically collected from various translated texts, allowing for a detailed examination of the specific challenges posed by lexical inaccuracies. By analyzing the error corpus, the study uncovers recurring patterns, such as mistranslations of medical terms, incorrect use of idiomatic expressions, and culturally inappropriate word choices. These errors can distort the intended meaning of critical public health messages, potentially leading to misunderstandings or even harmful outcomes. Moreover, the corpus-based approach allows for a quantitative assessment of error frequency and severity, offering valuable insights into the limitations of Google Translate when handling specialized terminology. For instance, errors in translating medical jargon or terms with nuanced meanings can affect the overall coherence and accuracy of a text, thereby undermining trust in the information provided. In contexts like the Covid-19 pandemic, where timely and precise communication is paramount, these issues become particularly significant. By focusing on a corpus of real-world errors, this research not only identifies specific lexical pitfalls in Machine Translation, but also underscores the need for human oversight, especially in critical areas such as healthcare communication.

Several recent studies have examined the errors made by Machine Translation systems, particularly Google Translate, when translating medical documents into various languages. These studies have highlighted significant challenges and inaccuracies that can arise in medical contexts. For example, a study by the University of California[38] evaluated Google Translate's accuracy in translating doctor's orders into Spanish and Chinese. The findings indicated that while the translations were generally understandable, there were notable errors, especially with longer, jargon-filled sentences. This raises concerns about the reliability of Machine Translation in medical settings. Research published in *The BMJ*[39] assessed the accuracy of Google Translate for medical phrase translations across multiple languages. The study found an overall accuracy of 57.7%, concluding that the tool should not be trusted for important medical communications due to the high potential for errors. An article in *Slator*[40] discussed a study indicating that while Google Translate has shown improvements, it remains unreliable for use in medical emergencies. The study highlighted that inaccuracies in translations could lead to serious misunderstandings in critical situations. An article from *The Verge*[41] reported on a study that found Google Translate still isn't reliable enough for medical instructions for people who don't speak English, emphasizing the potential risks of relying on Machine Translation for medical communication. All these studies underscore the limitations of Machine Translation tools like Google Translate in accurately conveying medical information across different languages. The potential for errors thus poses significant risks in healthcare settings, and it emphasizes the need for professional human translators to ensure accuracy and patient safety.

---

[38]    https://www.ucsf.edu/news/2019/02/413376/google-translates-doctors-orders-spanish-and-chinese-few-significant-errors.

The findings of this study underscore the importance of recognizing the limitations of Machine Translation systems, such as Google Translate, especially when dealing with specialized domains like medicine and public health. The lexical errors introduced during the translation process can lead to confusion and misinformation, potentially compromising the accuracy and reliability of the translated content. In specialized fields like medicine, where precision and clarity are of utmost importance, even minor mistranslations can lead to severe and potentially life-threatening consequences. Researchers such as García-Serrano and Cruz (2020) emphasize that the stakes in medical translation are exceptionally high. Errors in translating medical instructions, for example, may result in patients misunderstanding how to take prescribed medications or follow treatment protocols, thereby jeopardizing their health. Similarly, inaccuracies in public health guidelines can lead to widespread confusion, misinformation, or improper implementation of critical health measures. This is particularly concerning in contexts such as the Covid-19 pandemic, where the accurate dissemination of health information plays a vital role in ensuring public safety and preventing the spread of the virus. The research highlights the critical need for rigorous quality assurance mechanisms in Machine Translation systems and underscores the importance of involving human translators, particularly in domains where precise and unambiguous communication is essential. These issues underline the critical need for human oversight in verifying machine-generated translations, particularly when conveying essential health information that may directly impact public safety. Furthermore, this study emphasizes the broader implications of relying solely on automated translation tools in high-stakes environments.

---

[39] https://www.bmj.com/content/349/bmj.g7392?.
[40] https://slator.com/google-translate-not-ready-for-use-in-medical-emergencies-but-improving-fast-study.
[41] https://www.theverge.com/2021/3/9/22319225/google-translate-medical-instructions-unreliable.

Without expert review, Machine Translation in critical fields like medicine and public health poses serious risks. Misinterpretations can undermine trust, compromise patient safety, and disrupt communication. García-Serrano and Cruz (2020) note that MT often fails to accurately translate medical jargon and culturally specific terms.

Such an approach ensures that the translation process does not merely convey the correct literal meaning but also maintains coherence, cultural sensitivity, and contextual accuracy across different languages. Cheong (2023) supports this notion by advocating for collaborative models that integrate human reviewers into the translation workflow, particularly for specialized content such as healthcare or legal documents. The involvement of human professionals provides a crucial layer of quality assurance, enabling corrections to lexical errors, improving fluency, and ensuring that culturally appropriate language is used. Furthermore, the hybrid approach is especially critical in public health communication, where inaccuracies in translation can create barriers to understanding and compliance with health measures. For example, during the Covid-19 pandemic, poorly translated public health guidelines were shown to exacerbate misinformation and reduce the effectiveness of interventions in multilingual communities. According to Lăzăroiu et al. (2021), language barriers and mistranslations contributed to vaccine hesitancy and the spread of misinformation. They emphasized the need for clear and accurate multilingual communication to build trust and ensure the effectiveness of public health interventions. By combining the speed and efficiency of Machine Translation with the nuanced understanding of human professionals, hybrid systems can ensure not only the accuracy of information but also its accessibility and relatability to diverse audiences. This model ultimately enhances trust, promotes inclusivity, and improves the overall quality of cross-linguistic communication in specialized domains.

During the pandemic, the demand for translation services surged, driven by the need to disseminate crucial information, translate medical research, facilitate international collaborations, and communicate with diverse populations. Machine Translation tools, such as online translation services or apps, experienced increased popularity due to their accessibility and convenience. However, the rapid pace at which translations were needed, combined with the unique challenges posed by the pandemic, resulted in an increase in translation errors. The language used in discussions about Covid-19 often includes domain-specific terminology and specialized vocabulary related to public health. Faber and Rodriguez (2012) delineate the distinction between domain-specific terminology and specialized vocabulary within the context of public health. Domain-specific terminology refers to a set of terms that are uniquely pertinent to a particular field of study, encompassing precise definitions that are universally recognized by professionals within that domain. In public health, this includes terms such as "epidemiology," "immunization," and "quarantine," which have specific meanings and applications understood by experts. Specialized vocabulary, on the other hand, comprises a broader set of terms that, while relevant to the field, may also be used in other contexts but acquire particular nuances within the domain. For instance, words like "screening" and "surveillance" are commonly understood in general language but carry additional, more specific connotations in public health. The differentiation highlighted by Faber and Rodriguez (2012) is crucial for ensuring clarity and precision in communication, particularly in interdisciplinary and multilingual environments where accurate understanding of terms is essential for effective public health practice and research. Translating such language requires familiarity with the subject matter and an understanding of its linguistic nuances to ensure accurate and effective communication. For example, the error made by Google Translate in translating "curfew" by "stare de asediu" underscores the importance of linguistic sensitivity, contextual understanding, and domain expertise in translation. Achieving accurate and contextually appropriate translations, especially in discussions about Covid-19, requires a nuanced approach of the source text.

Google Translate relies on statistical models and lacks human-like comprehension. It can effectively translate simple phrases, but often struggles with context-rich and specialized language, such as medical or scientific jargon. This is because such language requires an understanding of specific terminology and nuanced meanings that statistical models cannot fully capture, leading to less accurate translations and underscoring the need for human expertise and advanced linguistic algorithms. For instance, Google Translate may translate terms like "smuggle" literally, without considering its contextual usage in discussions of public health or illicit vaccine transportation. Similarly, highly specialized terms such as "contact tracer," "virus transmission rate," "discharge from the nose," "aches and pain," "respiratory etiquette," "Covax facility," "cold chain," "respiratory medicine," "respiratory droplets," and "fitted mask" were rendered inaccurately or ambiguously. Such mistranslations can lead to confusion among patients, healthcare providers, and policymakers, undermining efforts to effectively communicate essential medical information or public health guidance. Moreover, terms such as "contact tracer", "Covax facility", "respiratory etiquette" represent essential concept in the medical Covid-19 related context, and might be misinterpreted if the system lacks cultural and contextual understanding of global health initiatives. Also, similar terms might be rendered in a way that confuses its practical application, in non-English-speaking contexts.

Relying on statistical models trained on large corpora of text, Google Translate can recognize patterns and associations between words, allowing it to provide quick translations across a wide range of languages. However, these models often struggle with understanding deeper linguistic and cultural context in the way humans do. While Google Translate excels at identifying frequent word combinations and basic translations, it lacks the ability to discern subtle distinctions, such as specific connotations, tone, or context-dependent meanings. This limitation becomes particularly evident when translating specialized terminology or phrases with nuanced meanings, like those associated with public health emergencies, such as Covid-19.

For example, the term "lockout," when used in the context of the Covid-19 pandemic, conveys a specific meaning related to widespread restrictions, quarantines, or shutdowns. Google Translate may not always accurately capture the precise connotations of this term, especially if the system is drawing from unrelated contexts (such as labor strikes or sports) where "lockout" has different implications. In contrast, a more contextually appropriate translation, like the Romanian word "izolare," may better convey the essence of "lockout" in relation to public health measures during the pandemic, emphasizing isolation and confinement. This highlights how nuanced understanding of context is crucial, especially when translating terms that may evolve or carry specific meanings in response to a crisis.

Moreover, translating words between languages is not merely a process of seeking direct formal equivalence, as discussed in translation theory by scholars such as Nida (1964). Instead, it requires achieving dynamic equivalence, where the translation conveys not only the linguistic structure, but also the cultural, pragmatic, and situational nuances that give the source text its full meaning. According to traductology principles, effective translation strategies must go beyond literal rendering, as specialized terms and expressions often carry connotations and contextual implications that cannot be conveyed through word-for-word translation. For example, the functional equivalence approach emphasizes adapting a term's meaning to fit the target audience's linguistic and cultural framework, ensuring the message resonates as it would in the original context. Translators must also employ strategies such as explicitation, modulation, or cultural substitution to bridge gaps in cultural understanding and accurately reflect the source text's intended meaning. In specialized domains such as medical or legal translation, these strategies are particularly critical, as terms often have no perfect lexical match in the target language and require contextual adaptation to maintain clarity and precision. The challenge is that languages rarely have one-to-one correspondence for words and concepts, especially for abstract or context-sensitive terms. In the case of "lockout" and "izolare," while both may refer to restrictions, the latter potentially carries a stronger connotation of enforced isolation, which better aligns with the public health objectives during a pandemic. This reflects the importance of achieving *dynamic equivalence*—conveying the intent and effect of the source text in the target language rather than merely mirroring its lexical structure. Producing accurate translations requires translators to go beyond formal equivalence and prioritize the communicative function of the terms within their specific context. Machine Translation systems like Google Translate, which rely on statistical or neural models, often fall short in this regard, as they lack the ability to interpret deeper semantic relationships or the cultural and situational nuances essential to contextual adaptation.

Another challenge for Machine Translation systems like Google Translate is keeping pace with the changing language and terminology surrounding dynamic events such as the Covid-19 pandemic. For example, words like "social distancing," "quarantine," and "lockdown" quickly became part of the everyday lexicon during the pandemic, and their meanings have shifted or expanded in various ways depending on the cultural and legal context. Google Translate, trained on large datasets, may not always reflect the most current usage or terminology, as its models may lag behind recent developments or fail to capture emerging linguistic trends. This delay can result in outdated or less accurate translations (see the translation for *breakthrough infection, curfew, discharge from the nose, infodemic, location, oncology pipeline, resolution,* *screening, service, shutdown, spike protein, T helper cells, etc.*), particularly in contexts where precision and immediacy are critical, such as public health communication.

Furthermore, pandemic-related terms are often laden with emotional weight and urgency, and subtle mistranslations could have far-reaching consequences. A failure to accurately convey the gravity of measures like "lockout" or "quarantine" could lead to misunderstandings or even non-compliance with public health directives. For instance, in a health crisis where isolation measures are central to controlling virus spread, mistranslating "lockout" as something less severe or less urgent might downplay the importance of those measures. This underscores the importance of not only translating words but also conveying the appropriate tone and severity that reflect the underlying meaning of those terms. While Google Translate can recognize word patterns and associations from large datasets, it struggles to accurately grasp the context and evolving nuances of language, particularly in specialized fields like public health. Machine translation tools may not always be capable of handling the rapid linguistic shifts that accompany crises like Covid-19, and they often fail to capture the deeper meaning and urgency that are critical in such scenarios. Therefore, in high-stakes situations like pandemic communication, it is essential to combine Machine Translation with human oversight to ensure translations reflect the most current and contextually appropriate usage of terminology, minimizing the risks of misunderstanding and misinformation.

The corpus analysis has attempted to provide useful information about translations that indicate particular problems, confusions, misunderstandings or meaning alterations. It has also attempted to identify common lexical and semantic errors, outlining the contextual challenges, questioning the need of improvements. The suggestions for improving the Google Translate system, especially in the context of Covid-19 related content (especially for Romanian) include refining algorithms, incorporating domain-specific dictionaries, or enhancing contextual understanding. Collocation errors are the dominant type of errors identified in this study. When translating specialized terms or concepts, it is essential for translation tools like Google Translate to consider not only individual word meanings but also the collocational patterns and idiomatic expressions that are prevalent in the target language. Failure to do so can result in translations that are inaccurate or contextually inappropriate.

In our analysis, the high Word Error Rate (WER) and BLEU scores noted highlight significant challenges when relying solely on Machine Translation systems for critical translations, such as those required for medical information during a pandemic. These metrics reveal that while Machine Translation tools can offer substantial support, they often fall short in capturing the nuanced and context-specific language essential for accurate communication.

Therefore, it is recommended that official translations of Covid-19-related information into Romanian, or any other language (especially if it is considered a minor one), involve a combination of Machine Translation and human review processes. This approach can help mitigate the risk of lexical errors and ensure that translated documents convey accurate and reliable information to the target audience. Additionally, ongoing advancements in Machine Translation technology and the development of specialized models for domain-specific translation tasks may offer opportunities to improve the efficacy and reliability of automated translation systems in the future.

# 2. Speculations for future research

The brief overview of the history of MT marks the continuous evolution of this translation service. The development of technology, in general, has also played an essential role in this respect. With advancements in technology, translation methods have evolved to incorporate various new features and capabilities. Machine Translation has undergone significant advancements that have profoundly transformed how we communicate across different languages. Among these notable enhancements are real-time translation capabilities, the integration of Neural Machine Translation techniques, and the development of speech-to-text features.

More recently, the emergence of Neural Machine Translation (NMT) has revolutionized the field. NMT employs deep learning techniques, specifically artificial neural networks, to translate text. This approach enables the model to learn representations of language and handle complex syntactic and semantic structures. NMT has demonstrated superior translation quality, with output that is often more fluent and accurate than previous methods. The rise of NMT has been further fuelled by the availability of large-scale parallel corpora and advancements in computational power. Koehn (2017: 188) highlights the superior translation quality of NMT systems, emphasizing how deep learning techniques and large-scale parallel corpora have enabled NMT to surpass traditional statistical methods in accuracy and fluency: "With the availability of large-scale parallel corpora, neural models can learn complex linguistic patterns, resulting in more natural and contextually appropriate translations compared to earlier statistical methods."

Machine Translation has profoundly influenced global communication by bridging language barriers across sectors such as business, healthcare, and diplomacy. It has facilitated multilingual interactions and content creation while evolving from rule-based systems to statistical models and, ultimately, Neural Machine Translation. Koehn (2017: 198) highlights NMT's transformative impact on translation quality and international communication, stating that "by leveraging deep learning architectures and large-scale training data, NMT has improved translation coherence, making it particularly effective for real-world applications in business, healthcare, and cross-cultural communication."

In the realm of Machine Translation, the analysis of lexical errors made by Google Translate in translating medical texts and articles related to Covid-19 from English into Romanian provides valuable insights for future research directions. As we look ahead, several avenues for exploration emerge, paving the way for advancements in Machine Translation technology and enhancing the quality of translations in specialized domains such as medicine and public health.

Future research could focus on a deeper understanding of the linguistic and computational factors that contribute to lexical errors in machine-translated medical texts. Given the complexity of medical discourse, which often involves specialized terminology and syntactic structures, examining the specific linguistic features of this domain is crucial. This would enable the development of more tailored algorithms and models capable of accurately translating medical concepts. For instance, Machine Translation systems must handle medical terminology that may not have direct equivalents in target languages, requiring advanced techniques to ensure both accuracy and fluency, according to García-Serrano and Cruz (2020).

As machine learning technologies continue to evolve, there is also a growing need to explore domain-specific training techniques. By utilizing domain adaptation methods and incorporating specialized medical knowledge into the training process, researchers can refine Machine Translation systems to better capture the nuances of medical language. This can help reduce lexical errors and improve the overall quality of translations in healthcare contexts. Advances in Neural Machine Translation, particularly the integration of domain-specific corpora, can improve both accuracy and contextual relevance in medical texts, according to Jurafsky and Martin (2019).

Moreover, the recent development of more sophisticated language models, such as ChatGPT and Google Gemini, offers promising opportunities for enhancing Machine Translation quality.

These models, which are trained on larger datasets and use complex architectures, have shown improved performance in specialized fields like medicine. By incorporating these advanced models into Machine Translation workflows, researchers could explore their potential to produce more accurate and contextually appropriate translations for Covid-19-related medical texts and beyond. The integration of these models would benefit from their ability to understand nuanced medical terms and phrases, which are crucial in preventing misunderstandings in critical healthcare communication.

Looking ahead, the continued advancement of Artificial Intelligence (AI) and Natural Language Processing (NLP) holds great promise for improving Machine Translation systems. Addressing the persistent challenges posed by lexical errors in medical translations could greatly enhance communication in global healthcare settings. Ultimately, this could contribute to better healthcare outcomes and more effective public health initiatives, ensuring that critical information reaches diverse populations in an accurate and culturally sensitive manner. This approach could improve the accessibility and reliability of translated medical texts, particularly in times of health crises like the Covid-19 pandemic, supporting informed decision-making and better patient care.

Reflecting on the lexical errors made by Google Translate during the COVID-19 pandemic, it becomes evident that early AI-driven translation systems faced substantial limitations, particularly when dealing with specialized terminology, context-dependent phrasing, and culturally sensitive language. The inaccuracies in translating critical health-related documents contributed to misinformation, misinterpretation, and, at times, a diminished trust in automated translation systems. However, the current state of AI translation, marked by advancements in large language models such as ChatGPT, Gemini, and other sophisticated neural networks, has significantly improved in terms of contextual understanding, fluency, and semantic accuracy. Unlike earlier models, which often relied on direct word-to-word translation, contemporary AI systems integrate deep learning, reinforcement mechanisms, and user feedback to generate more coherent, context-aware translations. Despite these advancements, challenges persist, particularly concerning domain-specific jargon, idiomatic expressions, and ethical concerns regarding AI's role in disseminating sensitive information. As AI translation continues to evolve, its future effectiveness will hinge on a symbiotic relationship between human expertise and machine learning. Regular validation by linguists, continuous refinement of training datasets, and ethical considerations in AI deployment will be crucial in ensuring that Machine Translation not only achieves linguistic accuracy, but also maintains clarity, reliability, and cultural sensitivity, particularly in high-stakes domains such as healthcare and crisis communication. The evolution from early statistical models to today's neural network-based systems marks a transformative shift, yet it also underscores the necessity for a balanced, critically engaged approach to AI-assisted language processing.

# Appendix

## 1. Abbreviations and acronyms

| Words in English | Adequate translation into Romanian | Words translated by GT into Romanian |
|---|---|---|
| ACT<br><br>Access to Covid-19 Tools | Acceleratorul accesului la instrumente pentru combaterea Covid-19 | Acceleratorul de acces la instrumente Covid-19 |
| Advance Market Commitment (AMC) | Angajament AMC | Angajament de piață în avans |
| AI (Artificial Intelligence) | Inteligență artificială | AI |
| APC<br><br>Antigen-presenting cell | Antigene macrofage | Angajament de cumpărare anticipată (APC) |
| ARDS<br><br>Acute Respiratory Distress Syndrome | Sindromul de detresă respiratorie acută | SDRA |
| ARI<br><br>Acute Respiratory Infection | SARI<br><br>Infecție respiratorie acută severă | ARI |
| Catholic Relief Services (CRS) | Delegație Caritas | Catholic Relief Services (CRS) |
| CEO | Director general | CEO |
| CDC | DSP | Centres for Disease Control and Prevention |

| Centres for Disease Control and Prevention | Direcția de sănătate publică | CDC |
|---|---|---|
| CFR<br><br>Case Fatality Rate | RM<br><br>Rata mortalității | CFR |
| Coalition for Epidemic Preparedness Innovations | *Coaliția pentru Inovații* în Pregătirea în caz de Epidemii | Coaliția pentru Inovații în Pregătirea Epidemiei (CEPI) |
| Covid-19<br><br>Covid-19 is the name of the disease that the novel coronavirus causes. It stands for coronavirus disease 2019. | Covid-19 | Covid-19 |
| FDA (food and drug administration) | Agenția Federală pentru Hrană și Alimente (AFHA) | FDA |
| Global Polio Eradication Initiative (GPEI) | Inițiativa globală de eradicare a poliomielitei | Global Polio Eradication Initiative (GPEI) |
| Intelectual Property (IP) | Proprietate Intelectuală (PI) | Proprietate intelectuală (IP) |
| International Development Association (IDA) | Asociația Internațională de Dezvoltare (AID) | Bank International Development Association (IDA) |
| Intensive Care Unit (ICU) | Unități de terapie intensivă (UATI) | Unități de terapie intensivă (ICU) |
| LMIC<br><br>Covid-19 in low and middle income countries | - | LMIC-uri |

| National Identity Document (NID) | Card de sănătare | NID |
|---|---|---|
| OPV (Olio Polivirus Vaccine) | *Poliovirus* viu administrat pe cale *orală* | Poliovirus oral (OPV) |
| PASC (Post-Acute Sequelae of Covid-19) | Sechele post-Covid | PASC |
| PCR Polymerase chain reaction | PCR | PCR |
| PPE Personal protective equipment | EIP Echipament individual de protective | EIP |
| PUI (Patient Under Investigation) | Pacient monitorizat | PUI |
| PCV Pneumococcal conjugate vaccines | VPG Vaccinuri pneumococice conjugate | Vaccinuri pneumococice conjugate (PCV) |
| Pediatric Inflammatory Multisystem (PIMS)colloc | Sidrom Multisistem inflamator pediatric | Sindromul multisistem inflamator pediatric (PIMS-TS) |
| PTSD | Tulburarea de stress post-traumatic | PTSD |
| SARS-CoV-2 Novel coronavirus 2019 is the name of the disease caused by SARS-CoV-2 | SARS-CoV-2 | SARS-CoV-2 |
| SURGE<br><br>Systematic Urgent Review Group Effort | - | Efortului Grupului de Evaluare Urgentă Sistematică Covid-19 (SURGE) |
| UHC<br><br>Universal Health Coverage | - | UHC |
| United Nations Development Progrramme UNDP | Programul Națiunilor Unite pentru Dezvoltare (PNUD) | PNUD |
| Vaccine-preventable disease (VPD) | Boli prevenibile prin vaccinare | VPD |
| WFH<br><br>Working from home | Lucru de acasă<br><br>Muncă la domiciliu | WFH |
| WHA<br><br>World Health Assembly | Adunarea Mondială a Sănătății | WHA |
| WHO<br><br>World Health Organization | OMS<br><br>Organizația mondială a sănătății | CARE |

## 2. False friends

| Words in English | Words translated by GT into Romanian | Adequate translation into Romanian |
|---|---|---|
| accommodation | acomodare | cazare |
| Actually | actual | de fapt |
| to alter | a altera | a modifica |
| appropriate | Apropiat | adecvat |
| to assist | a asista | a ajuta |
| to assume | a asuma | a presupune |
| Audience | audiență | public |
| to advertise | a avertiza | a promova |
| Common | comun | banal |
| challenging road | drum provocatory | drum dificil |
| eventually | eventual | în cele din urmă |
| Evidence | evidență | dovadă |
| Location | locație | loc |
| resolution | rezoluție | obiectiv |
| Support | support | sprijin |

# 3. Borrowings

| Words in English | Words translated by GT into Romanian | Adequate translation into Romanian |
|---|---|---|
| advocacy | advocacy | susținere |
| Bandaid | bandaid | plasture |
| big data | big data | volume mari de date |
| bioterror attack | bioterror attack | atac biologic |
| end-to-end solution | end-to-end solution | soluție completă |
| Briefing | briefing | prezentare succintă |
| T helper cells | T helper cells | ajutătoare |
| Cluster | cluster | mulțime |
| Curfew | curfew | restricții de oră |
| Hotspots | hotspots | punctele-cheie |
| Infodemic | infodemic | dezimformare |
| know-how | know-how | soluția |
| long Covid | long Covid | Covid de lungă durată |
| vaccines pillar | vaccines pillar | pilonul de vaccinuri |
| pre-print | pre-print | înainte de imprimare |
| Randomizat | randomizat | la întâmplare |
| proxy protection | proxy protection | protejarea proximității |
| spike protein | spike protein | proteina de vârf |
| Screening | screening | testarea și examinarea |
| service (service workers) | Service | persoane care lucrează cu publicul |
| Shibboleths | shibboleths | o anumită credință |
| Smartphone | smartphone | telefon smart |
| Station | Station | ordin de a sta în casă |
| Superspreader | superspreader | un prim contaminat extrem de contagios |
| disease cluster | disease cluster | focar de boli |

| plug and play vaccines | plug and play vaccines | vaccinurile ARNm care pot construi o versiune rapidă pentru o potenţială nouă tulpină |
| --- | --- | --- |

# 4. Distortions

| Words in English | Words translated by GT into Romanian | Adequate translation into Romanian |
| --- | --- | --- |
| Unhospitalised | nehospitalizaţi | Nespitalizaţi |
| to isolate myself | să mă autoizol | să mă autoizolez |
| clear messages | mesaje clarate | mesaje clare |

# 5. Semantic errors

| Words in English | Words translated by GT into Romanian | Adequate translation into Romanian |
| --- | --- | --- |
| stagnate water | apă stagnate | apă stătută |
| detection | detecţie | Detectare |
| distribution | distribuţie | Distribuire |
| informal worker | lucrător informal | muncitor nedeclarat |
| to list | a lista | a afişa |
| Lockout | blocare | Izolare |
| Masking | mascare | a purta mască |
| additional | suplimentari | în plus |
| transmission | transmisie | Transmitere |
| Zooming | mărire | a petrece timp pe Zoom |

# 6. Collocation errors

| Words in English | Words translated by GT into Romanian | Adequate translation into Romanian |
|---|---|---|
| the long grass approach | abordarea cu iarbă lungă | abordarea firul ierbii |
| clinical trials are speeding up | se accelerează studiile clinice | studiile clinice sunt făcute rapid |
| Depth | adâncime | profunzime |
| let's flatten the infodemic curve | să aplatizăm curba infodemică | haideți să reducem valul de dezinformare |
| Shutdown | închis | închidere totală |
| what is the difference between efficacy and effectiveness | care este diferența dintre eficacitate și eficacitate? | care este diferența dintre eficacitate și eficiență |
| financial expenses | cheltuielile financiare | costuri |
| vaccination could prove a vital lifeline | vaccinarea s-ar putea dovedi un colac vital | vaccinarea s-ar putea dovedi o soluție vitală |
| oncology pipeline | conductă de oncologie | domeniul oncologic |
| adverse consequences | consecințele adverse | efecte adverse |
| we all have biases | cu toții avem părtiniri | toți suntem părtinitori |
| discharge from the nose | descărcări din nas | scurgeri nazale |
| after a steep fall-off in prenatal visits | după o scădere abruptă a vizitelor prenatale | după o reducere drastică a vizitelor prenatale |
| aches and pain | Durere | durere și suferință |
| respiratory etiquette | etichete respiratorii | coduri de conduită |
| background | Fundal | istoric Covid-19 |
| to think outside the box | a gândi în afara cutiei | a gândi liber |
| information, misinformation, and disinformation | informare, dezinformare și dezinformare | informație, dezinformare și dezinformare intenționată |

| to smuggle the instructions | a introduce contrabandă instrucțiunile | a introduce în celule instrucțiunile |
|---|---|---|
| Covax facility | instalația COVAX | inițiativa COVAX |
| cold chain | lanț de frig | lanțul rece |
| shortages of hospital beds | există lipsuri de paturi de spital | există deficit de paturi de spital |
| office worker | lucrător la birou | angajat la birou |
| respiratory medicine | medicină respiratorie | pneumologie |
| to get around this issue | pentru a ocoli această problemă | a vita această problemă |
| poor spelling and grammar | ortografie și gramatică slabe | greșeli de ortografie și de gramatică |
| respiratory droplets | picături respiratorii | picături nazale |
| to catch Covid-19 | a prinde Covid-19 | a lua Covid-19 |
| global public | public global | public larg |
| wear a properly fitted mask | purtați mască corect montată | purtați correct masca |
| the situation is cloudier | situația este tulbure | situația este neclară |
| Curfew | stare de asediu | restricții de oră |
| to break the pattern | a sparge tiparele | a ieși din tipare |
| a nose or throat swab | un tampon pentru nas și gât | test antigen cu bețișor cu tampon de vată cu care se recoltează din nas și din gât |
| this is all well and good | toate acestea sunt bune și bune | asta e foarte bine |
| to keep the baby | a ține copilul | a păstra copilul |
| contact tracers | urmăritorii de contact | personal de investigare epidemiologică |
| distribution of some vaccines that require sub-zero storage | vaccinuri care necesită stocare sub zero | distribuția unor vaccinuri care necesită păstrare la temperaturi sub zero grade |
| fact checkers | verificatorii de fapte | analist de presă |

# 7. Arbitrary combination

| Words in English | Words translated by GT into Romanian | Adequate translation into Romanian |
|---|---|---|
| the hammer approach | abordarea ciocanului | abordarea agresivă |
| breakthrough infection | infecție episcopală | infecție care a recidivat |

# 8. Preposition partners

| Words in English | Words translated by GT into Romanian | Adequate translation into Romanian |
|---|---|---|
| to become ill with Covid-19 | a se îmbolnăvi cu Covid | a se îmbolnăvi de Covid |
| it helps us to protect ourselves from illness | ne ajută să ne protejăm în a ne îmbolnăvi de acel germen în viitor | ne ajută să ne protejăm de îmbolnăvirea cu acel germen în viitor |
| on August 27th | la 27 august | pe 27 august |
| to be involved with | a fi implicat cu | a fi implicat în |

# 9. Stylistic errors

| Words in English | Words translated by GT into Romanian | Adequate translation into Romanian |
|---|---|---|
| The new variant is spreading quickly | Noua variantă răspândire rapidă | Noua variantă se răspândeşte rapid |
| The vaccine protects against severe illness | Vaccinul protejează de boli grave | Vaccinul protejează de îmbolnăvire gravă |
| people wear masks | poartă oamenii măşti | oamenii poartă măşti |
| flatten the curve | curbă netezită | netezirea curbei |
| virus transmission rate | virus transmitere rată | rata de transmitere a virusului |
| Whose agenda might I be supporting by sharing it? | A cui agendă aş putea sprijini prin împărtăşirea ei? | Pe cine aş putea ajuta prin distribuirea acestei informaţii? |
| A bad rush | O erupţie cutanată rău | O erupţie cutanată cu evoluţie gravă |
| Relative ease | Relativă uşurinţă | Uşurinţei relative |
| Antibodies take time to appear | Anticorpii durează timp să apară | Durează până apar anticorpii |
| places with people crowded together | locuri cu oameni înghesuiţi | locuri aglomerate |
| Spending less time in the workplace | Petrecerea mai puţin timp la locul de muncă | dacă petreci mai puţin timp la locul de muncă |